\documentclass{article}
\input{tikzit.tikzdefs}

\title{\vspace*{-1.8cm}Towards Compositional Interpretability for XAI}

\author{Sean Tull, Robin Lorenz, Stephen Clark, Ilyas Khan, Bob Coecke\\[0.04cm]
{\small  \{\nolinkurl{sean.tull, robin.lorenz, steve.clark, ilyas, bob.coecke}\}\nolinkurl{@quantinuum.com}} \\[0.05cm]
Quantinuum, 17 Beaumont Street, Oxford, UK }
\date{}

\begin{document}

\maketitle

\begin{abstract}
Artificial intelligence (AI) based on machine learning, while highly successful in many applications, currently relies largely on black-box models which lack \emph{interpretability}. The developing field of eXplainable AI (XAI) strives to address this major concern, being most critical in high-stakes areas such as the financial, legal and health sectors.

We present an approach to defining AI models and studying their interpretability based on \emph{category theory}. For this we take a compositional viewpoint, employing the notion of a \emph{compositional model}, which sees a model in terms of formal \emph{string diagrams} which capture its abstract structure together with its concrete implementation. This view is comprehensive and incorporates deterministic, probabilistic and  quantum models. We demonstrate explicitly how a wide range of AI models can be seen as compositional models, ranging from linear and rule-based models, to (recurrent) neural networks, transformers, VAEs, and causal and DisCoCirc models. 

This analysis provides the grounds for a meaningful comparison of different models, along with a definition of \emph{interpretation} of a model in terms of its compositional structure. We demonstrate how one may analyse the interpretability of a model and use this to clarify common themes in XAI within this broad compositional perspective. In particular, the approach recognises the standard notion of `intrinsically interpretable' models as essentially compositional, finding that what makes these models so transparent is brought out most clearly diagrammatically. This leads us to the new, more general notion of what we call \emph{compositionally-interpretable} (CI) models, which in addition to linear and rule-based models include, for instance, causal models, conceptual space models, and DisCoCirc models.

We explicitly demonstrate the explainability benefits that CI models can offer, based on their rich compositional structure, none of which are available for black-box architectures. Firstly, their structure may allow the computation of other quantities of interest in terms of their components, and may correspond to structure in the phenomenon being modelled, facilitating inference about the world from the model. Secondly, we show that CI models allow for several forms of diagrammatic explanations for their behaviour, respectively in terms of \emph{influence constraints}, \emph{diagram surgery} and the novel notion of \emph{rewrite explanations} using graphical equations. 
Finally, we discuss many directions for a further exploration of the approach and overarching vision, in particular raising the question of how to learn such meaningfully structured models in practice.
\end{abstract}

\tableofcontents


\section{Introduction} \label{sec:intro}




Artificial intelligence (AI) based on machine learning (ML) has achieved enormous practical successes over the past decade, but today is faced with major concerns about  its lack of \emph{interpretability}. Indeed, the explainability of AI has become an increasingly prevalent theme in public discussions around its safety and role in our society. Ultimately, this can be said to stem from the 
\rlc{\emph{black-box}} nature of ML models: though neural networks can be trained effectively to solve a given task, their lack of explicit meaningful structure means that the manner in which they do so is largely inaccessible to us from the outside. 
The absence of interpretability is a particular concern in sensitive areas, such as the \irs{financial}, legal and health sectors, where accountability and a strong kind of transparency are required and an ethical and responsible use of AI needs to be established \cite{Rudin_2019_StopExplainingBlackBoxes,gunning2019xai,xu2019explainable}.


In response, the growing field of \emph{eXplainable AI (XAI)} is attempting to solve the interpretability problem, largely through the exploration of so-called `post-hoc' techniques which take a trained AI model and aim to give explanations for either its overall behaviour, or individual outputs. \ir{Examples include saliency maps for images \cite{simonyan2013deep, zeiler2014visualizing, selvaraju2017grad}, Shapley values which rank the importance of input features \cite{LundbergEtAl_2017_SHAP_Paper_UnifiedApproachToInterpreting}, and counterfactual explanations of outputs \cite{WachterEtAl_2017_CounterfactualExplanations}.} 

However, the field of XAI has numerous issues to overcome \cite{FreieslebenEtAl_2023_DearXAICommunity}, and several authors including Rudin \cite{Rudin_2019_StopExplainingBlackBoxes} have pointed out the problematic nature of many of the `explanations' produced by such methods. 
Rather than relying on after-the-fact, and by their nature limited, explanations of standard AI models, these authors argue that \st{in sensitive high-stakes areas, such as in decisions on bail, loans or hiring,} one should instead make use of models which are \emph{intrinsically interpretable}: that is, come with explicit structure which is meaningful to us from the outside. 
\rl{Such intrinsically interpretable models do not require the post-hoc methods of XAI, serving instead as their own explanation, and one of a deeper kind, which is manifest, principled and not approximate. 
Crucially, in many cases they can perform just as well as \st{state-of-the-art} black-box models \cite{Rudin_2019_StopExplainingBlackBoxes} and hence should not be dismissed in critical high-stakes situations.}  To address these concerns, it would be desirable to have more theoretical and foundational tools which allow us to assess the interpretability of a given model. Further, one would ideally hope to broaden the scope and applicability of intrinsically interpretable models beyond the simple examples of rule-based and linear models which are typically considered. 


In this work, we present a theoretical framework for both defining AI models and analysing their interpretability. Our approach is based on \emph{category theory}, a mathematical language for describing processes and their \emph{composition}, which has in recent years found a number of applications in ML \cite{fong2019backprop,cruttwell2022categorical,shiebler2021category}. A benefit of the categorical approach is its use of the simple but formal \emph{graphical calculus} which allows the description of processes in terms of intuitive \emph{string diagrams} \cite{selinger2011survey,piedeleu2023introduction}. We show how a wide range of AI models can be described in terms of string diagrams, in order to demonstrate the broad applicability of the categorical approach,  and introduce categories to those with an ML background. Amongst many others we discuss linear and rule-based models, neural networks (NNs), recurrent NNs \cite{elman1990finding} and transformers \cite{vaswani2017attention}, \emph{DisCoCirc} and \emph{DisCoCat} models in natural language processing (NLP) \cite{coecke2010mathematical,coecke2021mathematics,wang2023distilling}, conceptual space models \cite{bolt2019interacting,QonceptsFull2024} and causal models \cite{pearl2009causality,
jacobs2019causal,lorenz2023causal}. These string diagrams generalise several existing graphical approaches, including decision trees, computational graphs for neural networks, DAGs for causal models, and even circuit diagrams in quantum computation, whilst also upgrading these approaches to allow for formal reasoning about the models themselves.

The diagrammatic approach provides a unified perspective in which to both compare AI models, and readily analyse the interpretability of a model in terms of its components. In doing so we see that the explainability of the classic examples of intrinsically interpretable models, namely linear and rule-based models, is evident diagrammatically. Beyond these examples, our analysis highlights the explainability benefits offered by models coming with rich interpretable \emph{compositional} structure, which we term \emph{\CI}\ (CI) models. This encourages us to broaden our perspective beyond the standard examples of intrinsically interpretable models, which form special cases, to more general CI models. Prominent examples of CI models include causal models, which are actively studied in the growing field of \emph{causal ML} \cite{scholkopf2022causality,scholkopf2021toward} and where compositional structure is given by causal structure, and DisCoCirc models in NLP \cite{coecke2021mathematics,wang2023distilling}, where compositional structure derives from grammatical structure.

For most \rlc{black-box models}, the only \rl{manifestly} interpretable components are the inputs and outputs, and thus XAI explanations \rl{often focus on} input-output behaviour.  CI models instead allow one to interpret and reason about their internal components. We make the interpretability benefits of such CI models explicit by showing how they allow for three specific forms of explanation: \rl{\emph{no-influence} arguments} as to their input-output influences; \emph{diagram surgery} arguments which generalise causal interventions; and finally the new notion of \emph{rewrite explanations}, which provide guarantees on, and explanations for, the outputs of compositional models satisfying certain interpretable equations. 

Beyond these arguments, we also use our framework to give precise definitions that clarify and disentangle aspects of common intuitions in XAI. These include: when a model itself is interpretable rather than just affords (approximate) explanations of particular outcomes;  notions of intrinsic interpretability; the distinction between `abstract' (structure-based) and `concrete' (semantics-based) interpretations; and the relation between structure in `the model' and structure in `the world'.  Causal models also form a prime example of interpretable compositional models, and we aim to help clarify the prominent role of causal intuitions in XAI, where for explanatory purposes it is crucial to distinguish between causal structure `just' induced by an NN-based model and that corresponding to the phenomena that the model is about. 

\rl{The central contributions of this work are:
\begin{itemize}
	\item The presentation of AI models as compositional models, providing a formal basis to compare different kinds of model on a common ground and study their interpretability. 
	\item The clarification of certain concepts in XAI using the compositional perspective. 
	\item The notion of \CI\ models, generalising intrinsically interpretable models.
	\item The kind of diagrammatic explanations as facilitated by \CI\ models. 
\end{itemize}}



Another noteworthy benefit of a categorical treatment is the ability to accommodate not only classical deterministic and probabilistic models, but also quantum models such as those based on \emph{quantum machine learning} (QML) and implemented on quantum computers. Indeed, formal diagrammatic reasoning has historically been heavily motivated by the study of quantum processes and information \cite{abramsky2004categorical,coecke2018picturing}, and quantum models are often described compositionally in terms of quantum circuits. Hence a categorical approach such as that proposed here can be seen as necessary in order to analyse interpretability of future quantum AI models.


Whether classical or quantum, the toolkit for analysing AI models and interpretability presented here naturally leads one to consider the explainability benefits of models which make use of rich compositions of interpretable components.  In future work, we hope to further explore the space of such compositional models, including the difficult problem of how and to what extent compositional structure can be learned from data (generalising both \emph{causal representation learning} for causal models \cite{scholkopf2021toward} and learning conceptual domain structure \cite{higgins2016beta,QonceptsFull2024}), as well as further sharpen our explainability techniques such as the notion of rewrite explanations.

\subsection{Overview} \label{sec:overview}

Let us now give a brief overview of the article and our diagrammatic treatment of AI models and interpretability as a whole. Our formalisation is based on category theory, the mathematics of \emph{processes} and their \emph{composition}, which allows us to describe the structure of models at an abstract level using \emph{string diagrams}.

Formally, a category consists of a collection of \emph{objects} $A, B, \dots$ and \emph{morphisms} (processes) between them. For example, consider a model which consists of a process $M$ from a collection of inputs $X_i$ to a collection of outputs $Y_j$. In diagrams, this would be depicted as below, where inputs and outputs (objects) are depicted as wires and the process as a box, read from bottom to top. 
\begin{equation} \label{eq:intro-model}
\input{model-box.tikz} 
\end{equation}
By choosing which category we are working in, we can consider different kinds of processes and thus models. For example, $M$ may describe a deterministic function, such as a neural network, a probability channel as in a Bayesian network, or even a quantum process as used in quantum computation.

In any case, to gain more understanding of the model we will need to `open up' this process further, by decomposing it as a string diagram of internal components, and this is where the categorical language becomes useful. For example, we may know that the process $M$ in fact takes the following form. 
\begin{equation} 
\label{eq:model-decompose}
\input{model-box-decompose-2.tikz}
\end{equation}
This string diagram factorises $M$ in terms of internal `variables' $W$ and $Z$, processes $f$ and $g$, and the state $z$. The diagram itself can be understood as describing the abstract \emph{structure} of the model. This structure is then given specific \emph{semantics} when it is implemented via a specific choice of category, along with specific choices of objects and processes for the wires and boxes. For example, for a semantics based on neural networks, we would assign a dimension to each wire, a vector to $z$, and neural networks to the boxes $f, g$.

We can formalise this view of a model in categorical terms with the notion of a \emph{compositional model} $\modelM$, which consists of the following. Firstly, a collection $\Sig$ of abstract \emph{variables} (wires in the diagram) and \emph{generators} (the boxes), as well as any equations that these may satisfy. From these we can generate an entire \emph{structure category} $\catS$, consisting of all abstract string diagrams which can be built from these components, such as the diagram in \eqref{eq:model-decompose}. Secondly, the model includes a choice of \emph{semantics} category $\catC$, along with a \emph{representation}  $\sem{\syn{V}}$, $\sem{\syn{f}}$ in $\catC$ of each  variable $\syn{V}$ and generator $\syn{f}$, respectively. Formally this is equivalent to specifying a \emph{functor} (mapping between categories): 
\[
\sem{ - } \colon \catS \to \catC
\]
which gives semantics to any diagram built from the variables and generators. This view of a model as a functor between categories of structure (or `syntax') and semantics has a long history in \ir{categorical logic \cite{Lawvere_PhD_Thesis, lawvere1963functorial}}. In practice, we usually define a compositional model implicitly by simply drawing one or several string diagrams which encode the variables and generators, such as that above in \eqref{eq:model-decompose}.

A large number of AI and ML models can be described string diagrammatically as compositional models in this way, and we cover many in this article. In typical cases the model essentially consists of a single input-output process, decomposed as a diagram such as \eqref{eq:model-decompose}. Examples include decision tree models, linear models, neural networks, as well as specific architectures such as the transformer (with fixed input length), and causal models. 

\begin{figure}[H]
	\centering
\begin{subfigure}{3.5cm}
		\centering
\figtikz{tree-intro}
		\caption{\label{Fig_:treeintro}}
	\end{subfigure}
		\begin{subfigure}{3.5cm}
		\centering
\figtikz{NN-pic}
		\caption{\label{Fig_:complevel-nn}}
	\end{subfigure}
	\begin{subfigure}{3.5cm}
		\centering
		\input{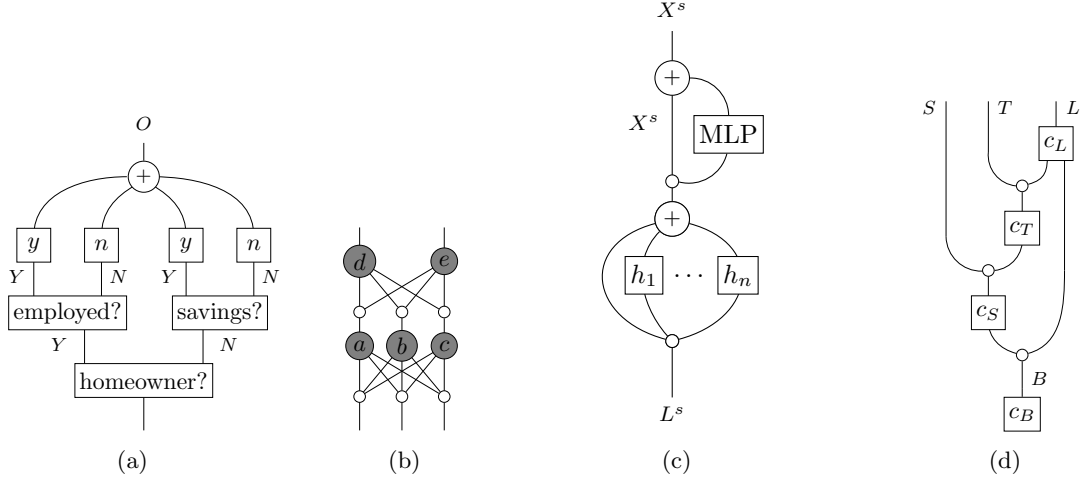}
		\caption{\label{Fig:complevel-transformer}}
	\end{subfigure}
	\begin{subfigure}{5cm}
		\centering
\figtikz{Fig_smoking_example}
		\caption{\label{Fig:replevel-causalmodel}}
	\end{subfigure}
	\caption{String diagrams for (a) decision tree, (b) neural network with layers of size 3, 2, (c) simplified transformer, (d) causal model.}
	 \label{fig:intro-oneprocessmodels}
\end{figure}

String diagrams for these models are shown in Figure \ref{fig:intro-oneprocessmodels}. In each case these diagrams in fact correspond to well-known graphical representations of such models, namely as trees, computational graphs, or DAGs. String diagrams upgrade these graphical systems to also capture semantics and thus the ability to reason about the model. For example, the string diagram of a causal model allows us to both represent it (as for a DAG) and categorically carry out \rl{causal and probabilistic reasoning with the model} \cite{jacobs2019causal,fritz2023d,lorenz2023causal}.

\begin{figure}[H]
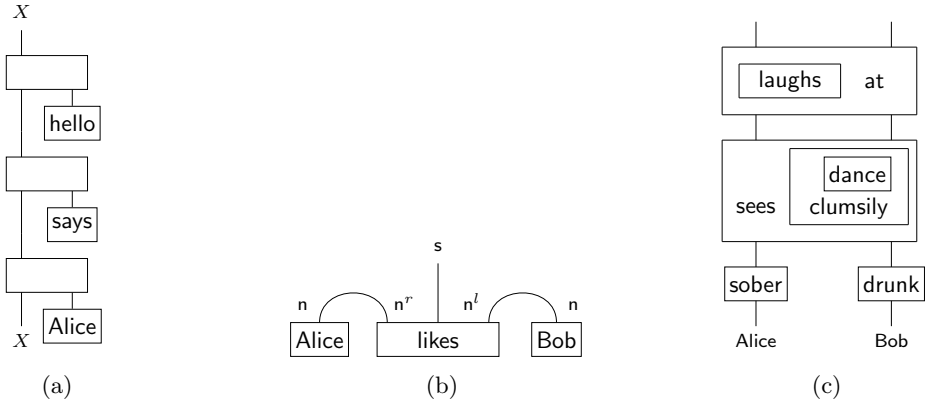
 
	\centering
	\begin{subfigure}{5cm}
		\centering
\figtikz{RNN-intro}
		\caption{\label{Fig:intro-RNN}}
	\end{subfigure}
		\begin{subfigure}{5cm}
		\centering
\figtikz{Alice-likes-Bob}
		\caption{\label{Fig:replevel-dcat}}
	\end{subfigure}
	\begin{subfigure}{5cm}
		\centering
\figtikz{drunkbob3intro}
		\caption{\label{Fig:replevel-dcirc}}
	\end{subfigure}
		\caption{Diagrams for text representations in (a) recurrent neural network (b) DisCoCat model (c) DisCoCirc model.}
	\label{fig:intro-NLPmodels}
\end{figure}

Another common form of model does not distinguish any particular input-output process but instead makes use of a collection of processes and diagrams, formally within the structure category $\catS$. For example, this is common for some structured models of language, where one may have a specific diagram for each input text which captures its representation (semantics) in the model. These include recurrent neural networks (RNNs), bag of words models, or the more categorically oriented DisCoCat and DisCoCirc models. Examples of string diagrams representing text meanings for some of these models are shown in Figure \ref{fig:intro-NLPmodels}.

Now, while a diagrammatic account can provide insights into a model, it doesn't yet tell us that a model is \emph{interpretable}. For example, in the string diagram above for a neural network the individual neurons may have no meaning to us from the outside. Hence any interpretation which exists must be provided in addition to the formal structure of the model. 

Here we will define an \emph{interpretation} of a compositional model semi-formally, as a partial mapping from variables and related processes into a collection  $\Human$ of `human-friendly' terms and concepts, which assigns them meaning. In more detail, an interpretation consists of two parts. Firstly, the \emph{abstract interpretation} assigns meanings in $\Human$ to the abstract structure of the model, namely its variables and generators. For example, we may say that a variable $\syn{V}$ in a model related to images corresponds to the concept `brightness'. Secondly, a \emph{concrete interpretation} assigns meanings to aspects of the semantics, i.e. processes in $\catC$, such as individual states. For example, a concrete interpretation could assert that the state $V = 0$ corresponds to `dark' and $V=1$ to `bright'. Typically, both mappings are only partial, since most processes in $\catC$ will not be interpreted, and neither may some of the variables or generators, such as those which are `latent' or hidden.



Using this setup, we can analyse many examples of AI models compositionally, and discuss their interpretability. For example, while transformer models can be drawn as string diagrams, their internal components (e.g. attention heads) typically lack any prior interpretation. In contrast, models typically considered to be intrinsically interpretable, such as rule-based models, come with concrete interpretations, and moreover the components of their associated string diagrams precisely show the manner in which they are usually deemed to be interpretable. 


This viewpoint suggests that we consider not only the standard examples of `intrinsically interpretable' model, namely linear and rule based models, but more general forms of interpreted compositional models. Moreover, when these models come with rich compositional structure, involving many interpreted processes which may be composed and re-combined in a number of ways, the interpretability benefits become yet more apparent. As already mentioned, we term such models as compositionally-interpretable (CI). Examples of CI models include (interpreted) causal models, as explored in the field of Causal ML \cite{scholkopf2021toward,scholkopf2022causality}, as well as DisCoCirc models in NLP \cite{coecke2021mathematics,wang2023distilling}.

\begin{figure}[H]
	\centering
	\begin{subfigure}{2.5cm}
		\centering
\figtikz{io-intro-pic}
		\caption{\label{Fig:intro-io}}
	\end{subfigure}
	\begin{subfigure}{3cm}
		\centering
\input{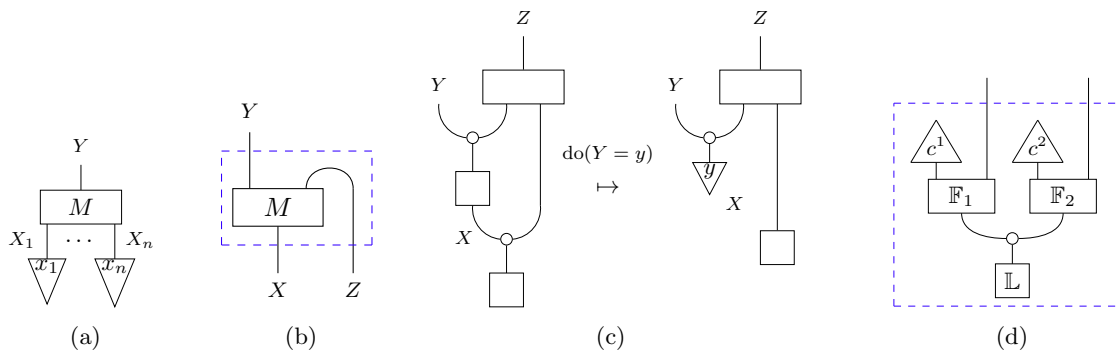}
		\caption{\label{Fig:intro-cond}}
	\end{subfigure}	
	\begin{subfigure}{5cm}
		\centering
\figtikz{int-intro-example}
		\caption{\label{Fig:intro-intervene}}
	\end{subfigure}
	\hspace*{0.6cm}
	\begin{subfigure}{4cm}
		\centering
\figtikz{cf-intro-illustration}
		\caption{\label{Fig:intro-cf}}
	\end{subfigure}
		\caption{Meaningful processes in various frameworks. (a) Applying inputs to a simple input-output model. (b) Conditional probability $P(Y | X, Z)$ in a statistical model. (c) Do-intervention on a causal model. (d) Counterfactual distribution for a functional causal model \ir{$\modelM = \mathbb{F} \circ \mathbb{L}$, where $\mathbb{F}$ and $\mathbb{L}$ denote the deterministic part for the endogenus variables and the product distribution over the exogenous variables, respectively.}}
	\label{fig:intro-frameworks}
\end{figure}

One way to make this notion of `rich' compositional structure precise is what we refer to as the \emph{framework} of a model, which captures the way that the compositional structure of the model is actually used, i.e.~which meaningful string diagrammatic calculations we can perform. For example, the framework of causal models is famously richer than that of plain statistical models, allowing us to compute not only conditional probabilities, but also \emph{interventions} and \emph{counterfactuals} \cite{lorenz2023causal}, as illustrated in Figure \ref{fig:intro-frameworks}. 

In order to make the proposed interpretability benefits of CI models more explicit, we can consider what forms of \emph{explanation} they can provide over generic \rlc{black-box models}. For the latter, typically it is only the inputs and outputs which have a concrete interpretation, and so most post-hoc XAI techniques \rl{view the model simply at the} `outside' level of input-output behaviour. For example, \emph{counterfactual explanations (CFEs)} involve searching for minor alterations $x'$ to an input $x$ for the model $M$ which produce a given output $y'$; i.e.~given that $M \circ x = y$, which is the `nearest' input $x'$ with $M \circ x' = y'$?

\begin{figure}[H]
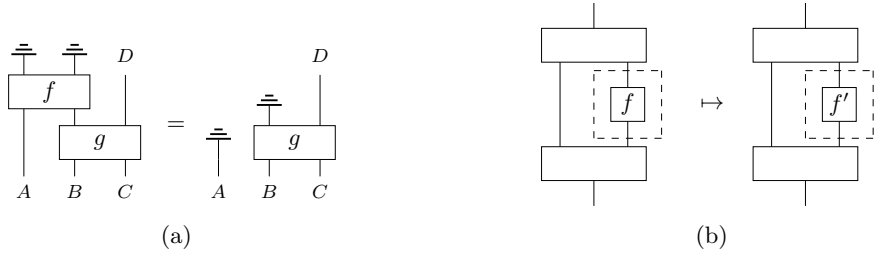

	\centering
	\begin{subfigure}{7cm}
		\centering
\figtikz{sig-intro} 
		\caption{\label{Fig:intro-signalling}}
	\end{subfigure}
	\begin{subfigure}{7cm}
		\centering
\figtikz{diag-surgery-intro}
		\caption{\label{Fig:intro-surgery}}
	\end{subfigure}
		\caption{(a) \rl{Example argument showing that $A$ cannot influence $D$} for a model of the left-hand form. (b) Illustration of diagram surgery in which we replace the component $f$ of a diagram with $f'$.}
	\label{fig:intro-diag-expl}
\end{figure}

In contrast, for a CI model the internal components are themselves interpretable, allowing us to provide richer explanations and constraints on their behaviour. We discuss several forms of explanations available for CI models in Section \ref{sec:expl-from-diags}. 
The most simple \rl{are \emph{(no-)influence arguments},} illustrated in Figure \ref{Fig:intro-signalling}, which allow us to reason about which variables can affect some \rl{particular} outputs of a model. Often these \rl{influence constraints} can be simply read off from the connectivity of the string diagram for a model, when it has non-trivial compositional structure. In contrast, the diagrams for typical neural networks are fully connected, offering no such constraints.

Next, given a string diagram for a model, we can consider altering or acting on it to examine how a given output is affected, or to inspect the model further at certain locations. Such \emph{diagram surgery}, illustrated in Figure \ref{Fig:intro-surgery}, can be seen to generalise interventions on causal models, as well as CFEs, by extending them beyond simply altering inputs to now also altering the internal processes of a model.

\begin{figure}[H]
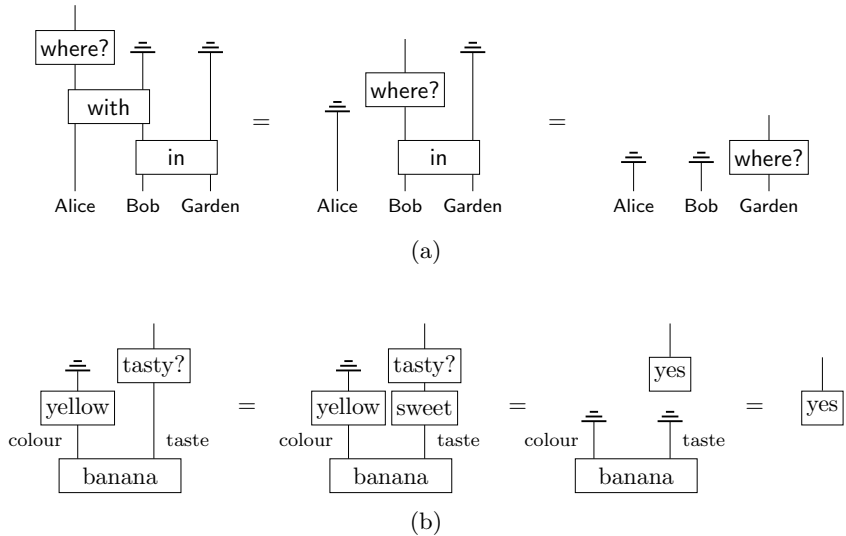

	\centering
	\begin{subfigure}{\linewidth}
		\centering
\figtikz{rewrite-illustration-intro}
		\caption{\label{Fig:rewrite-expl-disco}}
	\end{subfigure}
	\\ \vspace{2em}
	\centering
\begin{subfigure}{\linewidth}
		\centering 
\figtikz{banana-rewrite-ex}
		\caption{\label{Fig:rewrite-expl-banana}}
	\end{subfigure}
\caption{Toy examples of rewrite explanations, in which the equations used are implicit in the rewrite steps. 
(a) A DisCoCirc type model, where we explain why \dc{Alice is with Bob in the Garden, where is Alice?} returns as its answer the location of the garden. 
\ir{The equations used in the rewriting express that if \dc{X is in/with Y} then the answer to \dc{Where is X} is simply \dc{Y}.}  
(b) A conceptual space type model, using information that yellow bananas are typically sweet to explain why they are output as tasty.  
\ir{The equation implicit in the first rewrite captures that a yellow banana is also sweet; in the second it states that sweetness on its own ensures tastiness.}
\label{fig:rewrite-expl} }
\end{figure}

Finally, the strongest constraints are provided by a novel form of explanation we call \emph{rewrite explanations}, applicable only to models with interpretable compositional structure. These require knowledge about a model in the form of diagrammatic equations, which may either be discovered empirically after training, encouraged in training via a loss function, or imposed in the definition of a model. From such equations, a rewrite explanation consists of a diagrammatic argument proving that a given diagram, such as that describing the model applied to some input, is (approximately) equal to another, such as a given output value. Examples of rewrite explanations are shown in Figure \ref{fig:rewrite-expl}. Crucially, for these to qualify as explanations, the processes featured in the argument must themselves be interpretable. Because of this, and the need for non-trivial diagrammatic structure, typical AI models -- unlike CI models -- will not allow for any non-trivial rewrite explanations of their behaviour. 


Overall, we hope that the compositional perspective can help to elucidate the interpretability of a wide range of AI models, whilst enabling the definition of new inherently categorical forms of model coming with explainability benefits. The compositional viewpoint can also help to clarify various notions from across XAI, 
\st{including the role of causal concepts, since causal models are a special case of compositional models (see Section \ref{sec:CI-and-causal-XAI}).}

\paragraph{Quantum models.}
A strength of the categorical view is the ability to accommodate both classical and quantum AI models uniformly, by choice of the semantics category. In this way the same abstract structure, such as that of a DisCoCirc NLP model illustrated below, may be implemented either classically in terms of neural networks (left), or as quantum circuits (right). 

\begin{figure}[H]
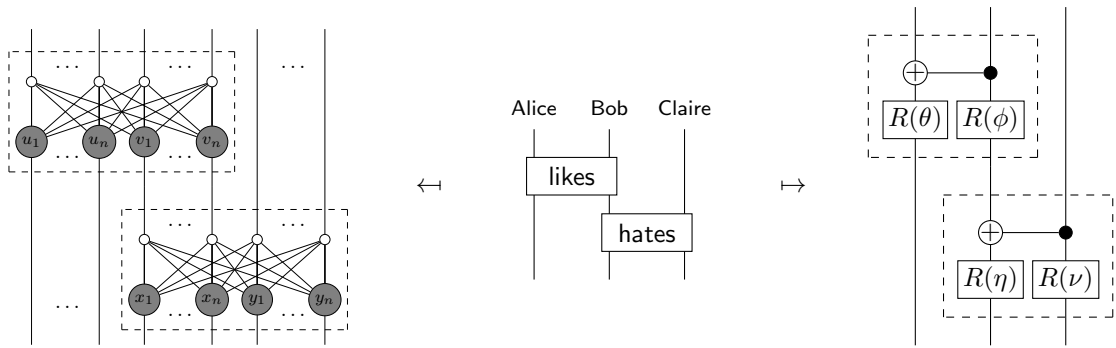

	\centering
\[
\scalebox{0.8}{\input{dcircneural3.tikz}}
\qquad
\mapsfrom
\qquad
\input{dcirclevel3.tikz}
\qquad 
\mapsto
\qquad
\input{dcircquantum.tikz}
\]
\caption{Neural (left) and quantum (right) implementations of a text circuit (centre).}
\label{fig:disco-circ-quant-nn}
\end{figure}
In Section \ref{sec:quantum} we discuss aspects of interpretability related specifically to quantum AI models. In fact, the compositional perspective has been relatively widely adopted in quantum information, and many developments in applied category theory have their origins in the study of quantum processes. In addition, quantum AI models are often defined compositionally, in terms of circuit diagrams. Hence we claim that a compositional framework capable of accommodating both classical and quantum models uniformly, such as that offered here, will be essential to assess the interpretability of quantum models. 
While features such as entangled states may make it more challenging to assign specific concrete interpretations to all of the states of a quantum model, the remainder of our compositional approach to interpretability, including all of the diagrammatic forms of explanation outlined above, applies just as straightforwardly to quantum models as to classical.

\paragraph{Structure of the article.}
We begin in Section \ref{sec:context} by giving as context for this work a short summary of the current state of XAI, and listing further related research. Following this, in Section \ref{sec:setup} we introduce the necessary mathematical background, introducing categories and string diagrams. In Section \ref{sec:compositional-models} we give our central definitions, namely that of a compositional model, and an interpretation of a model. To convey these abstract definitions, in Section \ref{sec:examples-compositional-models} we describe a comprehensive list of AI models as compositional models, and discuss their interpretability, including decision trees, neural networks, transformers, recurrent neural networks, variational autoencoders, causal models, and DisCo-style models in NLP.  In Section \ref{sec:frameworks} we then introduce the further theoretical notion of a compositional framework, which can articulate the use of models with `rich' and `meaningful' compositional structure. We make use of our compositional perspective to discuss various issues \rl{in conventional XAI} in Section \ref{sec:CI-and-causal-XAI}, including the model-vs-world distinction, comparing CFEs and Pearlian counterfactuals, and causal abstraction. We then take stock of all we have covered and make our observations on the relation between compositionality and interpretability in Section \ref{sec:comp-and-interp}. In Section \ref{sec:expl-from-diags} we sharpen these observations by giving explicit forms of explanations which can be offered by interpretable compositional models, based on signalling, diagram surgery, and the notion of rewrite explanations. In Section \ref{sec:quantum} we discuss aspects of interpretability related specifically to quantum AI models. Finally, in Section \ref{sec:future} we discuss future directions for research, including the exploration of further forms of compositional model and the key question of how compositional structure can be learned from data.

\section{The context} \label{sec:context}

In this section we briefly review the background context of XAI, its issues, and further related research.

\subsection{Conventional XAI} \label{sec:context-XAI}

\subsubsection{A rough overview} \label{subsec:XAI_in_nutshell}

The field of XAI is concerned with the \ety\ and \ity\ of ML models. 
What exactly these two terms are supposed to refer to and how they relate to each other varies, but roughly speaking, XAI studies what can be said about a model in human intelligible terms so as to make understandable how and why that model works and gives particular outputs. 
Typical questions to introduce and illustrate the field are  
``How does the model make predictions?"  \cite{GrazianiEtAl_2023_GlobalTaxonomyInterpretableAI},  
``For situation X, why was the outcome Y and not Z?" \cite{GoyalEtAl_2019_CFVisualExplanations} and  
``Was it a specific feature that caused the decision made by the model?" \cite{MoraffahEtAl_2020_CausalInterpretabilitySurvey}. 

The reasons why we would like to answer such and related questions, and strive for better \ety\ of ML models, can be as broad as: 
justifying a model or its predictions, e.g. to avoid unfair bias and build trust;  
controlling and debugging a model;  
improving a model in terms of accuracy through better \ety; and  
discovering scientific knowledge through the model \cite{CarloniEtAl_2023_RoleOfCausalityInXAI, MolnarEtAl_2020_InterpretableMLHistory}.
Arguably, the first reason drives most of the efforts in XAI, especially with a view to so called \textit{high-stakes decision problems}, where lives are directly impacted by the model's output.
High-stakes areas include the healthcare sector (e.g.~medical image analysis for diagnosis and treatment recommendations), 
the financial sector (e.g.~automated loan approval and investment advice),  
the legal sector (e.g.~bail decisions),  
and the educational sector (e.g.~assessing university applications) \cite{HarradonEtAl_2018CausalLearning, Freiesleben_2021_IntriguingRelation}.


The lack of \ity\ of a model can be \irs{due} to a lack of knowledge of the identity of the overall input-output function, be it because it is proprietary or because of some other epistemic restriction on the model parameters. We can refer to this as the \textit{black-boxness} of the model. Alternatively, the identity of the model may be known, but it may lack \ity\ because it is, loosely speaking, too complex or obscure to be `made sense of'. We can refer to this instead as the \textit{opacity} of a model; that is, a white-box, the opposite of a \rlc{black box}, may still be opaque. 



The XAI literature is large, somewhat piecemeal and in most parts not guided by an agreed-upon conceptual foundation with clear definitions (see below).  
The following intends to only touch on some of the main themes in XAI for the unfamiliar reader, who can otherwise refer to the \rlb{the many existing review articles for a more detailed background; see for instance \cite{GrazianiEtAl_2023_GlobalTaxonomyInterpretableAI, PalacioEtAl_2021_XAIHandbook, MolnarEtAl_2020_InterpretableMLHistory, arrieta2020explainable, DasEtAl_2020_OpportunitiesAndChallengesXAI, DoshiEtAl_2017_TowardsRigorousIML}.}

\paragraph{Post-hoc vs intrinsic.}


A main distinction is that between post-hoc and intrinsic (or ante-hoc) \ity. 
Some models are regarded as manifestly interpretable due to the properties they have \emph{by design} and are thus referred to as \textit{intrinsically \ible\ models}. A precise definition of the concept is never given \ir{in the literature}, but the canonical set of examples include linear (regression) models and rule-based models (also called if-then or logical models) such as decision lists, decision trees, and scoring systems. 
Any of these qualify as intrinsically \ible\ only subject to size conditions like sparsity. 
The intuitions behind intrinsic \ity\ will be discussed in more detail in Secs.~\ref{sec:examples-compositional-models} and \ref{sec:comp-and-interp}. 
In contrast, \textit{post-hoc \ity}\ is the kind achieved through XAI methods applied to a given model after training, when it is opaque and lacks any intrinsic \ity. Post-hoc methods are usually referred to as producing `explanations' of the model of some sort and the vast majority of the XAI literature is concerned with the toolbox of such methods. 

\paragraph{Typical post-hoc methods and dimensions of distinction.}

Post-hoc methods are commonly compared and classified with respect to various dimensions. 
Explanations are either \textit{local or global}, where `local' means that the explanandum concerns a particular prediction, i.e. input-output pair -- like `why was person $X$ predicted to default on the loan?' -- and `global' tends to refer to some form of averaging or combining of local explanations, which is then considered to be explanatory of the model and its workings as a whole. 
A method may be \textit{model-agnostic}, in which case it can be applied to a \rlc{black box}, i.e.~it only relies on access to the input-output behaviour; or the method may be \textit{model-specific}, i.e. rely on facts about the given model, such as the parameters to compute gradients for back-propagation. 
Molnar \cite{MolnarEtAl_2020_InterpretableMLHistory} moreover distinguishes methods in three (not mutually exclusive) ways, by whether they: 
\textit{analyse the components of the model} (model-specific as reliant on a useful decomposition of the model), 
\textit{study the sensitivity of the model} (investigating the input-output behaviour),  
or \textit{use a surrogate model} (an intrinsically \ible\ model trained to approximate the given black-box or opaque model in a specific local input region). 

The majority of methods study the sensitivity of the model in some form or another. 
For instance, a counterfactual explanation (CFE) of a given input-output pair $(x,y)$, an idea first introduced in \cite{WachterEtAl_2017_CounterfactualExplanations}, may be roughly paraphrased as an alternative input $x'$, minimally different from $x$ and such that the model produces some (usually desired, but in any case predefined) output $y'\neq y$. 
Note that CFEs are in general highly non-unique, and see \cite{Freiesleben_2021_IntriguingRelation} for other pitfalls and a generally insightful analysis of CFEs, as well as Sec.~\ref{subsec:CFEs-vs-CFs} for further details.

Many of the other, non-CFE techniques fall into the broad category of \emph{feature attribution methods}. 
The features may be pixels, tabular data or words, and the basic idea of feature attribution is that it `explains' a particular output by assigning each input feature a number that is interpreted as a weight or contribution towards the output (negatively or positively). 
Prominent examples are LIME and SHAP. 
Local Interpretable Model-agnostic Explanations (LIME), first introduced in \cite{RibeiroEtAL_2016_WhyShouldITrustYou}, is a local method that fits a linear regression model -- as an intrinsically interpretable surrogate model -- in feature space within the vicinity of the concrete data point to be `explained'. The output is the list of coefficients, understood as approximate feature importance weights.
Shapley Additive exPlanations (SHAP), first introduced in \cite{LundbergEtAl_2017_SHAP_Paper_UnifiedApproachToInterpreting}, is a local method that also fits a surrogate model, though such that its coefficients (locally) approximate analogues of game-theoretic Shapley values, again interpreted as the different features' contributions to a given outcome.

\paragraph{Terminology, desiderata and definitions.} 

The basic intuitions, and what exactly \ity\ and \ety\ mean, differ significantly across XAI. For instance, ``is it an approximation of a complex model", ``an assignment of causal responsibility", ``a set of human intelligible features contributing to a model’s prediction", or ``a function mapping a less interpretable object into a more interpretable one"? \cite{PalacioEtAl_2021_XAIHandbook}. 
Is it about ``means to engender trust [...], faith in a model’s performance, [...] or a low-level mechanistic understanding of our models"? \cite{Lipton_2018_MythosModelInterpretability}. 
In light of this, review articles have collated and compared different definitions (see, e.g.~\cite{GrazianiEtAl_2023_GlobalTaxonomyInterpretableAI, PalacioEtAl_2021_XAIHandbook, DasEtAl_2020_OpportunitiesAndChallengesXAI})  
with some attempting to provide an overarching definition to `capture them all' and to give guidance.
Seeing as the field is broad, it is perhaps not surprising that one then ends up with rather vague and high-level suggestions. 
For instance, in~\cite{GrazianiEtAl_2023_GlobalTaxonomyInterpretableAI} ``an AI system is interpretable if it is possible to translate its working principles and outcomes in human-understandable language without affecting the validity of the system". 
In~\cite{PalacioEtAl_2021_XAIHandbook} one finds ``an explanation is the process of describing one or more facts, such that it facilitates the understanding of aspects related to said facts (by a human consumer)" and that an ``interpretation is the assignment of meaning (to an explanation)", so that ``for ML, the assigned meaning refers to notions of the high-level task for which the explanans is provided as evidence. An interpretation is therefore bridging the gap between under-specified non-functional requirements of the original task and its representation in formal, low-level primitives." 

Certainly, it is important to be explicit about what the explanandum, the explanans and the explanatory link between the two is, as voiced for example in~\cite{FreieslebenEtAl_2023_DearXAICommunity, PalacioEtAl_2021_XAIHandbook, Creel_2020_TransparencyInComplexCOmputationalSystems}. 
Rather than searching for a distinguished, overarching definition though, one may also attempt to disentangle and clearly name different aspects of `interpretability'. 
There will not be one unique way to do this, but for instance Lipton \cite{Lipton_2018_MythosModelInterpretability} distinguishes :  
(1) \textit{simulatability} that focuses on a simplicity aspect, which calls ``a model transparent if a person can contemplate the entire model at once";  
(2) \textit{decomposability} that demands that ``each part of the model -- each input, parameter, and calculation -- admits an intuitive explanation"; and 
(3) \textit{algorithmic transparency}, which could be given by e.g. proven guarantees that training converges to a unique solution.  
This work's focus is closest to the second.  

An aspect which has attracted considerable attention is the role of the `human element' in XAI. This includes the common attitude of admitting some degree of \textit{subjectivity in desiderata for satisfying explanations} among (non-expert) humans (see, e.g., \cite{GrazianiEtAl_2023_GlobalTaxonomyInterpretableAI, MillerEtAl_2017_ExplainableAI_InmatesRunningAsylum}), but also includes calls to turn to the social sciences for theories of explanation (see, e.g., \cite{MillerEtAl_2017_ExplainableAI_InmatesRunningAsylum, Sullivan_2022_InductiveRisk}). Note the call is not for scientific explanations, but of how humans generally give explanations to each other, and proposals to explicitly model the human in an \ity\ framework \cite{MarconatoEtAl_2023_InterpretabilityInMIndOfBeholder}.

\paragraph{Role of causality in XAI.}

Causality plays a prominent and increasingly explicit role in XAI, though one that is multifaceted and that can cause confusion. 
See \cite{CarloniEtAl_2023_RoleOfCausalityInXAI} for a review covering some of its aspects. 

As is well-known, many ML models can be seen to (implicitly) induce a causal model. 
The details of how exactly and what kind of causal model we leave to Sec.~\ref{sec:CI-and-causal-XAI}, but this observation gives context to how many post-hoc methods are in fact ways to uncover causal properties of the model and to make causal statements about it. 
Obviously, CFEs have a decidedly causal flavour (though as we argue in Sec.~\ref{subsec:CFEs-vs-CFs}, unlike often claimed, a CFE is not a counterfactual in the Pearlian causal model sense). 
Also, feature attribution methods are essentially methods which vary the input features in order to study the (causal) effect on outputs.
Some have indeed argued that ideal explanations in XAI are in fact causal explanations. 
``Explanations can be established by pointing to associations between the explanans and the explanandum [...]. Usually, however, the relationship we are interested in is causal, that is, the explanans makes a difference for the explanandum" \cite{FreieslebenEtAl_2023_DearXAICommunity}; also see \cite{HarradonEtAl_2018CausalLearning, CarloniEtAl_2023_RoleOfCausalityInXAI}.
A crucial but easily overlooked distinction in the context of striving for causal explanations in XAI is the distinction between causal structure of `just' the model vs that of the phenomenon in the world that the model is about. 
This point, which has important consequences for the purposes of interpretability, has been raised by many authors \cite{Beckers_2022_CausalExplanationsAndXAI, FreieslebenEtAl_2023_DearXAICommunity, FreieslebenEtAl_2022ScientificInferenceAndIML, MolnarEtAl_2020_InterpretableMLHistory, Sullivan_2022_UnderstandingFromMLModels, Watson_2022_ConceptualChallengesForIML, CarloniEtAl_2023_RoleOfCausalityInXAI} and will be discussed in more detail in Section~\ref{subsec:model-vs-world}.

\subsubsection{The issues}  \label{subsec:Issues_with_XAI}


Despite the undoubtedly many useful methods and insights that the field of XAI has produced, there also are serious issues to point out -- beyond just the fact that the field appears to lack a conceptually well-grounded framework and a consensus on the definition of \ity. 
Two particularly clear and insightful pieces in this regard are the ones by Rudin \cite{Rudin_2019_StopExplainingBlackBoxes} 
and Freiesleben \etal\ \cite{FreieslebenEtAl_2023_DearXAICommunity}, to which we refer readers for the details and a more comprehensive list of references, while here we only briefly mention those issues that are most pertinent to motivating this work's approach; see also~\cite{Lipton_2018_MythosModelInterpretability, Watson_2022_ConceptualChallengesForIML, MillerEtAl_2017_ExplainableAI_InmatesRunningAsylum, MolnarEtAl_2020_GeneralPitfallsForIML, Creel_2020_TransparencyInComplexCOmputationalSystems, Sullivan_2022_InductiveRisk, DoshiEtAl_2017_TowardsRigorousIML}.

\paragraph{Arbitrariness and falsehood of XAI explanations.}
Many authors have pointed out that post-hoc explanations may be untrustworthy, due to two related facts. Firstly, they are based on approximations to the original model which are inherently `unfaithful', and secondly they are often non-unique and might even seem adjustable at will \cite{Rudin_2019_StopExplainingBlackBoxes, GrazianiEtAl_2023_GlobalTaxonomyInterpretableAI, FreieslebenEtAl_2023_DearXAICommunity}.  
Rudin argues that ``[Post-hoc] explanations must be wrong. [...] If the explanation was completely faithful to what the original model computes, the explanation would equal the original model"  \cite{Rudin_2019_StopExplainingBlackBoxes}, noting that this would do away with the need for the original model in the first place, while an explainer who is correct only 90\% of the time would be wrong in 10\% of cases, which is unacceptable in certain high-stakes situations.

\paragraph{The myth of learning human concepts.}
The belief that deep neural networks must automatically learn the concepts which humans use to reason about the data and task, Freiesleben \etal\ argue, underlies many explanation techniques, but is one of which we should be sceptical.  
Indeed, why should a model learn human concepts when just optimised on a particular task like classification or sequence prediction? 
The underlying assumption of course is that the good performance that a model achieves on some task is only conceivable if it indeed uses our concepts and hence all one needs to do is 
``use XAI techniques like activation maximization or network dissection to discover/reveal which nodes in the network stand for which concept, and then – tada – we have a fully transparent model where every part of the model stands for something"  \cite{FreieslebenEtAl_2023_DearXAICommunity}. 
Crucially, they argue that while certain nodes in the network may co-activate when certain concepts are present (at least to some degree), they lack the causal role that the corresponding concept would have for us, and moreover point out that learning techniques such as drop-out would actually incentivise models to \textit{not} learn concepts as localised in specific neurons. 

\paragraph{The what vs the how.} 
Rudin points out that typical methods like saliency maps may tell us \textit{what} was relevant, but not \textit{how} it was actually used by the model. 
``Knowing where the network is looking within the image does not tell the user what it is doing with that part of the image [...]. In fact, the saliency maps for multiple classes could be essentially the same; in that case, the explanation for why the image might contain a Siberian husky would be the same as the explanation for why the image might contain a transverse flute." \cite{Rudin_2019_StopExplainingBlackBoxes}

\paragraph{The inadequacy of probing a model.}  
Perturbing data to then study the model's outputs on these additional inputs is very common in XAI, but involves applying the model to inputs outside the given data distribution (that is, outside the training and test data). 
Yet, guarantees for a model to correctly generalise to an out-of-distribution setting are notoriously hard to come by \cite{scholkopf2021toward}. 
Conclusions about how a perturbation-based method apparently makes the model more transparent should be seen critically. 
Put yet more strongly: ``In extrapolation regions, models disagree even when fitted to exactly the same data and achieve similar high performance on a test set. Asking an ML model to extrapolate is like asking a five-year-old kid who hasn’t gone to school about her insights into algebraic topology. You might get an answer, but that answer will not really help you." \cite{FreieslebenEtAl_2023_DearXAICommunity}

\paragraph{The accuracy-transparency trade-off, or myth thereof.} 
A prevalent attitude consists in the belief that `perfect' accuracy and \ity\ are mutually exclusive and that one generally has to trade off one for the other to some degree. 
Rudin debunks this belief as a myth that has had much influence on the types of models considered and research undertaken \cite{Rudin_2019_StopExplainingBlackBoxes}. 
While opaque models such as ones built from deep NNs may perform with higher accuracy on a larger class of tasks and domains, this does not imply the impossibility of vanilla transparency for an equally accurate model if one is willing to `pay the cost' of finding models tailored to the \textit{specific problem at hand} -- costs being time, effort and possibly expert knowledge. 
Rudin finds that ``particularly when the data are structured, with a good representation in terms of naturally meaningful features" there is little difference in performance between complex models and classic examples of intrinsically interpretable models. 
Crucially, many of the high-stakes decision problems, where \ity\ arguably matters the most, are in fact such data science problems. Rudin lists a series of concrete examples, where \ible\ models with state-of-the-art accuracy do exist.

\subsection{Compositionality} \label{subsec:compositionality-context} 


The term `compositionality' has appeared in various fields, but often has a distinct meaning and relates to distinct questions in each setting. Here we give a brief summary of various uses of the term, to clarify what compositionality as understood in this work does and does not refer to.


\paragraph{In linguistics and philosophy.}
There is a long tradition of studying notions of compositionality in linguistics and philosophy, including formal treatments as pioneered by Frege \cite{frege1892sinn, Frege_1914_LetterToJourdain} and Montague \cite{montague1970universal}. 
A focus lies on the \emph{principle of compositionality}, which roughly states that 
``the meaning of a complex expression is determined by its structure and the meanings of its constituents" \cite{sep-compositionality}.
This principle aims to explain phenomena such as the productivity and systematicity of language, which allow us to understand sequences of words we have never encountered before. 
An important question concerns the interplay between this `bottom-up' view of meaning determination and the contrasting `top-down' view expressed by the so-called \emph{context principle}.\footnote{Note that principles of compositionality are often attributed to Frege incorrectly \cite{nefdt2020puzzle, janssen2001frege, szabo2014problems, pelletier2001did}.}  
See \cite{sep-compositionality} and \cite{werning2012oxford} for encyclopaedic overviews of the field.

\paragraph{In cognitive science.}
\rlc{The \emph{language of thought hypothesis} (LOTH) is the proposal that our thought processes happen in a mental language, Mentalese, and plays a prominent role in cognitive science, going back to the influential works of Fodor \cite{fodor1975language, fodor1983modularity, fodor1988connectionism} (in turn building on a long history in philosophy). 
Natural language and mental language differ in multiple ways; crucially, however, they also resemble each other in important ways according to LOTH. In particular, Mentalese features sentences and words with the meanings of sentences being built from word meanings in a systematic way. 
\rlc{Indeed, Fodor argued that Mentalese is a compositional language which explains the compositional nature of thought processes.}
Assuming that mental events that represent mental (intentional) states are instantiated neurologically, one question to study, for instance, is how the compositional structure of the state is reflected in the instantiation -- are there individuated instantiations of the separately meaningful components of a mental state, or thought process? 
See \cite{sep-language-thought} for an overview.}

\paragraph{In NLP and AI more generally.}
Assuming that there indeed is a principle of compositionality behind human language use, and more generally behind human reasoning, there are questions concerning AI which have attracted much attention, especially in light of recent successes of \rlc{deep NN-based} NLP models. How is it that models generalise so well, despite not being systematic or rule-based systems?
And can they genuinely \emph{generalise compositionally}? \cite{baroni2020linguistic, nefdt2020puzzle, lake2018generalization}
Some have argued that part of the difficulty in answering such questions is the lack of clarity on what `generalising compositionally' should mean for modern, connectionist NLP models. This has led to a focus on the identification of datasets and \emph{compositional tasks} that are designed to test a model's capability to generalise compositionally, based on the conviction that such a task requires a compositional approach to reach the correct output \cite{socher2013recursive, hupkes2020compositionality, dziri2024faith}.
Broadly speaking, we can refer to the engagement of AI research with compositionality in this sense -- studying model performance on compositional tasks -- as the study of the \emph{compositional behaviour} of models. This is conceptually distinct from questions of the `modularity' of the model, i.e. the compositional structure of the model at an architectural or implementation level, which is closer to our focus here. However, some works have explored the design of a model's structure with a view to improving its compositional behaviour \cite{socher2013recursive, yessenalina2011compositional, csordas2021are, QDisCoCirc}. 
For an overview on the `challenge of compositionality in AI' also see \cite{Gary_website_ChallengeOfCompositionality}.

\paragraph{In category theory.} 
Finally, our use of the term `compositionality' is rooted in category theory, and especially the tradition of the \emph{Applied Category Theory} (ACT) community. It refers to the formal study of compositional structure, that is, how systems and processes compose and how structural constraints for well-behaved composition can guide abstract reasoning about them. This aproach can be applied in many contexts including theoretical computer science, fundamental physics, quantum computing, AI, and engineering. See \cite{selinger2011survey, fong2018seven, piedeleu2023introduction, coecke2021compositionality, baez2011physics} for some introductory pieces, Section~\ref{subsec:related-work} for the works most closely related to ours, and Section~\ref{sec:setup} for the formal details.

\subsection{Related work} \label{subsec:related-work} 

Alongside the articles introducing the many AI models referenced throughout the text, there is a wide range of work related to our approach. 

\paragraph{Compositional Intelligence.} Broadly speaking, this work can be seen as part of the research programme of \emph{Compositional Intelligence}, which explores applications of compositional models in (quantum) AI, and theories of intelligence more broadly. In particular, this work aims to make precise the common intuition that categorically structured models come with interpretability benefits over \rlc{black-box models}. Another aim is to build on the proposed definitions of `meaningful' compositional structure put forward by Coecke \cite{coecke2021compositionality}. 

Aside from interpretability, one could argue for further benefits of such models as a desired `middle ground' between `too strict' symbolic models and `too loose' connectionist models, similarly to the aims of neuro-symbolic AI \cite{hitzler2022neuro}. In our case, symbolic aspects appear through explicit diagrammatic structure, and connectionist aspects through the implementation. These benefits could include robust reasoning and increased efficiency from training on smaller representational spaces, with the latter perhaps being beneficial for QML \cite{mcclean2018barren}.  

Rodatz et al.~\cite{PatternLanguage} develop an approach to defining and training ML models similarly to compositional models in our sense, in terms of generators whose behaviour is specified by string diagrammatic equations which constrain loss functions. These would also allow for the rewrite explanations in our sense, and we hope to connect with this approach in future work. Also related is a recent quantum implementation of the DisCoCirc framework in \cite{QDisCoCirc}, in which the authors discuss the compositional interpretability of the trained model, making use of diagrammatic axioms similar to those considered for our rewrite explanations.

\paragraph{Compositional models.}
Our notion of a compositional model is the standard one in \ir{categorical logic \cite{Lawvere_PhD_Thesis, lawvere1963functorial}}, and is used for example in DisCoCat \cite{coecke2010mathematical}, and categorical treatments of causal models \cite{fong2013causal,jacobs2019causal,lorenz2023causal}. For those wanting a more formal treatment, we recommend Giovanni de Felice's DPhil thesis \cite{de2022categorical}, which describes a number of the compositional models explored here, with a focus on NLP.

The \texttt{discopy} python library may be seen as a software implementation of compositional models \cite{de2020discopy}, allowing for a range of semantics, such as neural networks or quantum circuits. The related \texttt{lambeq} toolkit \cite{kartsaklis2021lambeq} handles the special case of quantum and tensor-network-based compositional models in NLP, discussed in Sections  \ref{subsec:NLP-sequence-model}, \ref{subsec:DisCoCat} and \ref{subsec:DisCoCirc}. 

While we focus on `strictly' compositional (i.e. strong monoidal) models, recent work has explored graded notions based on \emph{obstructions} to compositionality, and related properties such as emergence \cite{puca2023obstructions, newerObstrPaper} and it would be interesting to explore connections to this area in future work.

\paragraph{Categories for AI.} 
The categorical perspective on ML has been developed in recent years by a number of authors, usually with a focus on viewing the training setups of models as `bidirectional transformations', or \emph{lenses}. These capture the forward pass of a trained model along with its backward `parameter updating' direction; see for example \cite{fong2019backprop,shiebler2021category}. Parameterised lenses have been used to formalise the general setup of gradient-based learning in \cite{cruttwell2022categorical}, and this is applied to models which produce `explanations' for their outputs in \cite{barbiero2023categorical}. 

Closely related is the field of categorical cybernetics \cite{capucci2021towards,smithe2021cyber} which observes that bidirectional transformations also capture probabilistic inference. This has even been used to make new proposals for the general forms of AI systems in \cite{swan2022road}. Here we focus on the (string diagrammatic) structure of a model as given, rather than the training process. Related ideas within the field of human cognition are theories of the \emph{Bayesian brain} such as \emph{Predictive Processing} and \emph{Active inference} \cite{parr2022active,smith2022step}, with the latter having received various recent treatments in terms of category theory \cite{smithe2021compositional,smithe2022compositional} and a related string diagrammatic treatment \cite{tull2023active}. 

Our account of causal models is based on the flourishing area of string diagrammatic approaches to probability theory \cite{cho2019disintegration,fritz2020synthetic}, and causality \cite{fong2013causal,jacobs2019causal,fritz2023d}, and in particular the string diagrammatic account of the Pearlian causal model framework in \cite{lorenz2023causal}. 

String diagrammatic formalisms for neural networks, and architectures such as transformers, have been presented informally and semi-formally \cite{elhage2021mathematical} while the upcoming article \cite{AnatomyOfAttention} aims to give a systematic diagrammatic view on a wide range of \st{attention-based architectures}. Another relatively structurally focused field of ML lies in \emph{geometric deep learning} \cite{bronstein2021geometric}, which has recently been given a categorical generalisation \cite{gavranovic2024categorical}, and which we hope to connect to in future work.

\section{Categories and string diagrams} \label{sec:setup}

To describe compositionality formally, we will make use of the mathematics of \emph{category theory} and its associated graphical language of \emph{string diagrams} \cite{coecke2006introducing,piedeleu2023introduction,selinger2011survey}. \stb{These diagrams provide a rigorous reasoning system which is often equivalent to working with traditional equations, but is more intuitive, allowing one to skip many mathematical details which are automatically accounted for by graphical rules.} 

A \emph{category} $\catC$ consists of a collection of \emph{objects} $A, B, \dots$ and \emph{morphisms} $f \colon A \to B$ between them, which we can compose in sequence. Such a morphism is also referred to as a \emph{process} $f$ from input $A$ to output $B$. We write $\catC(A,B)$ for the collection of morphisms from $A$ to $B$ in $\catC$. In string diagrams an object $A$ is depicted as a wire labelled by $A$, and a morphism $f \colon A \to B$ as a box with lower input wire $A$ and upper output wire $B$, read from bottom to top.
\[
\begin{tikzpicture}
	\begin{pgfonlayer}{nodelayer}
		\node [style=map] (0) at (0, 0) {$f$};
		\node [style=none] (1) at (0, 1) {};
		\node [style=none] (2) at (0, -1) {};
		\node [style=label] (3) at (0, 1.5) {$B$};
		\node [style=label] (4) at (0, -1.5) {$A$};
	\end{pgfonlayer}
	\begin{pgfonlayer}{edgelayer}
		\draw (2.center) to (1.center);
	\end{pgfonlayer}
\end{tikzpicture}
 
\]
For any morphisms $f \colon A \to B$ and $g \colon B \to C$ whose respective outputs and inputs match, we can form the sequential composite $g \circ f \colon A \to C$, depicted as follows.
\[
\input{composite-1.tikz}
\]
Each object $A$ also comes with an \emph{identity} morphism $\id{A} \colon A \to A$ depicted as a plain wire:
\[
\input{identity.tikz}
\]
The identity acts as a unit for composition, with $\id{B} \circ f = f = f \circ \id{A}$ for any morphism $f \colon A \to B$.

We will work with categories with the following extra structure, allowing us to also compose processes `side by side'. In formal terms, a \emph{symmetric monoidal category (SMC)} is a category $\catC$ coming with a functor $\otimes \colon \catC \times \catC \to \catC$, distinguished object $I$ and natural transformations which express that $\otimes$ is suitably associative and symmetric, with $I$ as a unit \cite{coecke2006introducing}. Spelling out what this means is most naturally done in string diagrams, as follows. 

Firstly, for any pair of objects $A, B$ we can form their parallel composite or \emph{tensor} $A \otimes B$. In diagrams (the identity morphism on) $A \otimes B$ appears as a wire labelled by $A$ alongside one labelled by $B$. 
\[
\input{tensor-ob.tikz}
\]
This allows us to have boxes with multiple inputs and outputs in diagrams. 
For example, a morphism $ f \colon A \otimes B \otimes C \to D \otimes E$ is drawn as below. 
\[ 
\input{tensor-f-2.tikz} 
\]
This parallel composition also extends to morphisms. Given morphisms $f \colon A \to C$ and $g \colon B \to D$ we can form their tensor $f \otimes g \colon A \otimes B \to C \otimes D$, depicted as below. 
\[
\input{tensor.tikz}
\]
The tensor is symmetric meaning that we can also `swap' pairs of wires past each other, such that swapping twice gives the identity (left-hand below), and boxes can move along the swaps (right-hand below). 
\[
\input{doubleswap.tikz} 
\qquad \qquad \qquad \qquad
\input{symmetry.tikz}
\]
In a monoidal category there is also a distinguished \emph{unit object} $I$, which is depicted simply as `empty space'. 
\begin{equation} \label{eq:empty-space}
\input{idI.tikz}  \ \ \quad 
\end{equation} 
Intuitively, tensoring any object by $I$ simply leaves it invariant.\footnote{Formally this is expressed via `coherence isomorphisms' $A \otimes I \simeq A \simeq I \otimes A$.} The unit object lets us now have special kinds of morphisms `without' inputs and/or outputs. A morphism $\omega \colon I \to A$ is called a \emph{state} of $A$, and is depicted with no input. An \emph{effect} is a morphism $e \colon A \to I$, depicted with no output.   
\[
\begin{tikzpicture}
	\begin{pgfonlayer}{nodelayer}
		\node [style=map] (0) at (0, 0) {$\omega$};
		\node [style=none] (1) at (0, 1.25) {};
		\node [style=label] (2) at (0, 1.75) {$A$};
	\end{pgfonlayer}
	\begin{pgfonlayer}{edgelayer}
		\draw (1.center) to (0);
	\end{pgfonlayer}
\end{tikzpicture}
 \qquad \qquad \qquad \begin{tikzpicture}[tikzfig]
	\begin{pgfonlayer}{nodelayer}
		\node [style=map] (0) at (0, 1.25) {$e$};
		\node [style=none] (1) at (0, 0) {};
		\node [style=label] (2) at (0, -0.5) {$A$};
	\end{pgfonlayer}
	\begin{pgfonlayer}{edgelayer}
		\draw (1.center) to (0);
	\end{pgfonlayer}
\end{tikzpicture}
 
\]
A \emph{scalar} is a morphism $r \colon I \to I$, depicted with no inputs or outputs. Scalars can move `freely' around diagrams, and also can be multiplied together via \ir{$r \cdot s := r \otimes s = r \circ s = s \cdot r$}, i.e.: 
\[ 
\input{scalars.tikz} 
\]
We denote the `empty space' scalar \eqref{eq:empty-space} by $1 := \id{I}$. By definition $1 \cdot r = r$ for all scalars $r$.  

The composition operations in a category satisfy numerous axioms which we mostly omit here but which are self-evident in the graphical language. For example, in any category associativity of composition: 
\[
(h \circ g) \circ f = h \circ g \circ f = h \circ (g \circ f)
\] 
is automatically apparent in string diagrams, with each expression given simply by drawing $f, g,$ and $h$ composed sequentially along a line. As another example, functoriality of $\otimes$ means that we always have:
\[
(f \otimes g) \circ (f' \otimes g') = (f \circ f') \otimes (g \circ g')
\]
In diagrams this is automatic, shown left-hand below, with a special case being the right-hand `interchange law' which lets us freely slide boxes along wires. 
\[
\input{functoriality2.tikz} \qquad \qquad \qquad 
\input{interchange2.tikz} 
\]
The categories we consider in this article typically come with some extra structure allowing us to `ignore' or `throw away' objects. 

\begin{definition} \cite{coecke2008axiomatic,coecke2014terminality} \label{def:disc-cat}
A \emph{discard-category} is an SMC in which every object comes with a chosen \emph{discarding} morphism, depicted as a ground symbol: 
\[
\begin{tikzpicture}
	\begin{pgfonlayer}{nodelayer}
		\node [style=upground] (0) at (4.5, 0.25) {};
		\node [style=none] (1) at (4.5, -1) {};
		\node [style=label] (3) at (4.5, -1.5) {$A$};
	\end{pgfonlayer}
	\begin{pgfonlayer}{edgelayer}
		\draw (1.center) to (0);
	\end{pgfonlayer}
\end{tikzpicture}

\]
Moreover we require the choice of discarding morphisms to be `natural' in that the following hold. 
\begin{equation} \label{eq:disc-nat}
\input{disc-nat.tikz} \qquad \qquad  \qquad \input{disc-I.tikz}
\end{equation}
A morphism $ f \colon A \to B$ is called a \emph{channel} when it preserves discarding:\footnote{Channels are also sometimes called `causal' \cite{coecke2014terminality} or `total' \cite{cho2015introduction}.}
\[
\input{causal2.tikz}
\]
 In particular, we call a state $\omega$ \emph{normalised} when the following holds.
\[
\begin{tikzpicture}
	\begin{pgfonlayer}{nodelayer}
		\node [style=map] (0) at (-0.5, -0.75) {$\omega$};
		\node [style=upground] (1) at (-0.5, 0.75) {};
		\node [style=none] (2) at (0.75, 0) {$=$};
		\node [style=none] (3) at (2, 0) {$1$};
	\end{pgfonlayer}
	\begin{pgfonlayer}{edgelayer}
		\draw (1) to (0);
	\end{pgfonlayer}
\end{tikzpicture}

\]
\end{definition}

The channels in $\catC$ form their own SMC $\catC_\channel$ in which $\discard{}$ is now the unique effect on any object\footnote{Formally this means that $I$ is a terminal object in $\catC_\channel$.}. A useful rule, which follows from \eqref{eq:disc-nat}, is that any channel with multiple inputs $f$ satisfies:
\[
\input{mech-disc.tikz}
\]
As we will see, categories with discarding provide a very general language for reasoning about composable processes of many kinds. One useful benefit is that discarding allows one to `ignore' certain outputs of morphisms. Given any morphism $f$ from $A$ to $B \otimes C$, its \emph{marginal} $A \to B$ is the following morphism. 
\[
\input{marginal.tikz} 
\]
 The final piece of extra structure we will use is the following ability to `copy' information. This is present for most but not all categories we will consider, being a signature of `classical' (deterministic or probabilistic) processes. 

\begin{definition}
\cite{cho2019disintegration}\label{def:cd_category}
A \emph{cd-category} (\emph{copy-discard category}) is a discard-category in which every object comes with a distinguished \emph{copying} morphism depicted as: 
\[
\input{just-copy.tikz}
\]
and satisfying the following:
\[
\input{markov-axioms.tikz}
\] 
Formally, this says that copying and discarding form a `commutative comonoid'. The copying morphisms are moreover required to be `natural' in that the following holds for all objects $A, B$.
\begin{equation} \label{eq:copy-nat-rules}
\input{copy-nat.tikz}
\end{equation}
A cd-category in which every morphism is a channel is called a \emph{Markov category} \cite{fritz2020synthetic}.
\end{definition}

Let us now introduce our main example categories for this article. These include deterministic processes, as used in classical computation and neural networks, probabilistic processes as used in Bayesian and causal models, and quantum processes as used in quantum computation.

\subsection{Deterministic processes}

Deterministic classical computation, as used in AI models based on neural networks, can be described by simply taking our processes to be functions, as follows.

In the cd-category $\Setcat$ the objects are sets $X, Y, \dots$ and morphisms are functions $f \colon X \to Y$. Composition $g \circ f$ is the usual composition of functions, with $\id{X}$ being the identity map $x \mapsto x$. The monoidal structure is the Cartesian product with $A \otimes B = A \times B$ and $f \otimes g = f \times g$, with $I=\{\star\}$ given by a singleton set. A state $x \colon I \to X$ of $X$ can be equated with an element $x \in X$ via $x = x(\star)$. Discarding on $X$ is given by the unique function $\discard{} \colon X \to I$, and copying $\tinycopy \colon X \to X \times X$ as expected by $x \mapsto (x,x)$.

$\Setcat$ is a Markov category, so every morphism is a channel. Moreover every morphism $f$ is in fact \emph{deterministic}, which means that the following holds. 
\begin{equation} \label{eq:deterministic}
\input{deterministic.tikz}
\end{equation}
In particular every state $x$ is deterministic, also called \emph{sharp}, meaning it is copied by the copy morphism:
\begin{equation} \label{eq:copy-points}
	\input{copy-points.tikz}
\end{equation}
In diagrams we draw such sharp states as triangles, as above.

Variants of this category are useful in AI in allowing us to model neural networks. We define the category $\NN$ just like $\Setcat$ but with objects being only sets of the form $V=\mathbb{R}^n$ for some $n$.\footnote{It is also common to define $\NN$ to only contain differentiable functions, to allow for the study of the training process, but we will not do so here and so for simplicity allow arbitrary functions.} Morphisms in $\NN$ are thus arbitrary (not necessarily linear)  functions $f \colon \mathbb{R}^n \to \mathbb{R}^m$. A state $v$ of $V$ corresponds to a vector $v \in \mathbb{R}^n$. The monoidal product is now often denoted $V \oplus W$ as it constitutes the direct sum of vector spaces, satisfying $\mathbb{R}^n \oplus \mathbb{R}^m = \mathbb{R}^{n + m}$. 
More broadly, it can be useful to include discrete sets in this setup. To do so, we define the category $\Comp$ just like $\Setcat$ but with objects being sets of the form $X_1 \times \dots \times X_n$, where each $X_j$ is either a finite discrete set or of the form $\mathbb{R}^n$ for some $n$.

String diagrams in these categories can precisely capture the `computational graphs' usually used to depict neural networks, as well as linear and rule based models, as we will see in Section \ref{sec:examples-compositional-models}.

\subsection{Probabilistic processes} \label{subsec:probabilistic-processes}

Recent work has demonstrated how cd-categories and Markov categories may be used to describe much of probability theory in string diagrams \cite{cho2019disintegration,fritz2020synthetic}. In particular, these provide a diagrammatic account of Bayesian networks and causal models \cite{fong2013causal,cho2019disintegration,lorenz2023causal}. For simplicity we will focus on probability with finite sets, but note that continuous probability theory can be treated in just the same way via suitable categories of `Markov kernels' \cite{fritz2020synthetic}.

In the cd-category $\MatR$, the objects are finite sets $X, Y, \dots$ and morphisms $M \colon X \to Y$ are functions $M \colon Y \times X \to \mathbb{R}^+$, where  $\mathbb{R}^+ := \{r \in \mathbb{R} \mid r \geq 0 \}$. We think of such a morphism as a positive `$X \times Y$ matrix', and following probabilistic notation denote its values by $M(y \mid x) := M(y,x)$. 
\[
\begin{tikzpicture}
	\begin{pgfonlayer}{nodelayer}
		\node [style=none] (0) at (0, 1) {};
		\node [style=none] (1) at (0, -1) {};
		\node [style=map] (2) at (0, 0) {$M$};
		\node [style=label] (3) at (0, 1.5) {$Y$};
		\node [style=label] (4) at (0, -1.5) {$X$};
	\end{pgfonlayer}
	\begin{pgfonlayer}{edgelayer}
		\draw (1.center) to (0.center);
	\end{pgfonlayer}
\end{tikzpicture}
 
\quad 
:: \quad   (y,x) \  \mapsto \ M(y \mid x)
\]
Sequential composition $N \circ M$ is given by the matrix product, i.e.~summation over internal wires. 
\[
\input{mat-comp.tikz}
\quad 
:: \quad  (z,x) \mapsto   \ \sum_{y \in Y} N(z \mid y) M(y \mid x)
\]
The tensor $\otimes$ is again given on objects by $X \otimes Y = X \times Y$, and on morphisms now by the Kronecker (tensor) product of matrices. 
\[
\input{mat-tens.tikz} 
\quad  :: \quad  (w, z, x ,y)  \  \mapsto \   M(w \mid x)N(z \mid y)
\]
The unit object is the singleton set $I = \{\star\}$, so that states $\omega$ and effects $e$ on $X$ correspond to positive functions on $X$:\footnote{Precisely, $\omega(x) := \omega(x \mid \star)$ and $e(x) := e(\star \mid x)$.} 
\[
\begin{tikzpicture}
	\begin{pgfonlayer}{nodelayer}
		\node [style=map] (0) at (0, -0.25) {$\omega$};
		\node [style=none] (1) at (0, 0.75) {};
		\node [style=label] (2) at (0, 1.25) {$X$};
	\end{pgfonlayer}
	\begin{pgfonlayer}{edgelayer}
		\draw (1.center) to (0);
	\end{pgfonlayer}
\end{tikzpicture}
   \ \ :: \ \ x  \ \mapsto  \ \omega(x)
\qquad \qquad \qquad 
\begin{tikzpicture}[tikzfig]
	\begin{pgfonlayer}{nodelayer}
		\node [style=map] (0) at (0, 0.75) {$e$};
		\node [style=none] (1) at (0, -0.5) {};
		\node [style=label] (2) at (0, -1) {$X$};
	\end{pgfonlayer}
	\begin{pgfonlayer}{edgelayer}
		\draw (1.center) to (0);
	\end{pgfonlayer}
\end{tikzpicture}
   \ \ ::  \ \ x  \ \mapsto  \ e(x)
\]
A scalar is precisely a positive real $r \in \mathbb{R}^+$, and composing scalars amounts to multiplication $r \otimes s = r \circ s = r \cdot s$ in $\mathbb{R}^+$. The copy map on $X$ is $\tinymultflip[whitedot](y,z \mid x) = \delta_{x,y,z}$ with value $1$ iff $x=y=z$ and $0$ otherwise. Discarding $\discard{}$ on $X$ is given by the function with $x \mapsto 1$ for all $x \in X$. 

A state $\omega$ is normalised precisely when it forms a normalised probability distribution over $X$, i.e. $\sum_{x \in X} \omega(x) = 1$. More generally, $M \colon X \to Y$ is a channel precisely when it forms a \emph{probability channel}, or equivalently the matrix is \emph{Stochastic}, meaning that it sends each $x \in X$ to a normalised distribution over $Y$:
\[
\begin{tikzpicture}
	\begin{pgfonlayer}{nodelayer}
		\node [style=none] (0) at (0, 1) {};
		\node [style=none] (1) at (0, -1) {};
		\node [style=map] (2) at (0, 0) {$M$};
		\node [style=label] (3) at (0, 1.5) {$Y$};
		\node [style=label] (4) at (0, -1.5) {$X$};
	\end{pgfonlayer}
	\begin{pgfonlayer}{edgelayer}
		\draw (1.center) to (0.center);
	\end{pgfonlayer}
\end{tikzpicture}
 \ \ \text{a channel} \ \ 
 \iff  \ \ 
\sum_{y \in Y} M(y \mid x) = 1 \ \ \forall x \in X.
\]
The category of channels in $\MatR$ is the Markov category $\FStoch$ of finite Stochastic matrices, describing probability channels between finite sets. Note that in both $\MatR$ and $\FStoch$ most states are not deterministic, i.e. in general we have:
\[
\input{notcopiable.tikz}
\]
In fact a normalised state $x$ is deterministic \irs{(sharp)} precisely when it is given by a point distribution $\delta_x$ for some $x \in X$. In general, deterministic morphisms $f \colon X \to Y$ in $\FStoch$ correspond precisely to functions $f' \colon X \to Y$, via $f \circ x = f'(x)$ for each sharp state $x$. Hence we can see deterministic processes (on finite sets) as living `inside' $\FStoch$.

String diagrams in $\FStoch$ can describe many notions from probability theory. For example, given any state or general process with outputs $Y, Z$, the marginal on $Y$ corresponds to taking the marginal in the usual sense, i.e.~by summation over the set corresponding to the discarded wire:
\[
\input{margM.tikz} :: (x,y) \mapsto \sum_{z \in Z}  M(y,z \mid x)
\]
We will see more aspects of probability theory cast as cd-diagrams later; in particular in Section \ref{subsec:causal-models} we will see how they can capture the structure of Bayesian networks and causal models. 

\subsection{Quantum processes} \label{subsec:quantum-processes}

The categorical approach allows us to consider very general kinds of processes, including those used in quantum computation. Here we briefly sketch the categorical treatment of quantum processes; for more details see \cite{abramsky2004categorical,coecke2018picturing}. 

In the category $\Quant{}$ the objects are finite-dimensional Hilbert spaces (equivalently, finite-dimensional complex vector spaces) $\hilbH, \hilbK, \dots$, with 
$\hilbH \otimes \hilbK$ given by the usual tensor product of vector spaces, and $I=\mathbb{C}$ being the one-dimensional space. We write $L(\hilbH)$ for the space of linear operators on $\hilbH$. A morphism $\hilbH \to \hilbK$ is a linear map 
\[
f \colon L(\hilbH) \to L(\hilbK)
\] which is \emph{completely positive} (CP). This means that $f$ is \emph{positive}, 
\ir{sending positive operators $a \in L(\hilbH)$ -- that is, ones for which there is some $b \in L(\hilbH)$ such that $a=b^\dagger b$ --}
to positive operators $f(a)$ in $L(\hilbK)$, and moreover is such that each operator $f \otimes \id{\hilbH'}$ is also positive, for any other Hilbert space $\hilbH'$. \stb{Here $b^\dagger$ denotes the Hermitian conjugate of the operator $b$, i.e.~the complex conjugate of the transpose.}

In general, an effect $e$ on $\hilbH$ may be equated with a positive operator $a \in L(\hilbH)$ via $e(b) = \Tr(ab)$, where $\Tr$ denotes the trace, with discarding given by the map $\discard{} \colon a \mapsto \Tr(a)$ and corresponding to the identity operator \irs{$1$}. 
Similarly, a state $\omega$ of $\hilbH$ may also be equated with a positive operator \irs{$\rho \in L(\hilbH)$ on $\hilbH$ via $\rho=\omega(1)$}. 

A normalised state is then precisely a density matrix (trace one positive operator), i.e.~a quantum state:
\[
\begin{tikzpicture}
	\begin{pgfonlayer}{nodelayer}
		\node [style=map] (0) at (-1, -0.5) {$\rho$};
		\node [style=none] (1) at (-1, 0.5) {};
		\node [style=none] (2) at (-1, -0.5) {};
		\node [style=label] (3) at (-1, 1) {$\hilbH$};
	\end{pgfonlayer}
	\begin{pgfonlayer}{edgelayer}
		\draw (2.center) to (1.center);
	\end{pgfonlayer}
\end{tikzpicture}
 \ \ \ \ \text{is normalised} \qquad \iff \qquad \rho \in L(\hilbH)\text{ is a density matrix  }
\]
In general, a channel is precisely a CP trace-preserving (CPTP) map, the standard notion of a quantum operation. Such a channel maps density matrices on its input to density matrices on its output. 
\[
\begin{tikzpicture}
	\begin{pgfonlayer}{nodelayer}
		\node [style=map] (0) at (-1, 0) {$f$};
		\node [style=none] (1) at (-1, 1) {};
		\node [style=none] (2) at (-1, -1) {};
		\node [style=label] (3) at (-1, 1.5) {$\hilbK$};
		\node [style=label] (4) at (-1, -1.5) {$\hilbH$};
	\end{pgfonlayer}
	\begin{pgfonlayer}{edgelayer}
		\draw (2.center) to (1.center);
	\end{pgfonlayer}
\end{tikzpicture}
 \ \ \ \ \text{ is a channel} \qquad \iff \qquad f \colon L(\hilbH) \to L(\hilbK) \text{ is a CPTP map }
\]
Taking the marginal of a CP map is given by the \emph{partial trace} operation. 
\[
\input{partial-trace.tikz} \ \ = \ \ \Tr_\hilbL (f)
\]
Of particular interest are those CP maps $\hat f \colon L(\hilbH) \to L(\hilbK)$ which are \emph{pure}, meaning they are of the form $\hat f(a) = f \circ a \circ f^\dagger$ for some linear map $f \colon \hilbH \to \hilbK$.\footnote{
Formally, the pure maps are the morphisms in the image of a monoidal `embedding' $\FHilb \to \Quant{} $ given by $f \mapsto \hat f$. This mapping equates any two linear maps only when they are equal up to global phase, i.e. $\hat f = \hat g \iff f = e^{i \theta} \cdot g$ for some $\theta \in [0,2\pi)$.} The normalised pure states are those of the form $\psi = \ket{\psi} \bra{\psi}$, where $\ket{\psi} \in \Hilb$ is a unit vector and $\bra{\psi}$ the corresponding dual linear functional on $\hilbH$. 

String diagrams in $\Quant{}$ can be used to describe the quantum circuits featured in quantum computation, including the preparation of states (density matrices), applications of unitary gates $\hat U$ for a unitary matrix $U \colon \hilbH \to \hilbH$, discarding of \irs{systems}, \irs{and post-selective measurements given by effects}, as depicted below. Each individual quantum wire typically corresponds to a `qubit', i.e. 2-dimensional system $\hilbH = \mathbb{C}^2$. 
\[
\input{generic-circuit.tikz}
\]
The compositional perspective has been particularly fruitful for modelling quantum processes due to the inherently compositional nature of quantum circuits, and thanks to the existence of a complete graphical language for describing pure quantum processes, the \emph{ZX-calculus} (and related \emph{ZW} and \emph{ZXW} calculi) \cite{coecke2008interacting,van2020zx,poor2023completeness,hadzihasanovic2018two}.

\begin{remark}
Full measurement setups, including the representation of classical measurement outcomes, can similarly be described in the broader category of finite-dimensional C*-algebras, which now includes both quantum and finite classical systems; for more details, see \cite{coecke2018picturing}. 
\end{remark}


\section{Compositional models} 
\label{sec:compositional-models}


We now introduce the general notion of a compositional model, through which we will later understand AI models. 
Intuitively, such a model consists of an abstract `structure', consisting of a collection of (usually learned) basic variables and processes between them, called `generators', which we may then compose to describe further processes. The model then specifies semantics for this structure in a particular category, e.g that of neural networks, or quantum circuits. We will start by giving all the formal notions needed for a model at once, before illustrating them together in an example. 

\begin{definition} 
A \emph{(monoidal) signature} $\Sig$ consists of: 
\begin{itemize}
\item 
a set $\Sigob$ of `objects', which we call \emph{variables}; 
\item 
a set $\Sigmor$ of  `morphisms' $\syn{f},\syn{g}, \dots$, called \emph{generators}, whose inputs and outputs are lists of objects\footnote{Formally, these come with functions $\inputs, \outputs \colon \Sigmor \to \Sigob^*$ sending each generator to a list of input variables and output variables, respectively.}; 
\item 
a set $\Sigeq$ of \emph{equations} of the form $\diagD_1 = \diagD_2$ where $\diagD_1, \diagD_2$ are string diagrams built from the generators.
\end{itemize}
\end{definition} 

\irs{Given a signature we depict its variables as wires, and generators as boxes with their input and output variables as input and output wires, just as for morphisms in a category. A generator with an empty list of inputs is again called a \emph{state} and depicted with no input wires}. 

Typically a signature is thought of as a collection of basic objects and processes which we combine together to generate a category, as follows. Given some fixed class of string diagrams (e.g. monoidal or cd-diagrams), we write $\Free(\Sig)$ for the category whose objects are lists $(\syn{A_i})^n_{i=1}$ of variables in $\Sigob$, and morphisms $\diagD \colon (\syn{A_i})^n_{i=1} \to (\syn{B_j})^m_{j=1}$ are such string diagrams $\diagD$ built from $\Sigmor$ with inputs $(\syn{A_i})^n_{i=1}$ and outputs $(\syn{B_j})^m_{j=1}$.\footnote{Composition is given simply by composing diagrams `on the page', the empty list is the monoidal unit and each identity morphism is the diagram of all plain wires.} We consider two diagrams $\diagD, \diagD'$ equal as morphisms if one can be rewritten to the other in finitely many steps using equations in $\Sigeq$ and standard rules for such string diagrams.

We now reach our main definition.

\begin{definition}
\label{def:compositional-model}
A \emph{compositional model} $\modelM=(\Sig,\catC,\sem{-})$ consists of: 
\begin{itemize}
	\item a signature $\Sig$, which induces a \emph{structure} category $\catS = \Free(\Sig)$;
	\item a \emph{semantics} category $\catC$;
	\item a \emph{representation} functor: 
\[
\sem{-} \colon \catS \to \catC
\] 
For any object or morphism $\syn{A}, \syn{f}$ in $\catS$ we call $\sem{\syn{A}}$, $\sem{\syn{f}}$ its \emph{semantics} or \emph{representation} in $\catC$.   
\end{itemize}
In the above $\catS$ and $\catC$ are always categories of some explicitly specified type (e.g. monoidal or cd-categories)  where $\catS=\Free(\G)$ consists of all diagrams of this kind, and $\sem{-}$ is then a functor which preserves this structure (e.g. a monoidal or cd-functor). 
\end{definition}

The definition is the standard notion of model from \ir{categorical logic \cite{Lawvere_PhD_Thesis, lawvere1963functorial}}. We refer to such a model $\modelM$ as a (compositional) model \emph{of $\catS$ in $\catC$}. Often we don't write $\sem{-}$ explicitly but distinguish objects and morphisms in $\catS$ from their representation in $\catC$ by using different fonts, so that objects and morphisms $\syn{A}, \syn{f}$ in $\catS$ are mapped to $A := \sem{\syn{A}}$ and $f := \sem{\syn{f}}$ in $\catC$.  

Since the structure category is freely built, specifying such a model simply amounts to specifying the representation of each variable and generator. That is, we specify an object $A = \sem{\syn{A}}$ of $\catC$ for each variable $\syn{A} \in \Sigob$ and a morphism $f$ in $\catC$ of the appropriate type, as below, for each generator $\syn{f}$: 
\begin{equation} \label{eq:generator-map}
\input{gen-map.tikz}
\end{equation}
such that $\sem{-}$ preserves the equations in $\Sigeq$.

\rl{For a reader not familiar with above notion of a model from categorical logic, and with a view to AI models, the upshot of the separation between syntax and semantics is to be able to clearly distinguish between, and yet relate to each other, the high-level abstract structure of a model and its concrete instantiation or implementation. The benefits of this distinction will be illustrated through numerous examples in Sec.~\ref{sec:examples-compositional-models}.}

\paragraph{Models from Diagrams.}

Often one can specify a model simply by drawing a single string diagram, where the variables and generators correspond to the wires and boxes. To make this precise, let $\diagD$ be a string diagram of a given kind. By a \emph{compositional model of $\diagD$} in $\catC$ we mean a compositional model $\modelM$ in $\catC$ whose signature $\G_\diagD$ has the wires in $\diagD$ as its variables and the boxes in $\diagD$ as its generators. 
The model $\modelM$ is understood to come with the diagram $\diagD$ itself as a distinguished morphism in $\catS$. 
We will also refer to the pair of a diagram $\diagD$ with a given model in $\catC$ as a \emph{diagram in $\catC$}.

Rather than labelling the wires and boxes abstractly as $\syn{A}, \syn{f}$, at times we may simply label them via their representations $A=\sem{\syn{A}}$, $f=\sem{\syn{f}}$ in $\catC$, thus capturing the whole model (structure + semantics) at once in a single diagram. 

\paragraph{}

The following toy example now illustrates together the notions of signature, model, and capturing a model in a diagram. 

\begin{example}
Let $\catC$ be a cd-category. Suppose that $\modelM$ is a compositional model in $\catC$, in the language of cd-categories, of the following form: 

\begin{equation} \label{eq:diag-in-C}
\input{diag-in-C.tikz}
\end{equation}
This means that the signature $\Sig$ consists of variables $\syn{A},\syn{B},\syn{C},\syn{D},\syn{E}$ and generators: 
\begin{equation} \label{eq:generators-ex}
\input{cm-blocks.tikz}
\end{equation}
The structure category $\catS = \Free(G)$ consists of all cd-diagrams that can be  built from these variables and generators, including for example the following diagrams as morphisms:
\begin{equation} \label{eq:free-diags}
\input{diags.tikz}
\end{equation}
Furthermore the model comes with the fourth diagram above as a distinguished morphism in $\catS$. Specifying the semantics of the generators $\G$ in \eqref{eq:generators-ex}, i.e. the representation functor $\sem{-}$, amounts to specifying corresponding objects $A, B, C, D, E$ of $\catC$ and morphisms in $\catC$ of the form:
\[
\input{cm-blocks.tikz}
\]
with $A = \sem{\syn{A}}$ and $a = \sem{\syn{a}}$ etc. The functor $\sem{-}$ then extends this mapping to give any morphism (i.e. diagram) $\diagD$ in $\catS$ a representation as a morphism $\sem{\diagD}$ in $\catC$. For example, if $\diagD$ is the fourth diagram in \eqref{eq:free-diags} then $\sem{\diagD}$ is equal to the morphism \eqref{eq:diag-in-C} in $\catC$.
\end{example}

\paragraph{Maps of signatures.}
Our representation functor can also be defined while referring only to the signature $\Sig$, using the following notion which will be helpful in what follows. Given signatures $\Sig, \Sig'$, by a \emph{map of signatures} $F \colon \Sig \to \Sig'$ we mean a map of variables $F \colon \Sigob \to \Sigob'$ along with a map $F \colon \Sigmor \to \Sigmor'$ which preserves types, i.e. of the form:
\[
\input{mor-signatures.tikz}
\]
and which also maps equations in $\Sigeq$ to equations in $\Sigeq'$.\footnote{\irs{Formally, the mapping equations condition is expressed as follows. Write $\hat\Sig, \hat\Sig'$ respectively for signatures $\Sig, \Sig'$ with their equations removed. Then $F$ induces a functor $F' \colon \Free(\hat \Sig) \to \Free(\hat \Sig')$. For each equation $\diagD_1 = \diagD_2$ in $\Sigeq$ we require that $F$ is such that $F'(\diagD_1) = F'(\diagD_2)$ in $\Free(\hat \Sig')$. Then $F$ extends to a functor $\Free(\Sig) \to \Free(\Sig')$.}}  As a special case, one can in fact view any monoidal category $\catC$ as a signature with a variable for every object, generator for every morphism, and every valid equation in its equation set. Then a (strong monoidal) functor $\sem{-} \colon \Free(\Sig) \to \catC$ as in a compositional model \eqref{eq:generator-map} is the same as a map of signatures $ \sem{-} \colon \Sig \to \catC$.

We will also need to consider partial such mappings. We define a \emph{partial map} of signatures $F$ in the same way, only requiring the maps $\Sigob \to \Sigob'$ and $\Sigmor \to \Sigmor'$ to be partial functions, such that if $\syn{f}$ is a generator and $F(\syn{f})$ is defined then so is $F(\syn{V})$ whenever $\syn{V}$ is an input or output of $\syn{f}$. Moreover, we no longer require any map on the equations of $\Sig$. 

\subsection{Interpretations of compositional models}
\label{subsec:interpretations-of-models}

Having introduced compositional models fully formally, in this section we will propose a notion of interpretability of a compositional model. This assigns meanings to (some of) the components and processes within a compositional model, from a collection of `human-friendly' concepts. 

Before doing so, we briefly introduce one more notion. To interpret processes in the semantics category $\catC$, one must know precisely what variables are being viewed as their inputs and outputs. \stb{To capture this, we define the category $\catC_\Sig$ to have as objects lists of variables $(\syn{A_i})^n_{i=1}$ with $\syn{A_i} \in \Sigob$ and with a morphism $f \colon (\syn{A_i})^n_{i=1} \to (\syn{B_j})^m_{j=1}$ in $\catC_\Sig$ for each morphism $f \colon \bigotimes^n_{i=1} A_i \to \bigotimes^m_{j=1} B_j$ in $\catC$. Composition operations are the same as in $\catC$.}

\stb{
Hence morphisms in $\catC_\Sig$ are those of $\catC$ but with their inputs and outputs indexed by variables from $\Sigob$. For example, states of variables $\syn{A}$, $\syn{B}$ are regarded as states of distinct objects in $\catC_\Sig$, even if the variables' semantics coincide so that $A=B$ in $\catC$. }






We now reach our definition.

\begin{definition}[Interpretation] 
\label{def:interpretation}
An \emph{interpretation} $\Interp = (\Human, \Interpabs, \Interpcon)$ for a compositional model $\modelM=(\Sig,\catC,\sem{-})$ consists of:
\begin{itemize}
	\item 
a signature $\Human$ of \emph{human-friendly} terms;
\item 
partial maps of signatures $\Interpabs$, the \emph{abstract interpretation}, and $\Interpcon$, the \emph{concrete interpretation}, such that the following diagram commutes:\footnote{Formally, each arrow represents a partial map of signatures, where we view the category $\catC_\Sig$ as a special case of a signature, and $\sem{-}$ denotes as expected the (total) map $\syn{V} \mapsto V = \sem{\syn{V}}$, $\syn{f} \mapsto f = \sem{\syn{f}}$. } 
\begin{equation} \label{eq:interp-commutes}
\input{concrete-triangle-2-nopartial.tikz}
\end{equation}
\end{itemize}
We call the image in $\Human$ of a variable $\syn{V}$, generator $\syn{f}$, or morphism $g$ in $\catC_\Sig$ its \emph{interpretation}. We say a variable, generator or entire signature \emph{has an abstract interpretation} when $\Interpabs$ is defined on it.
Finally, we will say that a variable $\syn{V}$ itself has a \emph{concrete interpretation} when $\Interpcon(\omega \colon \syn{I} \to \syn{V})$ is defined for every state $\omega$ of $V$ in $\catC$. 
\end{definition}

Let us now unpack this definition in detail, before making it more vivid in an example shortly. The definition is \st{not fully formal} in that we have not specified what makes a term `human-friendly'. Formally, we have only asked that these are organised into a signature $\Human$. This simply means we can separate the terms into collections of interpretations $\Humanob$ for variables and objects, and $\Humanmor$ for processes, and the above mappings respect these, explicitly taking the form:
\begin{equation} \label{eq:abstr-conc-pic}
\input{abstr-interp-explicit.tikz}
\qquad \qquad \qquad \qquad 
\input{conc-interp-explicit.tikz}
\end{equation}
The commutativity condition \eqref{eq:interp-commutes} means that the abstract and concrete interpretations coincide on variables and generators; that is, 
$\Interpcon(\syn{V}) = \Interpabs(\syn{V})$ for all variables $\syn{V}$ and $\Interpcon(\sem{\syn{f}}) = \Interpabs(\syn{f})$ for all generators $\syn{f}$, in each case whenever the former is defined. 

First, let us consider a model with an abstract interpretation $\Interpabs \colon \Sig \to \Human$ only. This assigns terms from $\Human$ to (some of) the variables and generators of the model. We can think of this as simply `labelling' them in human-friendly terms. For example we may say that a variable $\syn{V}$ corresponds to `brightness'. 
Note that an abstract interpretation depends only on the structure $(\Sig, \catS)$ of the model, being independent from its semantics $(\catC,\sem{-})$.\footnote{Informally, we may also think of an abstract interpretation as often coming with assigned meanings for the composition operations $(\circ, \otimes, \syn{I})$ of $\catS$ in relation to the concepts of $\Human$. This allows one to extend the interpretation to one of diagrams in $\catS$.}

It is common however to also wish to interpret additional parts of the semantics of a model. For example, given our `brightness' variable with semantics $V=\mathbb{R}$, we may wish to interpret each of its states as a degree of brightness, e.g. $0 \mapsto$ `Dark', $1 \mapsto$ `Bright' etc. This is the role of a concrete interpretation $\Interpcon$. Explicitly, given lists of variables such as $\syn{A_i}$ and $\syn{B_j}$, this provides a partial map allowing us to interpret morphisms $g$ in $\catC$ from these inputs to outputs, as shown right-hand above. An important special case is that of states, where a variable $\syn{V}$ is said to have a concrete interpretation when we can assign an interpretation $\Interpcon(\omega)$ to every state $\omega$ of $V$. 

The fact that $\Interpcon$ is indexed by variables, and so defined on $\catC_\Sig$ rather than simply on  $\catC$, is essential as we must know what variables are being considered to apply an interpretation. For example, consider variables $V_1 = V_2 = \mathbb{R}$ interpreted as `brightness' and `size' respectively. A particular value $r = 0.5$ forms a state of $\mathbb{R}$ in $\catC$ but can only be interpreted once a variable is chosen, i.e.~as a state in $\catC_\Sig$ of either $\syn{V_1}$ (a brightness) or $\syn{V_2}$ (a size). 

Note that typically the mappings of an interpretation are only partial, in that some variables or generators may have a clear meaning, while others may not. We say a model has a \emph{complete interpretation} when every variable and generator has an abstract interpretation, and a \emph{complete concrete interpretation} when moreover every variable has a concrete interpretation.

\begin{example} \label{ex:sprinkler-interpretation}
Consider a compositional model predicting how slippery the floor will typically be outside a house, depending on the season, generated by the diagram below.\footnote{This is a classic example of an (open) causal model, treated more generally in Section \ref{subsec:causal-models}.} 
\begin{equation} \label{eq:sprinkler-diagram}
\input{sprinkler-nd-2.tikz}
\end{equation}
Hence our signature contains variables $\syn{Se}, \rl{\syn{R}, \syn{Sp},}  \syn{W}, \syn{Sl}$ and generators $\syn{f},\syn{g},\syn{h},\syn{k}$ as above.

Let us now equip this structure with the intended abstract interpretation. Our collection of human-friendly concepts $\Human$ will have as its set of interpretations for objects 
\rl{$\Humanob := \{ \text{`Season'}, \text{`Rain'}, \text{`Sprinkler'}, \text{`Wetness'}, \allowbreak \text{`Slipperiness'} \}$.} 
The interpretation of variables $\Interpabs \colon \Sigob \to \Humanob$ acts as follows: 
\begin{align*}
\syn{Se} &\mapsto \text{`Season'} \\ 
\syn{R} &\mapsto \text{`Rain'} \\ 
\syn{Sp} &\mapsto \text{`Sprinkler'} \\ 
\syn{W} &\mapsto \rl{\text{`Wetness'}} \\ 
\syn{Sl} &\mapsto \rl{\text{`Slipperines'}} 
\end{align*}

On generating morphisms, the mapping $\Interpabs \colon \Sigmor \to \Humanmor$ acts as below, where $\Humanmor$ consists precisely of all the morphisms on the right-handsides, with inputs and outputs as shown. 
\begin{align*}
\rl{\input{sprinkler-interp-combined_v2.tikz}}
\end{align*}
In each case the morphism in $\Human$ has a label giving a short-hand for our understanding of the concept. For example \rl{` avg. rainfall'} reflects our interpretation of $\syn{g}$ as the mapping from seasons to their average rainfall.

 In this scenario, we might expect to furthermore make use of a concrete interpretation. Suppose our semantics is in the category $\catC = \Setcat$ of sets and functions, with the specific semantics $V = \sem{\syn{V}}$ for each variable $\syn{V}$ given by $Se = \{w,sp,su,a\}, Sp = \{0,1\}, R = W = Sl = \mathbb{R}$. 
The states of each variable in $\Setcat$ are then its elements as a set. Suppose we expand $\Human$ to include states of `Sprinkler' called `on' and `off'. A concrete interpretation for $\syn{Sp}$ could then be given by the mapping: 
\[
\input{conc-interp-sprinkler.tikz}
\]
Similarly, a concrete interpretation for $\syn{Se}$ could be given by mapping the states $w, a, sp, su$ of $Se$ to states of \text{`Season'} in $\Human$ called `winter', `autumn', `spring' and `summer', respectively.

To give a concrete interpretation of $R$, we could include states of `Rain' in $\Human$ called `no rain' and `$r$-mm' for each $r > 0$, and then specify the following mapping from states of $R$ (i.e. real numbers) to states in $\Human$:  
\[
\begin{tikzpicture}
	\begin{pgfonlayer}{nodelayer}
		\node [style=map] (0) at (-5, -0.5) {$r$};
		\node [style=none] (1) at (-5, 0.75) {};
		\node [style=label] (2) at (-5, 1.25) {$R$};
	\end{pgfonlayer}
	\begin{pgfonlayer}{edgelayer}
		\draw (1.center) to (0);
	\end{pgfonlayer}
\end{tikzpicture}
 \ \ \  \mapsto \ \ \ \ 
\begin{cases}
\begin{tikzpicture}
	\begin{pgfonlayer}{nodelayer}
		\node [style=map] (0) at (-5, -0.5) {no rain};
		\node [style=none] (1) at (-5, 0.75) {};
		\node [style=label] (2) at (-5, 1.25) {Rain};
	\end{pgfonlayer}
	\begin{pgfonlayer}{edgelayer}
		\draw (1.center) to (0);
	\end{pgfonlayer}
\end{tikzpicture}
 & r \leq 0 \\ \begin{tikzpicture}
	\begin{pgfonlayer}{nodelayer}
		\node [style=map] (0) at (-5, -0.5) {$r$ mm };
		\node [style=none] (1) at (-5, 0.75) {};
		\node [style=label] (2) at (-5, 1.25) {Rain};
		\node [style=none] (4) at (-5, 2.5) {};
	\end{pgfonlayer}
	\begin{pgfonlayer}{edgelayer}
		\draw (1.center) to (0);
	\end{pgfonlayer}
\end{tikzpicture}
 & r > 0 
\end{cases}
\]
Similarly, each state $r \in \mathbb{R}$ of $W$ could denote the total mm of water on the ground, with $r \leq 0$ denoting dry, while each state $r \in [0,1]$ of $S$ could denote a degree of slipperiness, with all other states of $S$ lacking an interpretation. Note that, even if we interpret all states of all variables, $\Interpcon$ is still only a partial map of signatures. For example we have not defined interpretations for any other morphisms (i.e. functions) $Se \to Sp$ or any morphisms $W \to R$. 
\end{example}



\begin{remark}[Specifying interpretations in practice] \label{rem:specifying-interrpetations}
For clarity, we have in Example \ref{ex:sprinkler-interpretation} gone to great lengths to spell out the interpretation explicitly. In practice, and in the remainder of this article, we typically specify it much more briefly. For example we may simply say `the variables $\syn{Se}, \syn{R}, \syn{Sp}, \syn{W}, \syn{Sl}$ have concrete interpretations in terms of season, rain, sprinkler settings, wetness, and slipperiness, and $f, g, h, k$ have interpretations as 
\rl{sprinkler activation, average rainfall, total moisture and inducing slipperiness, respectively'.} 
Typically for brevity we will also not specify the signature $\Human$ or mappings $\Interp$ explicitly, but only implicitly by directly using human-friendly concepts as names for generators and for morphisms in $\catC$. Thus we may instead describe the model with a diagram such as:  
\[
\rl{\input{sprinkler-nd-interpreted_v2.tikz}}
\]
\end{remark}

\begin{remark}[Functorial interpretations]
From a category-theoretic perspective, the view of $\Human$ as a signature may suggest that it should come with its own composition operations also, and thus form a (monoidal) category. Though we will not require this notion here, we can call an interpretation \emph{functorial} when $\Human$ itself forms a monoidal category and $\Interpcon$ a (partial) monoidal functor. In this case, $\Interpabs$ extends freely to a functor on $\catS=\Free(\G)$ and the following is a commutative diagram of (partial) functors: 
\[
\input{concrete-triangle-nopartial.tikz}
\]
\stb{This more richly structured form of interpretation is appealing categorically, and may be found to be the natural one from a compositional perspective, making it worth exploring in future work. However, we do not make this functorial notion of interpretation our default here since the majority of models relevant to XAI do not come with such a well-structured collection of terms, and so their $\Human$ does not obviously form a category.}
\end{remark}

\section{Examples of compositional models} \label{sec:examples-compositional-models}

In this section we will describe numerous examples of our notion of compositional model, from across AI. This serves to illustrate the definitions of compositional model and interpretation from the previous section, and also demonstrate how familiar (and perhaps less familiar) models may be cast diagrammatically. Each model will be defined in terms of string diagrams, which implicitly specify the signature $\Sig$ and the language of string diagrams it is based on, from which the structure category $\catS=\Free(\Sig)$ is in turn defined. For example, for most models the diagrams contain copying morphisms, meaning that $\catS$ and $\catC$ are cd-categories and $\sem{-} \colon \catS \to \catC$ is a cd-functor. 

We emphasise that it is not essential for a reader to understand every example in detail. The reader is invited to pursue the list of examples from the Contents and choose those which interest them most. In Section \ref{sec:comp-and-interp} we will summarise our main conclusions from these examples regarding the role of compositionality in interpretability.

\subsection{Classification models} \label{subsec:classify-model}

As a simple example of a compositional model in our sense, let us consider a basic classification problem. Suppose we are given labelled data coming from a distribution $\syn{Data}(x,l)$ over \emph{inputs} $x \in X$ and \emph{labels} $l \in L$. For simplicity assume $X, L$ are both finite sets, so that $\syn{Data}$ forms a normalised state in $\FStoch$. A \emph{classification model} is then a compositional model of the $\syn{Data}$ in $\FStoch$ of the following form:  
\begin{equation} \label{eq:regression-simple}
\input{regression-simpler.tikz}
\end{equation}
Thus the model has variables $\syn{X}, \syn{L}$, and generators $\syn{f}$, $\syn{Data}$. By the choice of category $f = \sem{\syn{f}} \colon X \to L$ is a channel (and $\sem{\syn{Data}}$ is the normalised distribution underlying the dataset). In standard notation, the above equation states that:
\[
\syn{Data}(x,l) =  f(l \mid x) \sum_{l'} \syn{Data}(x, l')
\]
If each data point $x$ has a unique label $l= l(x)$ for which $\syn{Data}(x,l)$ is non-zero, then it follows that $f(l \mid x) = \delta_{l,l(x)}$, so that $f$ chooses the correct label. We may then apply $f$ to classify new inputs. In practice one is unlikely to have a perfect classifier, and so equation \eqref{eq:regression-simple} may only hold approximately, with $f$ given by the outcome of a training procedure.

\paragraph{Interpretation.}
In a classification model, typically the space of labels $L$ is a (finite) set with a concrete interpretation, so that each label $l \in L$  comes with a specific interpretation. The inputs $X$ may either be given in terms of interpretable features $X_1,\dots X_n$, as in typical data science scenarios, or as an abstract latent space in a broader ML model, lacking a concrete interpretation. The generator $\syn{Data}$ is interpreted as the distribution underlying the dataset. The map $\syn{f}$ has a simple interpretation as mapping each input to its correct label, but without further decomposition of $X$ and $f$ we may not be able to interpret how this is done any further.

\subsection{Encoder-decoder models} \label{subsec:encoder-decoder-model}



For our next example, we consider the common form of model in which data from some input space is encoded via a  latent space. By a \emph{strict encoder-decoder model} we mean a compositional model $\modelM$ in the language of cd-categories, of the following form. The model features an \emph{input space} variable $\syn{X}$, a \emph{representation space} variable $\syn{Z}$, and the following generators:
\[ 
\input{encoder-decoder.tikz}
\]
Here $d$ is called the \emph{decoder}, $e$ the \emph{encoder}, $\sigma$ the \emph{prior} over $Z$, and $\sem{\syn{Data}}$ is equal to a given data distribution over inputs $X$. 
To ensure that these do behave as one would expect of an encoder and decoder, we include equations stating that all of these are channels, and that the following holds:  
\begin{equation} \label{eq:strict-encoder-binv}
\input{encoder-binv.tikz}
\end{equation}
This equation relates the distribution over $X$ underlying the data to our prior distribution over $Z$, via the encoder and decoder. In fact it states precisely that the decoder forms the \emph{Bayesian inverse} of the encoder with respect to this prior $\sigma$. In particular, given such a model we can sample from $Z$ via the prior $\sigma$ and apply the decoder, to give a distribution which generates new inputs from $X$:
\[ 
\input{generate-simpler.tikz} 
\]
Marginalizing out $Z$ in the equation \eqref{eq:strict-encoder-binv} tells us that the above state is equal to $\sem{\latdata}$, i.e. that this newly generated data will match the original distribution.  If we include an extra equation specifying that $\syn{d}$ is deterministic, one may then prove graphically the desirable property that encoding and then decoding leaves all data points invariant, i.e. the following holds. 
\begin{equation} \label{eq:strong-encoder}
\input{encoder2.tikz}
\end{equation}

\paragraph{VAEs.}
Practical encoder-decoders such as the \emph{Variational Autoencoder (VAE)} \cite{kingma2013auto} do not require equation \eqref{eq:strict-encoder-binv} to hold strictly, instead using a loss function to minimize some distance between both sides.\footnote{In a VAE it is also typically assumed that all channels (aside from the data distribution) take a particular Gaussian form.} 

For a model in $\FStoch$, let us define the \emph{structural loss} $L(\model{V})$ as the Kullback-Leibler divergence $D_{KL}$ between the distributions over $Z, X$ given by the LHS and RHS of \eqref{eq:strict-encoder-binv}. Then minimizing this loss would minimize $D_{KL}$ between their marginals on $X$, so that generating new inputs via $d \circ \sigma$ returns points of $X$ distributed closely to $\latdata$. In practice, this structural loss is impractical to compute directly, and so a VAE is instead trained by minimizing the \emph{ELBO loss} defined by: 
\begin{equation} \label{eq:ELBO-loss}
\mathsf{ELBO}(\model{V}) := \Expval_{x \sim \latdata} 
\left[ D_{KL}(e(x),\sigma) -\Expval_{z \sim e(x)}\log d(x \mid z)  \right]
\end{equation}
Indeed one may prove directly that $L(\model{V}) \leq \mathsf{ELBO}(\model{V})$, so by minimizing the latter we can reduce $L(\model{V})$. \irs{For more details on VAEs and the ELBO loss see for instance \cite{doersch2016tutorial}}.

Encoder-decoder models can be extended in various ways, such as by including a collection of `concepts' over the representation space, as in the Conceptual VAE \cite{QonceptsFull2024} outlined in Appendix \ref{subsec:concept-encoders}, and related to conceptual space models considered in Section \ref{subsec:conceptual-space-models}.

\paragraph{Interpretation.}
Encoder-decoder models can appear within models of many kinds. A common case is where each data point in the space $X$ has a concrete interpretation, for example, $X=[0,1]^{3 \times n \times m}$ may be the space of images given by $n \times m$ many pixels represented as RGB values in $[0,1]^3$, and the data a collection of images. In contrast, $Z$ is often an uninterpreted hidden state space used to compress the data. As a result the only interpretation for the encoder and decoder is simply to say that they are indeed an encoder and decoder, with their precise workings uninterpreted.

\subsection{Linear models} \label{subsec:linear-models}

The simple example of a linear model is based on the use of specific morphisms available in the category $\catC=\NN$. Any object $V=\mathbb{R}^n$ in this category comes with the following distinguished morphisms:
\begin{equation} \label{eq:linear-generators}
\input{linear-generators.tikz}
\end{equation}
corresponding to \emph{addition} and \emph{scalar multiplication} respectively. As expected, addition is the function $(v,w) \mapsto v + w$ and scalar multiplication is $(r, v) \mapsto r \cdot v$. For any state $r$ of $\mathbb{R}$, the operation $v \mapsto r \cdot v$ of `multiplication by $r$' is then given by:
\begin{equation} \label{eq:rmult}
\input{rmult.tikz}
\end{equation} 
One may draw string diagrams expressing the usual relations satisfied by each of these operations, for example that $r \cdot (v + w) = r \cdot v + r \cdot w$. For more details, see Appendix \ref{appendix:linear-models}.

Using the above structure we can now specify linear models as string diagrams. A \emph{linear model} is a compositional model in $\catC=\NN$ of the form: 
\begin{equation} \label{eq:linearmodel}
\input{linearmodel3.tikz} 
\end{equation} 
where we assert that $X_1 = \dots = X_n = Y = \mathbb{R}$, each box $w_j$ denotes multiplication by $w_j \in \mathbb{R}$ as in \eqref{eq:rmult} and $\sem{+}$ is the standard addition map above. Explicitly, this means the model has variables $\syn{X_1},\dots,\syn{X_n},\syn{Y}$, (all with representation $\mathbb{R}$) and generators given by the weights and bias: 
\[
\input{weights-bias.tikz} 
\]
The model comes with a distinguished morphism given by the diagram above, which we denote (the representation of) by $L(w,b)$. Concretely, as a function this maps each $x=(x_1,\dots,x_n) \in \mathbb{R}^n$ to the `linear' (in fact affine) combination:
\[
L(w,b)(x) = \langle x, w \rangle + b
\]
with weights \ir{$w=(w_1,\dots,w_n) \in \mathbb{R}^n$, bias $b \in \mathbb{R}$ and $ \langle \_ , \_ \rangle$ the inner product in $\mathbb{R}^n$.}\footnote{More generally, one may define \emph{multi-linear models} by a string diagram in which the inputs $X_1, \dots, X_n$ are copied to $m$ different linear models, each with a distinct \ir{output wire $Y_j$}, for $j=1,\dots,m$.}


\paragraph{Interpretation.} 
While they could be applied to uninterpreted data, or within a broader model such as a neural network, a linear model used in isolation is usually given with a complete concrete interpretation. Indeed, linear models are a canonical example of an intrinsically interpretable model. Thus the variables $\syn{X_1},\dots,\syn{X_n},\syn{Y}$ typically come with abstract interpretations as to what they represent, as well as concrete interpretations for each of their possible values in $\mathbb{R}$. Each generator $\syn{w_i}$ has an interpretation as the `strength of association' between variable $\syn{X_i}$ and $\syn{Y}$ and the generator $\syn{b}$ as a bias within the space of outputs $\syn{Y}$. The diagram \eqref{eq:linearmodel} provides an internal view of the model showing all of the usually interpreted components, with the inputs, outputs, weights and biases each appearing. 

In practice, however, note that a linear model is typically only asserted to be truly interpretable when it satisfies a sparsity requirement, meaning that only a `small' number of weights $w_j$ are non-zero.

\subsection{Rule-based models} \label{subsec:rule-model}

Aside from linear models, the classic examples of intrinsically interpretable models are \emph{rule-based} models, of which there are several variants, each admitting a natural diagrammatic description. These require the use of discrete sets, as well as continuous spaces $\mathbb{R}^n$, and so we work in the category $\Comp$.

 Within this category is the object corresponding to the set $\mathbb{B} = \{1, 0\}$ of Boolean truth values, understood as `true' and `false'. A \emph{rule} on data $X$ is then a function $R \colon X \to \mathbb{B}$, returning either true or false on each input. We can return a fixed output $r \in \mathbb{R}$ whenever the rule is `true' by composing with the function \irs{also denoted} $r \colon \mathbb{B} \to \mathbb{R}$ given by $0 \mapsto 0$ and $1 \mapsto r$, which in diagrams we denote as a box labelled $r$ (similarly to \eqref{eq:rmult}). 

One can go on to define logical operations on rules, such as conjunctions and disjunctions, in diagrams also. For example, often we may define a rule as a conjunction of conditions $c_j$, so that the rule is satisfied iff every condition is satisfied. Such a conjunction is depicted as follows:   
\[
\input{rule-alt.tikz}
\]
Here $\wedge$ is conjunction in $\mathbb{B}$, returning $1$ iff all inputs are $1$.




\paragraph{Scoring systems.}

A \emph{scoring system} is a model which, given inputs $X_1, \dots, X_n$, passes each to a number of rules $R_j$, each of which maps to a fixed score value $s_j$. \ir{These scores} are then totalled to give the final output in $Y$. 
Equivalently, the final output is an $s$-weighted linear combination of outputs from each rule, where rule $R_j$ contributes $s_j$ to the sum whenever $R_j(x) = 1$: 
\begin{equation} \label{eq:score-equation}
\mathsf{score}(x) = \sum_j s_j \cdot R_j(x)
\end{equation}
For example, in \cite{Rudin_2019_StopExplainingBlackBoxes} the following example of a score-based model is discussed as an alternative to the proprietary COMPAS deep neural network in predicting subsequent convictions in the two years following prison release.\footnote{Following the calculation of the score, the model then includes a further function converting each score to a final output probability.} 
\begin{center}
\begin{tabular}{ll}
\hline
Prior Arrests $ \geq 2$            & 1 point \\ 
Prior Arrests $\geq 5$             & 1 point  \\
Prior Arrests for Local Ordinance  & 1 point  \\
Age at Release between 18 to 24    & 1 point  \\
Age at Release $\geq 40$          & -1 point \\
\hline
\multicolumn{1}{r}{\textbf{Score}} &   = ...   
\\ \hline
\end{tabular}
\end{center}
Writing variables $P$ for the number of priors, $L$ whether the subject has a local prior ($1$ or $0$), and $A$ for the age at time of release, we can represent this as a compositional model in $\Setcat$ of the following form:
\[
\input{score-example-3.tikz}
\]
where each wire and rule box takes the semantics indicated above. 


\paragraph{Decision lists.}

A closely related variant are \emph{decision lists}, which are functions of the form `if $R_1$ return $s_1$ else if $R_2$ return $s_2$ \dots' where the $R_j$ are rules and $s_j$ their associated outputs.  Such a model is defined just like a scoring system but replacing the final addition morphism with the function `first' which returns the first of its inputs $(x_1,\dots,x_n)$ which is non-zero, and returns zero if all inputs are zero. 

Also in \cite{Rudin_2019_StopExplainingBlackBoxes} the following decision-list model is claimed to give equal performance to the COMPAS model, despite its simplicity: 
\begin{equation}  \label{eq:dlist-table}
\begin{tabular}{rcl}
\hline
IF      & age 18-20 and sex is male        & THEN predict arrest (within 2 years) \\
ELSE IF & age 21-23 and 2-3 prior offences & THEN predict arrest                  \\
ELSE IF & more than three priors           & THEN predict arrest                  \\
ELSE    & predict no arrest.               &       
\\ \hline                              
\end{tabular}
\end{equation}
We can represent this model diagrammatically as follows, where $S, A ,P$ denote sex, age and number of priors respectively, and the output $O = \mathbb{B}$ is either true (predict arrest) or false (predict no arrest). Here we have explicitly shown each rule as a conjunction.\footnote{Note that in this example all the rules return the same output value $s_j=1$ and so the `first' box can be equivalently replaced with an `or' function. } 
 \begin{equation} \label{eq:decision-list}
\input{dec-list-full.tikz}
\end{equation}

\paragraph{Decision trees.}
A final variant of rule-based models are \emph{decision trees}. Each decision tree can be thought of as a finite series of yes/no questions, where each answer leads either to an output (represented by a leaf node in the tree) or to a further question. As an overall function, such a tree is a special case of a score-based system \eqref{eq:score-equation} with a rule $R_j$ for each of its leaf nodes, true whenever all the conditions in the unique path which leads to this leaf are true. These form a partition on the inputs, so that precisely one output $s_j$ contributes to the sum on any input.




For example, consider the following decision tree, adapted from \cite{Molnar_2020_InterpretableML}, which determines an expected number of bike rentals from four possible output values $o_1,o_2,o_3,o_4$, given input data $X$ consisting of the current temperature $T$ and number of days $D$ since 2011. 

\begin{equation} \label{eq:tree-as-sd-1}
\input{tree-converted.tikz}
\end{equation}
The tree first asks whether it has been $ \leq$ 400 days since 2011. If yes,  it asks whether it has in fact been $ \leq$ 100 days 2011, and if no asks whether the temperature is $\geq 11$ degrees. Each of the four answer sets yields a disinct output $o_1,\dots,o_4$. Note that the intuitive diagram one would draw for such a tree corresponds precisely to (the lower portion of the) corresponding string diagram, read from bottom to top. \irs{At the top of the diagram, we then sum over the outputs from each leaf. As we will see, on any input only one leaf $j$ is reached, with all other leafs contributing zero to the sum, yielding $o_j$ as the final output. }

Let us now explain how such a string diagram can indeed be given a semantics which yields the input-output function of the decision tree.\footnote{\irs{To the authors' knowledge, this approach is novel here.}} In the diagram \eqref{eq:tree-as-sd-1}, $Y, N$ are merely convenient labels (for `yes', `no'), with all wires below the output boxes $o_j$ on the same variable $\syn{X^*}$ with $\sem{\syn{X^*}} = X \times \mathbb{B}$, where $X$ is the space of inputs. We view each question as a rule $Q \colon X \to \mathbb{B}$, and then each of the `$Q?$' boxes above is of the left-hand form below, from one to two copies of $X^*$, labelled $Y$ and $N$. Here $\neg$ is the `not' function which swaps $0$ and $1$. 
\begin{equation} \label{eq:q-box}
\input{C-box-yes-no-3.tikz} 
\qquad \qquad \qquad \qquad
\input{output-from-x-bool.tikz}
\end{equation}

Applied to an input $(x,b) \in X^*$, the value of $x$ is simply copied to both outputs. The $Y$ output will carry true iff the `control' input $b$  is true and $Q(x)$ is true, and $N$ will carry true iff the $b$ is true and $Q(x)$ is false. If $b$ is false both output Booleans will be false. Each output box $o_j$ is of the right-hand form above, returning its output $o_j$ if its Boolean input is true. By construction, one may verify that the overall diagram then computes the function on $X$ represented by the decision tree.

It is clear that any form of decision tree can be represented by a string diagram in just the same way. In appendix \ref{appendix:dec-tree} we give an alternate string diagram for the tree which explicitly shows the dependencies of each condition on only a subset of inputs.

\paragraph{Interpretation.}
As for linear models, rule-based models are considered standard examples of intrinsically interpretable models. Rule-based models are typically applied where all of their input and output variables come with a concrete interpretation, so that we know what each variable, and each value of each variable, represents. Moreover each generator, i.e.~each process in the diagram, has an interpretation as a simple logical function on these inputs. 
For example, in the diagram \eqref{eq:decision-list} for a decision list, each state of $A$ is interpreted as a specific age, and we can concretely interpret the box `18-20' as simply asking whether an age falls within this range. Similarly the box `400 days since 2011' is understood to ask the question of whether for this input it has been 400 days since the year 2011, and the outputs interpreted as yes or no. We note that the components by which we would typically interpret a rule-based model (e.g.~the rows in the table \eqref{eq:dlist-table}) are the same components as in their string diagram (e.g.~the boxes in \eqref{eq:decision-list}), and even the branching structure of a decision tree is apparent from its diagram \eqref{eq:tree-as-sd-1}.

\subsection{Neural networks} \label{subsec:NN-model}

A typical \emph{neuron} may be described as a compositional model in $\NN$ given by composing the output of a linear model with a (usually non-linear) fixed activation function $\sigma$, taking the form: 
\[
\input{neuron-explicit.tikz}
\]
where as for linear models each wire is represented by $\mathbb{R}$, and each $w_j$ box is a scalar multiplication by $w_j$. This implements the function 
$f(x) = \sigma ( \sum^n_{i=1} w_i x_i + b)$ on inputs $x \in \mathbb{R}^n$, with generators given by the weights $w \in \mathbb{R}^n$, bias $b$ and activation function $\sigma$. 


A feed-forward \emph{neural network}, also known as a multi-layer perceptron (MLP), is a compositional model in $\NN$ given by sequentially composing layers of such neurons, 
\ir{each layer defining a function $\mathbb{R}^n \to \mathbb{R}^m$ for some $n,m$, 
and with the generators of the compositional model} given by the union of generators for all of its neurons, i.e.~their collective set of weights, biases, and choices of activation function. Typically all neurons in a layer use the same activation function and so each such neuron is determined by its weights and bias. An entire neural network can be drawn as a single (typically huge) string diagram, in which a copy morphism passes the inputs of each layer to all of its neurons. These string diagrams are essentially equivalent to the \emph{computational graphs} commonly used to describe neural networks. 

In the examples below each grey dot is a single neuron, with its own weights and bias. The left-hand diagram shows the general shape of a many-layered network with $n$ inputs and $k$ outputs. For example, the lower input layer has $n$ inputs and $m$ outputs, with neurons $a_1,\dots,a_m$, where each $a_j$ encodes both the weights and bias for a neuron. Similarly, the upper layer has neurons $b_1, \dots, b_k$. The right-hand diagram shows a network with input size 3 and layers of size 3, 2, 3.
\[
\input{NNntom2two.tikz}
\qquad 
\qquad 
\qquad 
\qquad 
\qquad 
\input{NN323.tikz}
\]

\paragraph{Interpretation.}
Neural networks can vary in their degree of interpretability. Typically, the inputs to the network come with a concrete interpretation, related to the training data, as do the outputs of the final layer (e.g.~as a classification or piece of generated data such as text). However, variables internal to the network typically lack any interpretation. Hence we can conclude that neural networks usually lack a (complete) interpretation. 

Various XAI methods exist for analysing the internal components of neural networks to attempt to provide a given neuron or layer either with an abstract or concrete interpretation, but these methods have been subject to criticism \ir{(see Sections \ref{subsec:Issues_with_XAI} and \ref{sec:CI-and-causal-XAI})}. Despite these issues, one may use a neural network to implement a more high-level model, whose variables (e.g. the inputs and outputs of the network) do come with an interpretation. An example would be a neural implementation of a conceptual space model, such as the conceptual VAE (see Section \ref{subsec:conceptual-space-models}).

\subsection{Transformers} \label{subsec:transformer-models}

Similarly to neural networks, common architectures such as the transformer can be viewed as compositional models. For simplicity we will focus on the encoder of a transformer, but the same approach can be applied to the whole architecture from \cite{vaswani2017attention}. For a more detailed diagrammatic account of the transformer see \cite{AnatomyOfAttention}, and also \cite{elhage2021mathematical}. 

A transformer is a compositional model in $\NN$ which specifies, for each sentence length $s \in \mathbb{N}$, a diagram of the following form. 
\begin{equation} \label{eq:transformer}
\input{transformer3.tikz}
\end{equation}
Here we write $Y^s$ (resp. $f^s$) as a short-hand for drawing $s$-many parallel copies of a wire $Y$ (resp. process $f$). Within such an encoder, an input sentence $(w_1,\dots,w_s)$ of length $s$ consisting of word labels from $w_i \in L$ is converted to a vector in $X^s$ where $X=\mathbb{R}^d$ for some $d$, by first applying an embedding map $E$ and adding positional information, summing with $n$ \emph{attention heads}, and then adding to the output of an MLP. Furthermore, the model comes with variables $\syn{Q}, \syn{K}, \syn{V}$ represented as linear spaces $Q=K=\mathbb{R}^{d_1}$ and $V=\mathbb{R}^{d_2}$, for some $d_1, d_2$ such that each attention head $h_i$ takes the following form, for learned linear maps $q_i, k_i, v_i, o_i$. 
\begin{equation} \label{eq:head}
\input{head3.tikz}
\end{equation}
In the above, $\sigma$ represents a softmax, `Mult' represents the `matrix multiplication' map and $\cdot$ the map which takes all pairs of inner products, using the fact that $Q=K$. We spell these out explicitly in Appendix  \ref{appendix:linear-models}, where we also show how to represent each in elementary terms as diagrams built from our earlier addition and scalar multiplication maps \eqref{eq:linear-generators}. 


In summary, to actually specify a transformer model one must specify the set of labels $L$, (the dimensions of) spaces $X=\mathbb{R}^{d_x}$, $Q=K=\mathbb{R}^{d_1}$, $V=\mathbb{R}^{d_2}$ and the linear generators $(q_i, k_i, v_i, o_i)^n_{i=1}$, as well as the generators $E$ and $\mbox{\emph{MLP}}$ (where the latter is further specified by a feed-forward NN, decomposed as a compositional model as we saw for neural networks in section \ref{subsec:NN-model}), and along with a specific choice of positional encoding vector $\mathsf{pos}$ on $X^s$ for each sentence length $s$. From these components one defines the heads $h_i$ as in \eqref{eq:head} and for each sentence length $s$ the encoder map in \eqref{eq:transformer}, with the meaning of all other components fixed. 


\paragraph{Interpretation.} 
Transformers face the same interpretability issues as neural networks, and their lack of interpretability underlies current safety issues with large language models (LLMs) based on them. Typically, for a transformer model as above, the labelled inputs $L$ have a concrete interpretation, in that each word $w \in L$ is understood to correspond to a specific word in natural language.\footnote{In practice, a label would correspond to a \emph{token} which can in fact be a word or a word fragment, produced by a statistically driven tokenizer. Some of these tokens themselves may lack a clear interpretation.} Beyond this, the individual spaces $X, Q, K, V$ and $q, k, v$ have no fixed meaning or interpretation by default. 
Nonetheless, as for neural networks, via post-hoc analysis one may hope to understand the meaning of each space to some extent. In this case one could provide, say, the space $\syn{V}$ of a particular head with an abstract (or concrete) interpretation, bringing the model closer to an interpretable one. For example, it has been argued that an individual attention head in an NLP model may at times pick up on certain grammatical features \cite{hewitt-manning-2019-structural}. 


\subsection{RNNs and text sequence models} \label{subsec:NLP-sequence-model}

In NLP models, compositionality can arise in the composing of the meanings of words to produce meanings for a sentence or text. Perhaps the simplest form of such composition, used for example in Recurrent Neural Networks (RNNs) \cite{elman1990finding}, is given by sequentially composing the meanings of words $w_1, \dots, w_n$ to yield a composite meaning for the sequence.


\begin{definition}
Let $\lexicon$ be a lexicon consisting of a set of words, e.g.~$\natlang{Alice}, \natlang{Bob}, \natlang{plays},\dots$.  A \emph{sequence model} of $\Sigma$ is a compositional model\footnote{For a sequence model one does not require $\otimes$ and so may take $\catC, \catS$ to be plain (non-monoidal) categories with a distinguished object $I$. However, here for consistency we assume both are monoidal with $I$ as the monoidal unit.} with a single variable $\syn{X}$, and for each word $\syn{w} \in \lexicon$ a generator: 
\begin{equation} \label{eq:sequence-generator}
\input{sequence-generators.tikz} 
\end{equation}
\end{definition} 

For any such model, given a sequence of words $(\syn{w_1},\dots,\syn{w_n})$ we can define a morphism (in $\catS$) given by their sequential composite $\syn{w_n} \circ \dots \circ \syn{w_1}$. In diagrams, such a sequence appears as below, with composition being simply sequential. 
\begin{equation} \label{eq:word-sequence}
\input{word-sequence-abstract.tikz} 
\end{equation}
\irs{The model represents the sequence in $\catC$ as a composition $w_n \circ \dots \circ w_1$ of their representations $w_j = \sem{\syn{w_j}}$ in the same way.} Note that since diagrams are read bottom to top, we read the text sequence from bottom to top also. Associativity of composition in a category means that we are free (in principle) to parse a given sequence by parsing any sub-sequences and composing these together. For example, the phrase \natlang{Alice says hello} is represented equivalently by any of the following. 
\[
\input{assoc-example.tikz}
\]
This associativity is a characteristic property of sequence models, not common to all structured models in NLP.

\begin{remark}
Formally, we can consider sequence models to just use the language of plain categories, i.e.~sequential composition only. The structure category $\catS$ may be identified with $\lexicon^*$, the \emph{free monoid} generated by the symbols $\lexicon$.\footnote{Recall that a monoid is a category with a single object; equivalently it is a set $S$ along with a binary operation $a, b \mapsto a \cdot b$ satisfying associativity $(a \cdot b) \cdot c = a \cdot (b \cdot c)$ and with a unit $1$ such that $a \cdot 1  = a = 1 \cdot a$ for all $a \in S$.} Morphisms of $\catS$ thus may be equated with finite sequences $s = (\syn{w_1},\dots,\syn{w_n})$ from $\lexicon$, with composition $s_1 \cdot s_2$ given by concatenation of sequences, and with $1= ()$ being the empty sequence. 

This simple form of model is well-studied in NLP, where the structure $\lexicon^*$ is also known as a \emph{sequence algebra} and the model itself as a \emph{parser} \cite{bernardy2022unitary}. Indeed a sequence model is equivalent to specifying an object $X$ of $\catC$ and a monoid homomorphism $\lexicon^* \to \catC(X,X)$, with the latter being the usual definition of a parser. Such a morphism amounts to specifying a morphism on $X$ for each $\syn{w} \in \lexicon$, since the homomorphism property enforces that each $(\syn{w_1},\dots,\syn{w_n})$ is mapped to $\syn{w_n} \circ \dots \circ \syn{w_1}$, just as above.
\end{remark}

In practice, one often wishes to obtain a specific \emph{state} (e.g. vector) for the meaning of a word or phrase, rather than a process. By a sequence model with an \emph{initial state} we mean one with an additional generator given by a state $\star$ on $\syn{X}$. For such a model, the \emph{state representation} of a word sequence is defined as the following state:
\begin{equation} \label{eq:state-rep}
\input{RNNstateform2.tikz}
\end{equation}
Henceforth we assume all sequence models to come with such an initial state. Various forms of these models have appeared in the NLP literature, including the following well-known example.

\paragraph{RNNs.} A sequence model implemented in the category $\catC = \NN$ of neural networks is called a \emph{Recurrent Neural Network (RNN)} \cite{elman1990finding}. This means that $X=\mathbb{R}^d$ for some dimension $d$, and the model sends each word $\syn{w} \in \lexicon$ to a typically non-linear function $w \colon X \to X$. The space $X$ is referred to as the space of `hidden states' of the network. 

Equivalently the whole RNN is often viewed as a single \emph{transition function} $f \colon X \times \lexicon \to X$, where $\lexicon$ is the discrete set of the lexicon, which we may view as a morphism in $\Comp$. Each function $f(-,w)$ represents the way a hidden state is updated by the word $w$. For any sequence the resulting update is given as below.
\begin{equation} \label{eq:RNN-composite}
\input{RNN-composite.tikz}
\end{equation}
The representation of a word or sequence from an RNN is then defined by composing with a fixed initial state vector  as in \eqref{eq:state-rep}, such as $\star = 0$.

\paragraph{Unitary RNNs.} A related variant of sequence models is the \emph{Unitary RNN}, which is argued to come with several interpretability benefits in  \cite{bernardy2022unitary}. A Unitary RNN is a sequence model in which for each word $\syn{w}$ its morphism $w$ on $X=\mathbb{R}^d$ is linear, and moreover represented by an orthogonal matrix $w$, so that $w^T w = w w^T = \id{X}$. Equivalently, it may be viewed as a sequence model in the category $\Orthog$ of orthogonal matrices.\footnote{$\Orthog$ is defined just like $\MatR$ but with matrices having values in $\mathbb{R}$ and being orthogonal only.} Since orthogonal matrices are closed under composition, the process representation of each word sequence \eqref{eq:RNN-composite} is also given by an orthogonal matrix and so can be directly stored within $\mathbb{R}^{d \times d}$. The space $\Orthog(d,d)$ of $d$-dimensional orthogonal matrices also has a linear structure (forming a real Hilbert space) which is argued to allow for ready analysis. In \cite{bernardy2022unitary} it is argued that this ability to efficiently store and analyse the representation of subsequences makes the unitary RNN inherently interpretable. 

The name `unitary' in URNN refers to the fact that an orthogonal matrix is a special case of a unitary matrix between Hilbert spaces, and indeed we can consider replacing neural networks with a quantum implementation. We could define a \emph{quantum URNN} to be a sequence model of $\lexicon$ in the category $\Quant{}$, such that each $w=\sem{\syn{w}}$ \irs{is both a channel and an isomorphism, and hence unitary}. Equivalently, it consists of a Hilbert space $\hilbH$ and unitary $U(s)$ on $\hilbH$ for each $s \in \lexicon$. The model then specifies a unitary for any sequence via $U(s_1,\dots,s_n) = U(s_1) \circ \dots \circ U(s_n)$. Quantum variants of the RNN are applied to a classification problem in computational biology in \cite{london2023peptide}.



\paragraph{Further sequence models.}
Further forms of sequence model appear in the \texttt{lambeq} python toolkit for compositional NLP \cite{kartsaklis2021lambeq}. For each model, the state representation of a word sequence $(w_1,\dots,w_n)$ is shown below.

\begin{figure}[H] \label{fig:seqmodels}
	\centering
	\begin{subfigure}{0.3\textwidth}
		\centering
		\scalebox{1}{\input{cupsreader-small.tikz}}
		\caption{}
	\end{subfigure}
	\begin{subfigure}{0.3\textwidth}
		\centering
		\scalebox{1}{\input{stairs-smaller-still.tikz}}
			\caption{}
	\end{subfigure}	
	\begin{subfigure}{0.3\textwidth}
		\centering
		\scalebox{1}{\input{bagwords-simpler.tikz}}
		\caption{}
	\end{subfigure}
\end{figure}

Firstly, (a) shows a \emph{cups reader}, a model on $X=\mathbb{R}^d$ where each process representation $w$ is linear. Equivalently, this is a sequence model in the category $\catC=\FVecR$ of finite-dimensional real-vector spaces. As in this diagram, a word $w$ is however typically not treated as a morphism $X \to X$, but instead equivalently as a state of $X^* \otimes X$, by applying `caps' (which we return to in Section \ref{subsec:DisCoCat}). 

Next, (b) shows a \emph{stairs reader}, a form of sequence model in which the representation is given by repeatedly composing the state representations of each word $w$ with some (typically linear) \emph{stairs} function $g$. In fact, such a model may be seen as a special case of a \irs{(linear)} RNN; see Appendix \ref{appendix:stairs-reader}. Finally, a \emph{bag of words model} is the special case of a stairs reader where $g$ is simply given by vector addition. Then the representation of a word sequence is simply the vector sum $w_1 + \dots + w_n$, shown in (c), and in particular is commutative, ignoring word order. 

\paragraph{Interpretation.} 
A sequence model forms perhaps the simplest example of an NLP model with interesting interpretable compositional structure. Here, the object $X$ comes with only a (somewhat vague) abstract interpretation simply as a `meaning' space, and $\star$ as an `initial' or `undetermined' meaning. Each generator for a word $\syn{w}$ however has a clearer abstract interpretation, namely as (the meaning of) that word. A diagram describing a text consists only of these interpretable components,  making it an `interpreted diagram' in the sense of Section \ref{subsec:rewrite-expl}, where we will see that this enables one to give interpreted (rewrite) explanations for the output of such a model. 

Note that these benefits exist despite the absence of a concrete interpretation for the space $X$, which is typically a generic representation space. It is essentially argued in \cite{bernardy2022unitary} that those sequence models for which the space $X$ is lower dimensional, and the processes $\syn{w}$ more efficiently representable, are more interpretable.  \ir{An example is an} URNN where the representation is given by an $n \times n$ (orthogonal) matrix, for a space $X$ of dimension $n$. This however would only yield more interpretability in our sense when each dimension, and thus the space $X$,  comes with a specific (concrete) interpretation.




\subsection{DisCoCat models} \label{subsec:DisCoCat}

So far we have considered natural language simply as sequences of un-typed words. To provide a more meaningful form of composition we can introduce grammar, where words are given specific types (such as nouns or verbs) and composed via grammatical rules which determine, for example, whether a sentence is valid. A form of \emph{categorial grammar} is provided by Lambek's notion of a `pregroup' \cite{lambek2008word}. Compositional models of pregroup grammars form the heart of the \emph{DisCoCat} \st{formalism} for (Q)NLP, which we now introduce. 

Formally, a \emph{pregroup} is a partially ordered set $(G, \leq)$ with a compatible monoid structure $(1, \cdot)$ such that for each element $x$ there is a distinguished \emph{left adjoint} $x^l$ and \emph{right adjoint} $x^r$, satisfying the following: 
\begin{align} \label{eq:pregroup}
x^l \cdot x \leq 1 \leq x \cdot x^l  & & &  x \cdot x^r \leq 1 \leq  x^r \cdot x 
\end{align}

Suppose we have a set $B$ of basic grammatical types, e.g. $B=\{\n,\s\}$, for `nouns' and `sentences', respectively. This generates a free pregroup $G(B)$, whose elements are terms $t$ such as $\n, \s, \n^r \cdot \s \cdot \n^l \dots$. A \emph{pregroup grammar} is then a lexicon $\lexicon$ consisting of a set of words $\syn{w} : t$, each with a given type $t \in G(B)$, such as $\natlang{Alice} : \n$,  $\natlang{plays} : \n^l \cdot \s \cdot \n^r$, or  $\natlang{football} : \n$. A sequence of words with types $t_1,\dots,t_n$ is considered to yield a grammatical sentence when their resulting type $t = t_1 \cdot \dots \cdot t_n$ satisfies $t \leq \s$. For example \natlang{Alice plays football} yields a valid sentence since:
\begin{equation} \label{eq:Alice-plays-football}
\n \cdot (\n^r \cdot \s \cdot \n^l) \cdot \n 
\leq 1 \cdot \s \cdot 1 = \s 
\end{equation}

To give semantics to pregroups, we must consider their generalisation to a special form of category.\footnote{A pregroup is precisely a rigid monoidal category which forms a partially ordered set, i.e. with at most a unique morphism $x \to y$ for each pair of objects $x, y$, denoted $\leq$ when it exists, and such that $x \leq y \leq x \implies x = y$. In this setting the existence of cups and caps correspond to the equations of pregroups \eqref{eq:pregroup}.} A monoidal category $\catC$ is \emph{rigid} when every object $A$ comes with a \emph{left adjoint} $A^l$ and a \emph{right adjoint} $A^r$, each respectively equipped with distinguished states and effects depicted as \emph{cups} and \emph{caps} as below, satisfying the so-called `yanking equations':\footnote{Additionally there are equations expressing $\id{A^l}$ and $\id{A^r}$ by yanking similarly.}
\[
\input{yanking-shorter.tikz}
\]

Any pregroup grammar $\lexicon$ over a free pregroup $G(B)$ generates a free rigid monoidal category $\FreeRigid(\lexicon)$ with $G(B)$ as its objects. For types $t, t'$ we have $t \otimes t' = t \cdot t'$, and for any type $t \in G(B)$, the elements $t^l, t^r$ form the left and right adjoints to $t$ as expected. Along with these cups and caps, for each word, type pair $\syn{w} : t$ in $\lexicon$ there is a generator given by a state $\syn{w}$ of $t$, such as: 
\[
\input{Alice-word-states.tikz}
\]
Morphisms in this category can describe pregroup derivations of sentences (and other types). For example, a pregroup derivation of the sentence `Alice plays football' would appear as:
\begin{equation} \label{eq:aliceplaysfootball}
\input{alice-plays-football.tikz}
\end{equation}
Any such diagram for a sentence can be read as a lower part consisting of the words, and an upper part encoding the grammatical structure, as a reduction of types $t \leq \s$ in the pregroup (in this example, precisely the derivation \eqref{eq:Alice-plays-football}). 

Now we have seen how to describe pregroup grammars in terms of a free structure category, we can consider models which give semantics to these words and derivations.   

\begin{definition} \cite{coecke2010mathematical} \label{def:DCatmodel}
Let $\lexicon$ be a pregroup grammar over some set of basic types $B$. A \emph{DisCoCat model} of $\lexicon$ is a compositional model of $\catS = \FreeRigid(\lexicon)$ in a rigid monoidal category $\catC$. 
\end{definition}

Explicitly then, the structure category contains diagrams such as \eqref{eq:aliceplaysfootball}. The model amounts to specifying in $\catC$ a representation for each basic type, e.g.~$\sem{n} = \syn{n}$, and a state $w = \sem{\syn{w}}$ of $t$ for each word of this type, e.g.~a state $\natlang{likes}$ of $n^r \otimes s \otimes n^l$.  Since $\sem{-}$ is a strong monoidal functor it automatically preserves adjoint objects, cups and caps, and so the representation of a sentence diagram such as \eqref{eq:aliceplaysfootball} is given by `the same diagram' in $\catC$.\footnote{That is, by composing the word representations with the caps in $\catC$ in the same way.}





\begin{example} \label{ex:dcat-vec}
A commonly used semantics category for DisCoCat models is the category $\catC=\FVecR$ of finite-dimensional real vector spaces and linear maps $f \colon V \to W$. Here $\otimes$ is the tensor product, with $I=\mathbb{R}$ the one-dimensional space. Each state $v$ of $V$ may be equated with a vector $v=v(1) \in V$. 

This category is \emph{compact}, meaning it is symmetric monoidal and that left and right adjoints coincide. For a space $V$ we have that $V^l = V^r = \FVecR(V, \mathbb{R})$, the \emph{dual space} of $V$. Given a choice of basis $\{\ket{i}\}^n_{i=1}$ for $V$ we can further equate the dual space with $V$ itself, thanks to finite dimensionality. Each cup is given by the state $\sum ^n_{i=1} \ket{i} \otimes \ket{i}$ of $V \otimes V$, while the caps are: 
\[
\begin{tikzpicture}
	\begin{pgfonlayer}{nodelayer}
		\node [style=none] (0) at (1, -0.25) {};
		\node [style=none] (1) at (-1, -0.25) {};
		\node [style=label] (2) at (1, -0.75) {$V$};
		\node [style=label] (3) at (-1, -0.75) {$V$};
	\end{pgfonlayer}
	\begin{pgfonlayer}{edgelayer}
		\draw [bend right=90, looseness=1.75] (0.center) to (1.center);
	\end{pgfonlayer}
\end{tikzpicture}
 \ \ :: \ \ \ket{i} \otimes \ket{j} \mapsto
\begin{cases}
1 & i = j \\ 
0 & \text{otherwise}
\end{cases}
\]

In such a DisCoCat model, a word such as \natlang{Plays} corresponds to a vector of the tensor product $n \otimes s \otimes n$, also simply called a \emph{tensor}. For example, the sentence \eqref{eq:aliceplaysfootball} is mapped to a vector of $s$  given by the tensor contraction:
\[
\ket{\natlang{Alice plays football}}_k 
\ = \  
\sum^n_{i=1} \sum^n_{j=1}
\ket{\natlang{Alice}}_i \ket{\natlang{plays}}_{i,k,j}
\ket{\natlang{football}}_j
\]

It is also common to take DisCoCat semantics in the compact category $\FHilb$ of finite-dimensional (complex) Hilbert spaces and linear maps $f \colon \hilbH \to \hilbK$. Again $\otimes$ is the tensor product, now with $I=\mathbb{C}$, and cups and caps are defined as above only now using the complex dual space of linear maps $\hilbH \to \mathbb{C}$. This latter category provides the semantics for quantum models of DisCoCat, as explored for example \rl{in \cite{meichanetzidis2023grammar, lorenz2023qnlp}}. 
\end{example}

\begin{example} \label{ex:dcat-consp}
Further choices for the semantics category include compact categories of relations, such as the category $\ConvRel$ of convex sets and convex relations. In \cite{bolt2019interacting} this is used to give a DisCoCat model of the conceptual spaces framework, which we return to in Example \ref{ex:consp-conrel}. Here each noun wire factorises in terms of domains such as \dc{colour} or \dc{taste}, and the meaning of a sentence is given by relational composition.
\end{example}

Various additions to \st{DisCoCat models} have been explored, including the modelling of relative pronouns through the use of \emph{Frobenius structures}, which may be thought of as `copying information' \cite{sadrzadeh2013frobenius}. These satisfy various equations which allow for simplification of diagrams, such as:
\begin{equation} \label{eq:dcat-frob-fruit}
\input{fruit-bitter.tikz}
\end{equation}
More recently, a \emph{higher-order} variant of DisCoCat has been introduced, which uses a free cartesian closed (rather than merely rigid) category as both its structure and target category, which describes higher-order operations on diagrams in the language of closed categories \cite{toumi2023higher}.

\paragraph{Interpretation.}
In a DisCoCat model, the variables $\syn{t}$ for each pregroup type have an abstract interpretation as representing that type; for example $\syn{n}$ corresponds to nouns. Similarly the generator $\syn{w}$ for each word has an abstract interpretation as representing (the meaning of) that word. Since the `caps' encode grammatical relationships, we can interpret every component of a diagram for a sentence such as \eqref{eq:aliceplaysfootball} (making it an interpreted diagram in the sense of Section \ref{subsec:rewrite-expl}), which we can say has an interpretation itself as the meaning of the sentence, with this specific grammatical structure. The fact that we can see how the meaning of a sentence is constructed grammatically, may provide additional explanatory benefits. For example, the sentence in \eqref{eq:dcat-frob-fruit} using Frobenius structures reduces to the readily inspectable composite form on the RHS. 
Specific DisCoCat models may have their interpretability enhanced by furthermore coming with concrete interpretations; for example the conceptual space models in Example \ref{ex:dcat-consp} are such that each noun space $\n$ has a concrete interpretation via a conceptual space factored in terms of colour, taste and texture.


\paragraph{CCG Models.}
Beyond pregroup grammars, one may define compositional models based on further forms of categorial grammar, notably Combinatory Categorial Grammar (CCG) \cite{steedman:2000}. We describe CCG models in detail in Appendix \ref{app:CCG-model}. In fact they are closely related to DisCoCat models; see Appendix \ref{appendix:disco-to-ccg}.

\subsection{DisCoCirc models} \label{subsec:DisCoCirc}

An alternative form of compositional structure in NLP is offered by the more recently developed \st{\emph{DisCoCirc} models}, originally from \cite{coecke2021mathematics} and treated more fully in \cite{wang2023distilling}. In this approach, the semantic space now decomposes in terms of multiple wires corresponding to the relevant nouns or \emph{discourse referents} in a text \cite{kamp1993discourse}. The text itself, including its grammatical structure, is encoded in the shape of a string diagram describing a process on these wires. For example, a sentence such as \natlang{Alice laughs and sees Bob who likes Charlie} could be modelled as a text circuit of the following form, acting on wires representing Alice, Bob and Charlie. 
\[ 
\input{text-circuit-example2.tikz}
\]

The DisCoCirc framework allows for a still richer class of string diagrams encoding more features of language and grammatical types. In particular it makes use of `higher order' processes which we can depict as a `box with a hole' into which we may insert a diagram to be modified. An example is when a verb is modified by inserting it into a higher order box representing an adverb. This allows us to produce diagrams known as \emph{text circuits}, such as the following representation of the sentence \natlang{Alice who is sober, sees Bob, who is drunk, dance clumsily and laughs at him} (from \cite{wang2023distilling}). 
\begin{equation} \label{eq:drunkbob}
\input{drunkbob3.tikz}
\end{equation}
\st{These diagrams} naturally allow the treatment of entire texts, rather than single sentences, and require only the structure of a monoidal category, on which such `higher order processes' may be defined. 

\paragraph{Text circuits.}
Let us now introduce \st{text circuits} in more detail. Let $\lexicon$ be a set of typed words, where the pair of a word $\syn{w}$ with a type $\syn{t}$ is denoted $\syn{w} : \syn{t}$. We will now list the small set of example types from \cite{wang2023distilling}, motivated by the types from a CCG grammar, along with their graphical representation.\footnote{For a practical NLP application a larger set of types would be required for the particular language in question.} A text circuit is a string diagram with $n$ input wires and $n$ output wires, which ultimately will each be labelled with distinct discourse referents from the text. 

Adjectives (e.g. \natlang{drunk} : \dc{ADJ}) and intransitive verbs (e.g. \natlang{laughs} : \dc{IV}) appear as boxes modifying a single noun wire, while transitive verbs act on two wires representing their subject and object. Thus for each adjective, intransitive verb or transitive verb in $\lexicon$ we have a graphical box of the corresponding form below. 
\begin{equation} \label{eq:dcirc-plaingates}
\input{adj.tikz}
\end{equation}
Adverbs (e.g. \natlang{quickly} : \dc{ADV}), and other higher order types, are represented graphically as boxes with a hole into which we can insert a text circuit of the appropriate shape, to build a new text circuit, for example: 
\[
\input{runquickly2.tikz}
\]
The various word types represented as higher-order maps are shown in \eqref{eq:dcirc-all}. For each adverb in $\lexicon$ we have a box for each of the two forms shown, which act on intransitive or transitive verbs respectively.
 Next, for each adposition (e.g.~\natlang{at} : \dc{ADP}) we have a box as shown, with a hole into which a verb is inserted and which returns a verb along with an additional noun.

Next are verbs with a sentential complement. For each verb  v : \dc{SC.V}, we have a generator as depicted below. For example, for \natlang{sees} : \dc{SC.V} we insert a text circuit for the sentence of what is seen, introducing an extra noun wire, as in \natlang{Alice sees Bob dance}. Finally, conjunctions allow us to combine pieces of text which overlap in some fragment of their nouns. For each conjunction we have an infinite family of generators of text circuits of the right-hand form below, for each possible circuit shape of its two arguments. 
\begin{equation} \label{eq:dcirc-all}
\input{adverb1.tikz} \qquad \ \ 
\input{adverb-smaller.tikz} \qquad \ \ 
\input{adp.tikz} \qquad \ \ 
\input{scv1.tikz} \qquad  \ \ 
\input{scv2.tikz} \qquad \ \  
\end{equation}

Now for each $n \in \mathbb{N}$ let $\TextCircuits_\lexicon(n)$ denote the collection of string diagrams that can be built in finitely many steps by composing and inserting these generators into holes, leaving no open holes remaining. Text circuits can be composed in sequence and parallel and so these form the morphisms $n \to n$ of an SMC $\TextCircuits_\lexicon$. 

We formalise this notion of `box with holes' more thoroughly in Appendix \ref{app:dcirc-hocat}. Roughly, by a `higher order morphism' on a category we mean a function which produces a new morphism from a set of input morphisms, each of fixed types. By a \emph{h.o.-category} we mean a monoidal category $\catC$ with a distinguished set $H_\catC$ of higher order morphisms. As expected, $\TextCircuits_\lexicon$ forms a h.o.-category with the generators \eqref{eq:dcirc-all} as its distinguished higher order morphisms. Finally, a `h.o.-functor' is a monoidal functor $F \colon \catC \to \catD$ between h.o.-categories along with a compatible mapping $H_F \colon H_\catC \to H_\catD$ between their higher order morphisms. 

\begin{definition} \label{Def:dcirc-model}
A \emph{DisCoCirc model} of lexicon $\lexicon$ is a compositional model of $\catS=\TextCircuits_\lexicon$ in a h.o.-category $\catC$, where $\sem{-}$ is a h.o.-functor.
\end{definition}
Equivalently, a DisCoCirc model amounts to specifying a single object $n = \sem{\n}$ in $\catC$ (for `noun') along with morphisms as in \eqref{eq:dcirc-plaingates} for each word $w$ of type $\dc{ADJ}, \dc{IV}, \dc{TV}$, and higher order morphisms as in \eqref{eq:dcirc-all} on $\catC$ for each possible hole shape and each word $w$ of type $\dc{ADV}, \dc{ADP}, \dc{SC.V}$ or \dc{CNJ}, as appropriate.  

\begin{remark}[Labelled text circuits]
The wires in actual text circuits should ultimately be labelled with nouns (or noun phrases) corresponding to the discourse referents in a text. Given a set of possible \emph{labels} $L$, e.g. $\natlang{Alice},\natlang{Bob},\natlang{Charlie},\dots $, we write $\TextCircuits_{\lexicon,L}$ for the family of labelled text circuits, where now each wire is given a distinct label from $L$, such that the inputs and outputs to every circuit are identical (and by convention aligned) as in \eqref{eq:drunkbob}. Any DisCoCirc model of $\lexicon$ extends, by simply `forgetting' wire labels, to give a h.o-functor $\TextCircuits_{\lexicon,L} \to \catC$, which we may call a \emph{labelled DisCoCirc model}. By construction, the semantics it provides to any text circuit does not depend on the choice of labels on its wires. 
\end{remark}

We will meet several examples of (labelled) text circuits in DisCoCirc in the context of explainability later in Section \ref{sec:expl-from-diags}.

\paragraph{Implementations.}
The theory of text circuits requires the semantics category $\catC$ to only be a symmetric monoidal category, allowing us to define DisCoCirc models very broadly. In particular, the same text circuit could be implemented either via neural networks, or quantum circuits, as depicted in Figure \ref{fig:disco-circ-quant-nn}. Taking $\catC=\NN$ gives neural DisCoCirc models, in which a wire such as \dc{Alice} is represented by a space $\mathbb{R}^n$ and each basic gate such as \dc{plays} by a neural network. Taking $\catC=\Quant{}$ instead gives quantum DisCoCirc models, in which the wires are quantum systems, such as collections of qubits, and the basic gates are CP maps (e.g. unitary gates). In \cite{QDisCoCirc} such a quantum DisCoCirc model is trained, with its interpretabiliy benefits discussed, and it is shown that the resulting quantum text circuits are hard to simulate classically. 

\paragraph{Interpretation.}
DisCoCirc models come with a natural abstract interpretation for their components.  For any labelled DisCoCirc model, there is a variable for each possible wire label, corresponding to a noun representing this discourse referent in the text. This variable has an abstract interpretation as representing the meaning state of this noun. The (higher order) generator morphism for each typed word $\syn{w}$ has an abstract interpretation as the meaning of that word, or rather the process of how it updates the meanings of its input words. 

As well as this, DisCoCirc models come with a rich compositional structure from the nature in which text circuits are constructed, combined with the fact that each of their components has an abstract interpretation as described above. In Section \ref{sec:expl-from-diags} we will see that this provides various explainability benefits, without requiring a concrete interpretation. A specific DisCoCirc model may come with further interpretability still by giving its noun space a concrete interpretation; for example by making it an interpreted conceptual space similarly to Example \ref{ex:dcat-consp}.

\subsection{Conceptual space models} \label{subsec:conceptual-space-models}

Our next form of model takes its inspiration from cognitive science. \Gardenfors' \emph{conceptual spaces} provide a framework for cognition in humans and AI, in which a cognitive space is viewed as a convex space \cite{gardenfors2004conceptual,gardenfors2014geometry}, which factorises in terms of elementary simpler spaces known as \emph{domains}, such as \domain{colour}, \domain{taste}, or \domain{sound}. Each such space (or domain) comes with a collection of concepts, such as \concept{red}, \concept{salty}, or \concept{loud}, modelled as convex subsets. 
Various formalisations and generalisations of the approach have been given, including compositional treatments based on string diagrams in \cite{bolt2019interacting,tull2021categorical,QonceptsFull2024}, which we give a simplied view of here.\footnote{In \cite{QonceptsFull2024} a more detailed notion of this model is introduced, which for brevity we have simplified, which makes use of `projection' morphisms which allow instances to be defined intrinsically and each conceptual space to form only a `subspace' of the product of domains, i.e. the domain of a projection.}

A \emph{conceptual space model} is a compositional model with a variable $\syn{Z}$, called the \emph{conceptual space}, and variables $\syn{Z_1},\dots,\syn{Z_n}$ called the \emph{domains}, along with the equation $\syn{Z} = \syn{Z_1} \otimes \dots \otimes \syn{Z_n}$. Additionally, it comes with a collection of generators, called the \emph{concepts}, of the form: 
\begin{equation} \label{eq:concept}
\input{concept-local.tikz}
\end{equation}
where $\dom(\syn{C}) \subseteq \{\syn{Z_1},\dots,\syn{Z_n}\}$ is a subset of domains. For example, a domain $\syn{Z_j}$ interpreted as \domain{colour} may have a concept \concept{red}; more generally a concept can be defined over multiple domains.

We can extend any such concept to the whole of $\syn{Z}$ by discarding the remaining unused domains. More generally, given any collection $(\syn{C_1},\dots,\syn{C_k})$ of concepts for which the subsets $\dom(\syn{C_i})$ are all disjoint, we define a product concept $\syn{C}= (\syn{C_1},\dots,\syn{C_k})$ over all of $\syn{Z}$ as follows:
\begin{equation} \label{eq:prod-concept}
\input{prod-concept3.tikz} 
\end{equation}
where $\syn{Z}'$ consists of all remaining domains. For example, we may have a conceptual space $Z$ for images of simple 2d shapes, with domains \domain{colour}, \domain{size}, \domain{shape}, \domain{position}. Given concepts \concept{red}, \concept{large}, \concept{square} on the first three domains, we can define a concept \concept{large red square} via their product (applying discarding to the unused \domain{position} domain). 

In addition to these generators, any specific conceptual space model should come with a distinguished collection of normalised states $\psi \in \Inst(Z_i)$ on each domain $Z_i=\sem{\syn{Z_i}}$, called the \emph{instances} of the domain \cite{clark2021formalising}. We then define an instance $\psi \in \Inst(Z)$ of the overall space $Z$ to be a product of instances $\psi_i \in \Inst(Z_i)$ 
on each factor, i.e. a state of the form: 
\[ 
\input{prod-instance.tikz} 
\]
Given any such instance, and any concept $\syn{C}$ on $\syn{Z}$, we can compose its representation with the instance to yield a scalar $C \circ \psi$ in $\catC$, which we think of as representing `how well' the instance fits the concept:
\begin{equation} \label{eq:instance}
\begin{tikzpicture}
	\begin{pgfonlayer}{nodelayer}
		\node [style=map] (0) at (0, 1) {$C$};
		\node [style=none] (1) at (0, -0.75) {};
		\node [style=map] (2) at (0, -0.75) {$\psi$};
		\node [style=label] (3) at (0.75, 0) {$Z$};
	\end{pgfonlayer}
	\begin{pgfonlayer}{edgelayer}
		\draw (1.center) to (0);
	\end{pgfonlayer}
\end{tikzpicture}
 
\end{equation}
Depending on the semantics category, this scalar can take values in $\{0,1\}$, for simply `yes' or `no', or $[0,1]$ or $\mathbb{R}^+$ to model `fuzzy' or `graded' concepts. 

Beyond this \st{simple treatment here, one can imagine enriching the setup further to include processes relating domains and representing conceptual reasoning, as suggested in \cite{tull2021categorical}. }

\begin{example} \label{ex:consp-conrel}
In \cite{bolt2019interacting} conceptual spaces are studied via the category $\catC=\ConvRel$ of \emph{convex relations}. Here the objects are convex sets, i.e. sets equipped with operations $(x,y) \mapsto p \cdot x + (1-p) \cdot y$ for $p \in [0,1]$, with $A \otimes B = A \times B$ and $I$ a singleton set. Morphisms $R \colon A \to B$ are convex relations, i.e. subsets $R \subseteq A \times B$ which are convex, meaning they are closed under convex combinations. In particular, effects, i.e.~concepts as in \eqref{eq:concept}, correspond to convex subsets of a product of domains, just as concepts are defined by \Gardenfors{} \cite{gardenfors2004conceptual}. 

The authors give examples of the domains $Z_1 = \domain{taste}$, $Z_2 = \domain{colour}$ and $Z_3 = \domain{texture}$. The domain $\domain{taste}$ is represented by a simplex, and $\domain{colour}$ by an RGB colour cube $[0,1]^3$, as below. The right-hand image shows an example concept \concept{yellow} on the colour domain. A toy example of a conceptual space $Z$ for \domain{food} is then defined as a product of these domains. 
\begin{figure}[H]
	\centering
\begin{subfigure}{5cm}
		\centering
\includegraphics[scale=0.15]{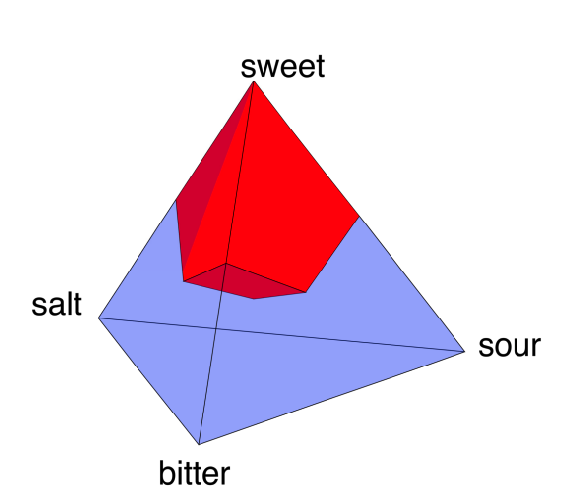}
	\end{subfigure}
\begin{subfigure}{5cm}
		\centering
\includegraphics[scale=0.15]{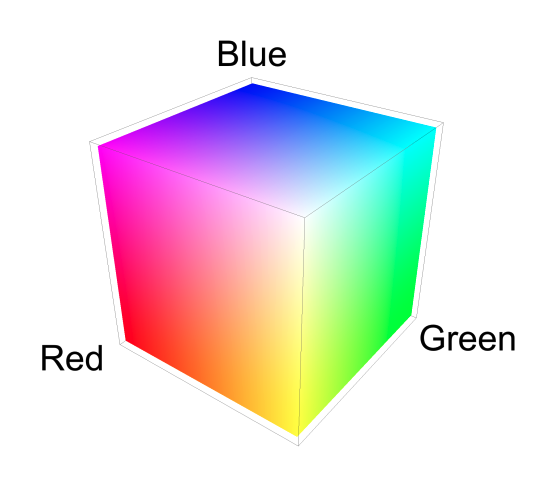}
	\end{subfigure}
\begin{subfigure}{5cm}
		\centering
\includegraphics[scale=0.15]{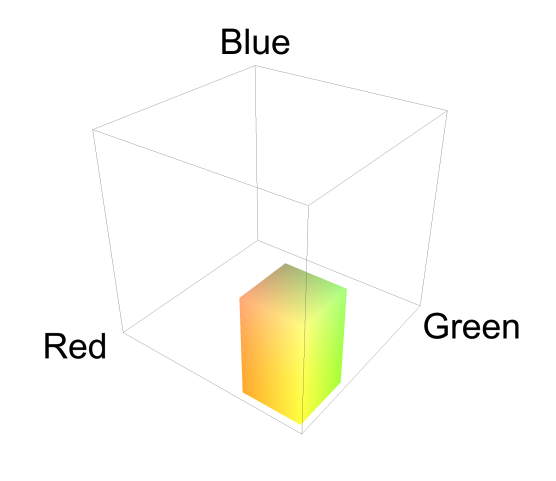}
	\end{subfigure}
\caption{A \domain{taste} domain with concept \concept{sweet} highlighted, a \domain{colour} domain, and a concept \concept{yellow}. Images from \cite{bolt2019interacting}.}
\end{figure}

Instances on these domains correspond simply to elements of their underlying sets; for example there is an instance of \domain{colour} for each RGB value in the colour cube. Thus an instance of a food space is given by a triple of a particular taste, colour, and texture. The scalar obtained by applying a concept to an instance is $1$ if the instance belongs to the subset or $0$ otherwise.

In \cite{tull2021categorical} an alternative semantics is given which allows \emph{fuzzy} concepts $C \colon Z \to [0,1]$ which now map instances to values in $[0,1]$. For this we use the category $\catC = \ConSp$ of (measurable) convex spaces and \emph{log-concave} probability channels. In particular, concepts \eqref{eq:concept} are now log-concave measurable functions $Z \to [0,1]$. 

Finally, in \cite{QonceptsFull2024} \emph{quantum conceptual models} are explored, by taking semantics in the category $\catC=\Quant{}$ of quantum processes. Now each domain is associated with a Hilbert space $Z_i = \hilbH_i$, and a conceptual space with their tensor product $\hilbH_1 \otimes \dots \otimes \hilbH_n$. Instances are defined as non-entangled pure states $\psi_1 \otimes \dots \otimes \psi_n$. By definition concepts are then positive operators $C$ on $\hilbH$, with $C \circ \psi = \Tr(C \ket{\psi_1} \otimes \dots \otimes \ket{\psi_n})$.  
\end{example}

\paragraph{Interpretation.}
Conceptual space models are typically applied with a clear abstract interpretation for each of their components. Each  $\syn{Z_i}$ is abstractly interpreted as a given cognitive domain (e.g.~\domain{colour}), and the overall space $\syn{Z}$ as a semantic space related to some class of entity (e.g.~\domain{food}). Each generator $\syn{C}$ is interpreted as a concept related to the domains on which it is relevant. 
These interpretations of concepts allow them to be related to training data, by labelling data points with appropriate concepts. Indeed training a conceptual space model on a latent space $Z$ can be seen as a way to partially interpret it; even if each point of $Z$ has no direct interpretation, we can understand it to some extent by saying how well it applies to each (interpreted) concept. Equipping latent spaces with conceptual structure in this way can make them more interpretable, a benefit of (classical and quantum) approaches such as in \cite{QonceptsFull2024}. However, in conceptual space models from cognitive science each domain usually does come with a concrete interpretation, allowing us to interpret each of its instances also. For example each instance of the domain \domain{colour} would be interpreted as a specific colour value.

\paragraph{Applying conceptual spaces to data.} Using conceptual space models in practice requires the training of an encoder $e \colon X \to Z$ from a space of inputs $X$ into a latent conceptual space $Z$, along with a set $L$ of concept labels applied to training data.  A quantum version of this setup is explored in  \cite{QonceptsFull2024}. Additionally, a decoder $d \colon Z \to X$ can allow concepts to be used as states on $Z$ to generate new inputs in $X$, as in the Conceptual VAE also defined in \cite{QonceptsFull2024}. For details of such conceptual encoder-decoder models, see Appendix \ref{subsec:concept-encoders}.

\paragraph{Disentangled representations.}
The factorisation of a conceptual space as a product of domains is closely related to the notion of a `disentangled representation' in structured ML, as formalised in \cite{higgins2018towards}. We discuss such representations as compositional models in Appendix \ref{app:disent-reps}.

\subsection{Causal models} \label{subsec:causal-models}

One class of model that plays an important role both in AI and interpretability is the class of causal models. The framework of causal models has been developed over the past four decades, originally with a perspective on AI that predates ML and partly comes from the philosophy of causation. A main goal has been to devise a framework that formalises and guides causal reasoning in a principled and scientific way, delineating it from mere Bayesian reasoning with statistical models.  
Pioneered by Pearl and his collaborators \cite{pearl2009causality}, as well as Spirtes, Glymour and Scheines \cite{SpirtesEtAL_2000_BookCausationPredictionSearch}, there is now a rich framework and a vast literature.

In recent years, it has also gained traction in NN-based ML and a number of authors have argued that the traditional statistical models learned by AI are not sufficient to carry out meaningful `human-like' reasoning, necessitating ML models equipped with explicit causal structure. Such causal ML systems are argued to gain robustness to adversarial attacks, greater generalisation skills and more efficient training, as well as increased interpretability \cite{scholkopf2022causality,higgins2018towards}. 
In fact, concepts and terminology from the causal model literature feature prominently in standard XAI methods.

Later in Section~\ref{sec:CI-and-causal-XAI} we will discuss the various roles causality has to play in interpretability. Here we will first introduce the formal notion of a causal model as a compositional model, and give a short overview of the diagrammatic account of the causal model framework from \cite{lorenz2023causal}, which builds on the previous works \cite{fong2013causal,jacobs2019causal,fritz2023d}. 




\subsubsection{Causal Bayesian networks} \label{subsubsec:CBNs}

Causal models are typically introduced through the notion of a \emph{Causal Bayesian Network} (CBN). 
This consists of a finite set of variables $V=\{\syn{X_1},\dots\syn{X_n}\}$, which form the vertices of a \emph{Directed Acyclic Graph} (DAG) $G$ representing the direct-cause relations. 
Each variable $\syn{X_i}$ takes a (finite) set of values $X_i$ and is associated with a probability channel $P(X_i \mid \Pa(X_i))$, its \emph{causal mechanism}, describing the probability distributions of its values given those of its \emph{parents} $\Pa(X_i)$ in $G$. 
The family of mechanisms induce a joint distribution over $V$, which, by construction, satisfies the causal Markov condition: 
$P(V) = \prod^n_{i=1} P(X_i \mid \Pa(X_i))$.\footnote{Note that the $P(X_i \mid \Pa(X_i))$ are given stochastic maps. As the notation suggests, they may in practice often be computed as conditionals from a joint distribution; however this need not be the case. Going between a joint distribution $P$ (plus DAG) and the causal mechanisms $P(X_i \mid \Pa(X_i))$ is only uniquely possible in both directions assuming full support of $P$. The definitional primitives of causal models are the causal mechanisms $P(X_i \mid \Pa(X_i))$.} 
In addition, it is common for only a subset $O \subseteq V$ of the variables in a model to be considered as \emph{outputs}, which we think of as `observed' variables, with the remaining variables referred to as `hidden' or `latent'. 

\begin{example} \label{Ex_CBN} 
	Let $S$ be a person's choice to smoke, $L$ whether or not they develop lung cancer, $A$ their age and $B$ a set of relevant background conditions like socio-economic status, education etc. 
A plausible causal model consists of the DAG below, where the vertices corresponding to output variables are circled, along with the specification of each of the probability channels listed to the right, which give the mechanisms.
	\begin{eqnarray}
		\input{Fig_Example_CBN_DAG.tikz} 
		\begin{minipage}{5cm}
			\centering
			$P(L | S, B, A)$ \\[0.3cm]
			$P(S | B)$ \\[0.3cm]
			$P(B | A)$ \\[0.3cm]
			$P(A)$	
		\end{minipage}
	\end{eqnarray}
	This data defines a marginal distribution over the outputs: $P(S, L, A) = \sum_{B} P(L | S, B, A) P(S | B) P(B | A) P(A)$.
\end{example}
 
It turns out that such a model can be understood as a compositional model induced by a single diagram. This is based on a correspondence between DAGs and the following class of string diagrams. 

 \begin{definition} \label{def:network-diagram}
A \emph{network diagram} is a string diagram $D$ built from single-output boxes, copy maps and discarding effects:
\[
\input{nd-box.tikz}
\qquad 
\input{nd-copy.tikz}
\qquad 
\begin{tikzpicture}[tikzfig]
	\begin{pgfonlayer}{nodelayer}
		\node [style=upground] (0) at (1.5, 0) {};
		\node [style=none] (1) at (1.5, -0.75) {};
	\end{pgfonlayer}
	\begin{pgfonlayer}{edgelayer}
		\draw (1.center) to (0);
	\end{pgfonlayer}
\end{tikzpicture}

\] 
along with labellings on the wires, such that any wires not connected by a sequence of copy maps are given distinct labels, and each label appears as an output at most once and as an input to any given box at most once.
\end{definition}

Examples of such diagrams will follow shortly in Ex.~\ref{ex:smoking-causal-model} and \ref{ex:open-DAGs}. 
For $G$ a DAG with vertices $V$ and $O \subseteq V$ a specified subset of output vertices, the pair $(G,O)$ is equivalent to a network diagram over wires $V$ with outputs $O$, and no inputs (see Ex.~\ref{ex:smoking-causal-model}). 
The correspondence works as follows: each vertex $X \in V$ is associated with a single-output box with $|\Pa(X)|$ many input wires and an output wire labeled by $X$. This wire is followed by a copy map that connects it with the boxes associated with $X$'s children $\Ch(X)$, and which also copies the wire one more time if it is an output, $X \in O$. 
More broadly, a general network diagram, which may have inputs, is equivalent to an \emph{open DAG} $(G,I,O)$, which now may come with a subset $I \subseteq V$ of \emph{inputs}, such that each input has no parents in the DAG. 

While a network diagram thus specifies a causal structure, an actual causal model also specifies causal mechanisms, by modelling the diagram in some suitable cd-category $\catC$, as follows. 

\begin{definition} \cite{lorenz2023causal} \label{def:causal model}
An \emph{open causal model} (resp. causal model) $\modelM$ in a cd-category $\catC$ is a compositional model of a network diagram $\diagD_\modelM$ in $\catC_\channel$ (resp.~with no inputs). We call the network diagram $\diagD_\modelM$, or the equivalent open DAG $G_\modelM = (G,I,O)$, the \emph{causal structure} of the model, and the variables which form the inputs (resp. outputs) of $\diagD_\modelM$ the inputs (resp. outputs) of the model. We call each generator of the model a \emph{(causal) mechanism}. 
\end{definition}

A causal model $\modelM$ in $\FStoch$ is hence equivalent to a CBN. The induced state in $\FStoch$ over all variables is equal to the distribution $P(V) = \prod^n_{i=1} P(X_i \mid \Pa(X_i))$, and $\morfrom{\modelM}$ to the marginal $P(O)$ over outputs.

By definition, a causal model amounts to a network diagram $\diagD$ and cd-functor $\Free(\Sig_\diagD) \to \catC_\channel$. This functorial view was noted by Jacobs, Kissinger and Zanasi in \cite{jacobs2019causal}, building on Fong's account of Bayesian networks \cite{fong2013causal}. 

\begin{example} \label{ex:smoking-causal-model}
For the DAG $G$ from example \eqref{Ex_CBN} the equivalent network diagram $\diagD_G$ is shown below. 
\[
\input{causal-model-example.tikz} 
\]
A corresponding causal model is given by a model of this diagram  in a cd-category $\catC$. Thus such a model has variables $\syn{A}, \syn{B}, \syn{L}, \syn{S}$ and generators $\syn{c_A}, \syn{c_B}, \syn{c_S}, \syn{c_L}$, represented  by corresponding objects $A, B, S, L$ and channels $c_L, c_S, c_B, c_A$ in $\catC$ as above.  For $\catC = \FStoch$ this is a CBN as in Ex.~\eqref{Ex_CBN}. 
\end{example}

\begin{example} \label{ex:open-DAGs}
The diagram below depicts an open DAG over $V = \{X_1,\dots,X_5\}$ with inputs $I = \{X_2,X_3\}$ and outputs $O=\{X_3, X_5\}$, as well as the equivalent network diagram. 
An open causal model $\modelM$ of this form in $\catC$ is given by specifying objects $X_1,\dots,X_5$ along with channels $a, b, c$ of the corresponding type.
\[
\input{open-DAG-rotate3.tikz} \qquad \iff \qquad 
\input{open-DAG-sd3.tikz}
\]
\end{example}

Open causal models with inputs, while not as common in the traditional causal model literature, arise naturally when studying neural networks, as well as interventions on models, as we will see later.

\paragraph{Interventions.}
A CBN is in particular a Bayesian network, encoding conditional independence relations in the probability distribution over all variables (see Sec.~\ref{sec:frameworks} for more details). What distinguishes a \emph{causal} Bayesian network, and causal models more generally, are \emph{interventions}. An intervention is thought of as a `reaching in' that changes the causal influences by altering some of the mechanisms of a model. The ability to do so is justified by the assertion that the channels $c_X \colon \Pa(X) \to X$ are causal mechanisms, and so in particular have the property of `autonomy'. Here we briefly sketch the string diagrammatic view of interventions, referring the reader to \cite{lorenz2023causal} for more details.

The simplest kind of intervention most commonly discussed is that of a \emph{do-intervention}.  
A do-intervention $\Do(X=x)$ replaces the mechanism for $X$ in the given (open) causal model $\modelM$ with a fixed sharp state $x$ on $X$, representing a `reaching in' and forcing the variable to take a certain value independent of what the direct causes of $X$ in $\modelM$ are.\footnote{A `sharp state' captures the idea of a fixed value $x$ of $X$. As a general technical term it is defined in Sec.~\ref{sec:setup}, though in $\FStoch$ a sharp state simply is a point distribution.}  
In compositional model terms, it thus is a meta-operation that alters the signature of the model $\modelM$ by replacing the mechanism $c_X$ for $X$ with a mechanism of the right-hand form below, such that $\sem{\syn{x}} = x$ in $\catC$.  
\begin{equation} \label{eq:do-int-map}
\input{do-meta.tikz}
\end{equation}
Suppose $\modelM$ has input variables $I$ and output variables $O$. Common notation is to write $P(O|I)$ for its induced channel $\sem{\modelM}$. Then $\Do(X=x)$ induces a causal model $\modelM'$, whose induced channel is now commonly denoted $P(O| I ; \Do(X=x))$.

\begin{example}
Consider the causal model from Ex.~\ref{ex:smoking-causal-model}. 
For an intervention $\Do(S=s)$ -- say, forcing someone to (not) smoke -- the transition between models yields the following channel $P(S,L,A;\Do(S=s))$ in $\catC$: 
\[
\input{open-example2.tikz}
\]
\end{example}

Various more general kinds of interventions have been considered in the literature such as soft and conditional interventions. 
They all fall under the following most general notion, which was first proposed in \cite{CorreaEtAl_2020_CalculusForStochasticInterventions} and then studied in string diagrammatic language in \cite{lorenz2023causal}. 
An \emph{intervention} $\sigma$ on variable $X$ of (open) causal model $\modelM$ replaces its mechanism $c_X$ with any other channel $c_X'$, subject to the condition of preserving acyclicity of the overall causal structure of the resultant new model $\sigma(\modelM)$. 
\[
\input{int-general.tikz}
\]
In particular the parents of $X$ may change. 
See \cite{lorenz2023causal} for a discussion of many interesting special cases other than do-interventions. 

\begin{example}
Returning once more to the smoking causal model from Ex.~\ref{ex:smoking-causal-model}, consider an intervention $\sigma$ which represents a public health policy such that all people under 21 years of age have a 90\% probability of not smoking, while all those above 21 years are unaffected by the policy \cite{CorreaEtAl_2020_CalculusForStochasticInterventions}. This replaces the mechanism for $S$ with a new mechanism $c'_S$ of the form:  
\[
\input{int-example.tikz} \qquad = \begin{cases}	0.9 \ \delta_{S,0} + 0.1 \ \delta_{S,1}  &  \forall A < 21 \\  c_S(B) &  \forall A \geq 21 \end{cases} 
		\label{Eq_WideLocalIntervention_NewChannel}
\]
where the channel $\eta$ may be deduced from the above expression for $c'_S$. 
This intervention yields a new model with the following output state: 
\[
\input{general-int-ex-2.tikz}
\]
\end{example}

We emphasise that causal models, and the causal reasoning performed on them such as interventions, are most naturally presented in string diagrams. This applies also to further aspects of the causal model framework, discussed in Section~\ref{sec:frameworks}.  

\paragraph{Interpretation.}  


A causal model does \emph{not} a priori come with an interpretation in our sense, be it abstract or concrete. However, in the contexts where CBNs tend to be used and where much of the causal model literature sees their intended purpose, usually all (at least observed) variables have an abstract and a concrete interpretation. 
This is because the models are applied to `data science problems' where there is observational or interventional (experimental) data involving variables with concrete interpretations. These variables are the aspects we consider causally relevant to the phenomenon being studied and given from the outset.  
For instance, in Ex.~\ref{Ex_CBN} all variables have an abstract and, with the exception of $B$, also a concrete interpretation. 
Note that a non-output variable $X$, i.e. a hidden or latent variable, might be part of the model simply because one cannot exclude the possibility of some confounding common cause unknown to us and it thus lacks any interpretation. 
However, also latent variables may have an abstract and even concrete interpretation, since being unobserved may just signify the lack of empirical data, despite being a specific variable that we `understand'. 

What does necessarily come with stating a causal model is the status of the variables as causal variables (causal relata) and the morphisms as causal mechanisms. In our setup, that status -- essential for the causal model framework with its many concepts and rules -- is best treated however using our general notion of a compositional framework, introduced in Section~\ref{sec:frameworks}.  

\subsubsection{Functional causal models} \label{subsubsec:FCMs}

In the causal model framework, there is also an important `more refined' form of causal model known as a \emph{Functional Causal Model} (FCM), or perhaps more commonly a \emph{Structural Causal Model} (SCM)  \cite{pearl2009causality}. 
Formally speaking, an SCM is a special case of a CBN, where each mechanism factorises in terms of an underlying `true' deterministic functional component, along with an additional `noise' variable with no parents, encoding our uncertainty about the underlying state of the world. 
By treating causal relations fundamentally as functional dependencies, formalised via SCMs, one can then go on to derive the rest of the causal model framework for CBNs as theorems, rather than via ad hoc reasoning.  In particular every CBN can be seen to arise (non-uniquely) from some SCM through marginalisation. From an interpretability viewpoint, SCMs are especially important in that they are the kind of causal model most pertient to studying neural network-based models, and allow us to define counterfactuals, which we treat in Section~\ref{sec:frameworks}.




Recall the definition of a deterministic morphism~\eqref{eq:deterministic}, which generalises the notion of a function. This allows us to define an FCM in any cd-category. 



\begin{definition} \label{def:FCM}
A \emph{Functional Causal Model (FCM)} in a cd-category $\catC$ is a causal model $\model{M}$ in $\catC$ whose variables are partitioned into \emph{exogenous} variables $\syn{U}=\{\syn{U_i}\}^n_{i=1}$, where each $\syn{U_i}$ is hidden and has no parents, having a mechanism of the form: 
\begin{equation} \label{eq:noise-var}
\begin{tikzpicture}[tikzfig]
    \begin{pgfonlayer}{nodelayer}
        \node [style=none] (0) at (0, 0.25) {};
        \node [style=map] (1) at (0, -0.75) {$\syn{\lambda_i}$};
        \node [style=label] (2) at (0, 0.75) {$\syn{U_i}$};
    \end{pgfonlayer}
    \begin{pgfonlayer}{edgelayer}
        \draw (0.center) to (1);
    \end{pgfonlayer}
\end{tikzpicture}
\end{equation}
and \emph{endogenous} variables $\syn{V} = \{\syn{X_i}\}^n_{i=1}$, where each $\syn{X_i}$ has a mechanism of the form:  
\begin{equation} \label{eq:SCM-2}
\input{SCM-2.tikz}
\end{equation}
where $\Pa'(\syn{X_i}) \subseteq V$ and $f_i = \sem{\syn{f_i}}$ is deterministic. 
\end{definition}

Formally, the signature for an FCM includes equations of the form \eqref{eq:deterministic} specifying that each generator $\syn{f_i}$ is deterministic. The usual kind of FCM considered in the literature is one with $\catC = \FStoch$, and we reserve the term SCM for these. An SCM is thus given by finite sets $X_1,\dots,X_n$, $U_1,\dots,U_n$, describing values of the endogenous and exogenous variables, respectively, along with a distribution $\lambda_i$ over each $U_i$ and for each $X_i$ a function $f_i \colon \Pa'(X_i) \times U_i \to X_i$ for a subset $\Pa'(X_i) \subseteq \{X_1,\dots, X_n\}$. 



\begin{example}
An SCM with variables $\syn{V=\{B,S,L,A\}}$ and $\syn{U=\{U_B, U_S,U_L,U_A\}}$ and the same causal structure amongst $\syn{V}$ as in Ex.~\ref{ex:smoking-causal-model} is given by a representation in $\FStoch$ of the network diagram below, such that each $f_X$ is deterministic, for $X \in \syn{V}$. The pairs $f_X, \lambda_X$ for $\syn{X \in V}$ may be chosen such that their composite recovers precisely the corresponding channel $c_X$ from our example CBN in Ex.~\ref{ex:smoking-causal-model}.
\begin{equation}
	\input{Fig_Example_FCM_DAG_ND.tikz} 
	\nonumber
\end{equation}
\end{example} 

%

FCMs are closely related to a final form of causal model which is most relevant to the study of (deterministic) ML models. We call an (open) causal model \emph{deterministic} when it is an (open) causal model but where every mechanism is a deterministic morphism. Thus it is given by a network diagram where every box is deterministic. We can thus arrive at a deterministic open causal model by starting from an FCM and `removing' the input state $\lambda_i$ for each `noise' variable $U_i$, turning them into input variables of the new model. 
Many of the models we met in Section \ref{sec:examples-compositional-models}, such as linear models, neural networks or decision trees, are specified by network diagrams represented as functions, and so indeed may also be viewed as deterministic open causal models. 
When doing so, we view each of the morphisms $+$ and $\mult$ as mechanisms themselves, with a single output. 
This causal view on deterministic models is in fact common in XAI and will be discussed further in Sec.~\ref{sec:CI-and-causal-XAI}.




\paragraph{Interpretation.} 

Just as for generic causal models, an FCM a priori has no interpretation, abstract or concrete. In the contexts where SCMs have been studied in the original causal model literature, they tend to play the role of a fine-graining or refinement of a CBN, and are supposed to capture the true (albeit usually unknown) underlying functional model in data science problems. 
Here the endogenous variables usually have an abstract and concrete interpretation. The exogenous `noise variables' however tend not to. In ML contexts where SCMs, or more precisely deterministic open causal models (see above), appear frequently, they do not always have either an abstract or concrete interpretation. Variables in such a model might include computational units, e.g.~a neuron's weights, which are not readily interpretable. See Sec.~\ref{sec:CI-and-causal-XAI} for a detailed discussion.

\section{Compositional frameworks}
\label{sec:frameworks}

The discussion in Section \ref{sec:examples-compositional-models} demonstrated how a wide range of AI models can be formally seen as compositional models, and that our definition of an interpretation (Definition \ref{def:interpretation}) lets us analyse a model's interpretability in terms of its components. However, there is a further aspect of the intuition of a model's interpretability that is not captured by this definition, but which features in XAI discussions, and which the compositional perspective can again help us to disentangle. The further aspect is what we call the \emph{compositional framework} of the model. Independently from having an (abstract or concrete) interpretation, a model may live in a particular framework that licences certain types of reasoning with its components. Roughly speaking, the framework answers the question of how useful the explicit structure of a model is once seen as a compositional model. That is, what kind of reasoning and what meaningful computations can be done using this structure? 

Here we will focus on examples of frameworks, revisiting several classes of models which share the same well-defined framework. We will not define the general notion of a framework formally here, with the question of whether there is an insightful such formalisation being left for future work.

\subsection{Framework of input-output models} \label{subsec:io-models}

Our first example can be seen as a `trivial' notion of a framework. The framework of input-output models applies to any model which comes with an associated overall input-output map $M$ in the semantics category. This includes any model generated by a single diagram $\diagD_\modelM$, where $M = \sem{\diagD_\modelM}$ is the resulting morphism in $\catC$ from inputs to outputs. Examples include linear, rule-based, and transformer (on fixed input length) models.
\[
	\input{io-model.tikz}
\]
What can we do with such a mapping? Firstly, one can consider different input states and obtain corresponding output states. Indeed, this is the way in which many models are intended to be used. In $\Setcat$ or $\NN$ any state necessarily is a product state, i.e.~just a list of values for the corresponding variables, as in the left-hand diagram below. More generally, the input state $s$ could be a non-product state, as on the right-hand side, and for $\FStoch$ it may in particular be the probability distribution underlying the given data in a training scenario. 
\[
		\input{io-model-input-product-state.tikz}
\qquad \qquad \qquad \qquad \qquad 
		\input{io-model-input-non-product-state.tikz}
\]

Many models have just one single output variable, but if the output is given by $k$ distinct variables as above, one can also compute the marginal on a subset of outputs, and apply this to input states as above. Below we depict the \rl{induced marginal over $R \subseteq \{Y_1,...,Y_k\}$.} 
\[
		\input{io-model-marginal-map.tikz}
\]




While it may seem rather trivial to regard the input-output behaviour as a framework, this is all that can be done with some models.
Consider for instance a (sparse) linear regression model for a simple data science problem. 
Such a model is `perfectly interpretable' with each component interpreted and appearing in the diagram. 
However, the structure of the model has \stb{no obvious further use}. There is no framework that would tell us how to use the components on their own for further reasoning or computing quantities of interest. 

\subsection{Framework of statistical models} \label{subsec:stats-models}

Our next framework considers models which allow for statistical reasoning about their variables, including conditioning. By an (open) \textit{statistical model} we will mean any compositional model $\modelM$ that has a distinguished morphism $M$ 
and semantics in $\catC_{\text{channel}}$, where $\catC$ is a cd-category that has suitable further structure to model probabilistic reasoning. For the details on this structure see~\cite{lorenz2023causal}, which builds on categorical probability theory \rl{\cite{CoeckeEtAl_2012_PicturingBayesianInference, cho2019disintegration, fritz2020synthetic, di2023evidential}}. For simplicity here we restrict to $\catC=\MatR$ with $\catC_{\text{channel}} = \FStoch$, i.e.~probability spaces over finite sets. 
The most common statistical models are without inputs, defining a probability distribution $M$ over the output variables, and we will focus on these. However, one can treat open models, where $M$ is a probability channel with inputs, in the same way.

What does the framework of statistical models allow? Firstly, statistical models live in particular in the framework of input-output models: feeding in an input distribution yields an output distribution and one can compute marginal probability distributions and channels.
In addition, however, one can \emph{compute conditionals}. 
In conventional notation, given a distribution $M=P(Y,X)$ over sets $X$ and $Y$, 
$P(Y|X=x)$ denotes the distribution over $Y$ conditional on $X=x$, and $P(Y|X)$ denotes the induced conditional stochastic map. In string diagrams, the former can be represented by the expression on the left below and the latter by the one on the right.
\[
	\input{conditional.tikz}
	\hspace*{3cm}
	\input{conditional-channel.tikz}
\]

To understand precisely the formal meaning of these dashed boxes we refer to~\cite{lorenz2023causal}, but roughly the meaning of the blue dashed box is to turn a morphism in $\MatR$ into a correctly normalised probability channel.\footnote{Technically it yields a partial probability channel, which on any input is either normalised or `undefined'.}  In particular it turns a process with no input into a probability distribution by mapping $P(Y,X=x)$ to $\frac{P(Y,X=x)}{P(X=x)}$. The cap (bent down wire) on the right corresponds to a Kronecker-delta.\footnote{Formally, $\MatR$ is a compact category with the Kronecker-delta as a cap in the sense of Section \ref{subsec:DisCoCat}.} These two diagrams for conditioning on a point $X=x$ are related by the following equalities:
\begin{equation}
	\input{sharp-updating.tikz}
	\label{eq:sharp-updating}
\end{equation}
The framework of statistical models can also include updating our knowledge with respect to fuzzy evidence given by a distribution $\omega$, rather than just sharp facts like $X=x$. Interestingly, there are in fact multiple meaningful ways in which we may update our knowledge with such fuzzy evidence. In contrast to the equality in \eqref{eq:sharp-updating}, for a generic state $\omega$ (i.e. not a point distribution) and its corresponding effect, the left and right-hand sides of the inequality below show two distinct yet natural ways to apply updating, called \emph{Jeffrey updating} and \emph{Pearl updating}, respectively, as discussed by Jacobs \cite{jacobs2019mathematics, jacobs2021learning}. 
\begin{equation}
	\input{different-updating.tikz}
	\label{eq:different-updating}
\end{equation}

\begin{figure}[h]
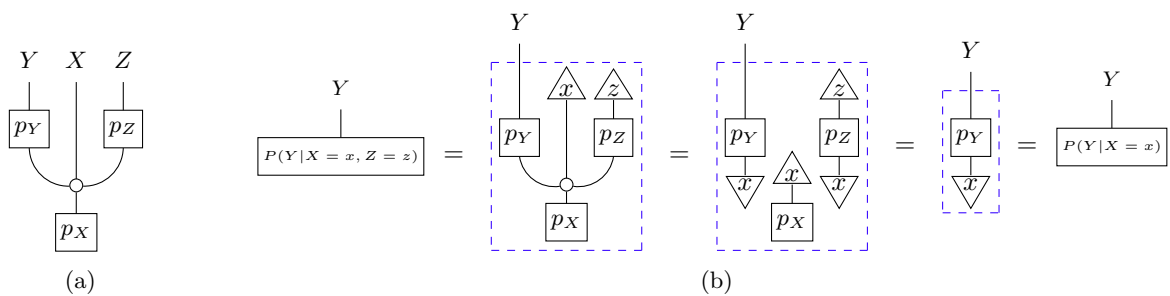

	\begin{subfigure}{0.18\textwidth}
		\centering
		\input{BN-example-fork.tikz}
		\caption{ }
	\end{subfigure}
	\hfill
	\begin{subfigure}{0.8\textwidth}
		\centering
		\input{BN-example-fork-conditioned.tikz}
		\caption{ }
	\end{subfigure}
	\caption{(a) Basic example of a BN: the \emph{fork} $Y \leftarrow X \rightarrow Z$; (b) Diagrammatic rewriting, using a variant of~\protect\eqref{eq:copy-points} and properties of the normalisation box \cite{lorenz2023causal}, to show how the structure of the network diagram can be used to infer the conditional independence implied by the fork. \label{fig:example-cond-ind}}
\end{figure}

The diagram associated with $\modelM$ may of course have internal structure. A particularly informative case is when it is a network diagram (see Def.~\ref{def:network-diagram}), in which case the model is a Bayesian network. 
The structure then encodes conditional independence relations as formalised through the notion of \emph{d-separation} \cite{GeigerEtAl_1990_IdentifyingIndependence, VermaEtAl_1990_CausalNetworks}. 
Both conditional independence relations and d-separation have been generalised within categorical probability theory and can be treated diagrammatically \cite{cho2019disintegration, lorenz2023causal, fritz2023d}, illustrated with a simple example in Figure~\ref{fig:example-cond-ind}.  

In summary, the framework of (open) statistical models allows the computation of any combination of conditioning, marginalising and fixing distributions for some inputs. 

\subsection{Framework of causal models} \label{subsec:framework-of-causal-models}

Section~\ref{subsec:causal-models} covered the notion of a \textit{causal model}, but barely touched on the many concepts and results of its rich framework, which the literature indeed often refers to as the \emph{framework} of causal models.  Being well established, for a full account of this framework we refer to \cite{lorenz2023causal, jacobs2019causal} for the compositional model view and to \cite{pearl2009causality, SpirtesEtAL_2000_BookCausationPredictionSearch, bareinboim2022pearl} for a conventional presentation. Here we only mention a few of its key aspects, with a view to the discussion of causal concepts in XAI, in Section~\ref{sec:CI-and-causal-XAI}.



\paragraph{Interventions.} A core aspect of the causal model framework are interventions, which we met in Section~\ref{subsec:causal-models}.  An intervention, be it a basic do-intervention or a more general kind, is a well-defined operation relative to any causal model $\modelM$ which transforms it to a new one $\modelM'$ with altered causal structure and semantics (by altering causal mechanisms). The overall process $\sem{\modelM'}$ of the new model is at times also referred to as the `post-intervention distribution' and as we saw denoted $P(Y; \Do(X))$ for a do-intervention.


The ability to compute interventions, and thus quantities that refer to a causal scenario distinct from the original one, is a feature of causal models over statistical ones. Formally, one is of course free to intervene on a general Bayesian network just as for a CBN. However, the resulting network
 would generally have nothing to do with the phenomenon or data generating process that was described by the original network, unless the structure is asserted as having a causal status.




\paragraph{Counterfactuals.} 
For a \emph{functional} causal model the framework is richer still, allowing the computation of well-defined `counterfactuals'. The literature is not always clear about what is and what is not a 
\rl{counterfactual,} 
and so for clarity we state a general definition below, after the following simple example. 




\begin{example} \label{ex:aspirin-counterfactual}
Consider the counterfactual question: \emph{`Had Mary taken an aspirin last night, would she still have woken up with a headache today?'} \cite{ShpitserEtAl_2008_CompleteIdentificationMethodCausalHierarchy}. 
To answer this, suppose an SCM of the below form is given with variables $A$ and $H$, denoting whether Mary takes an aspirin and gets a headache, respectively, 
each represented by $\{y,n\}$ for `yes', `no':  
\[
\input{cf-aspirin-1.tikz}
\]
Here $U_A, U_H$ are the respective exogenous variables with distributions $\lambda$, $\lambda'$. 
To evaluate the counterfactual, we intuitively wish to restrict to those (albeit unknown) background conditions that led to the event which occurred, namely that $A=n$, $H=y$, but now intervene so that an aspirin is being taken, and determine the resulting distribution over headaches. The counterfactual distribution is thus represented by the string diagram on the LHS below, which is then simplified via rewrite rules \ir{(see Sec. 8.1 of \cite{lorenz2023causal} for the details)}. 
\[
\input{cf-aspirin-2.tikz}
\]
We can think of the LHS diagram in terms of `parallel worlds', which share the same background variables (through perfectly correlated copies of $U_A$ and $U_H$), where the left-hand world shows (via the conditioning) what actually occurred, and the right-hand one a hypothetical world where aspirin was taken, given by the intervention.  
\end{example}

The key ideas are that interventions allow the construction of hypothetical worlds, that functional dependencies allow the background conditions to be kept the same and `stitch the worlds together', and that conditioning incorporates the facts that obtained in the actual world. These lead to the following definition of counterfactuals from \cite{lorenz2023causal}, which matches (but slightly generalises) the notion from the conventional causal model literature 
\cite{pearl2009causality, Pearl_2011_AlgorithmizationOfCounterfactuals, BareinboimEtAl_2021_PearlsHierarchy, HalpernEtAl_2005CausesAndExplanationsI}. 

Given an FCM $\modelM$, consider its factorisation into a model $\model{L}$, consisting only of its (`latent') exogenous variables with the product state over them \eqref{eq:noise-var}, and the model $\model{F}$ consisting of its endogenous variables with their deterministic mechanisms \eqref{eq:SCM-2}. Each computable counterfactual is given by a diagram in $\catC$ of the following form: 
\[
\input{counterfactal-simpler-2.tikz} \label{eq:counterfactual-simpler}
\]
Here the exogenous variables $U$ are copied to $n$ many `parallel worlds', each given by applying a do-intervention $\sigma^{j}$ to the deterministic part $\modelF$ of the original FCM, as well as conditioning on some variables via $c^{j}$ and marginalising some $D^{j}$.\footnote{Further we require that for some $j \neq j'$ we have $C^{j} \neq \emptyset \neq E^{j'}$.} 

Counterfactuals are thus another quantity of interest computable for a particular class of models, determined by the status of their compositional structure, but in principle independent from an interpretation in our earlier sense. We return to discuss their relation to counterfactual explanations in XAI in Section~\ref{subsec:CFEs-vs-CFs}.


\paragraph{Identifiability problems and the causal hierarchy.} 
In data science problems, one is rarely confident as to what exactly the causal structure is. There may be unknown common causes, and even if one is aware of all of them they may lack empirical data. As a result much of the causal model literature is concerned with \emph{causal identifiability problems}: without a fully specified SCM or CBN, under which conditions on the causal structure and the available empirical data can a certain quantity still be inferred unambiguously? Such quantities include post-intervention distributions and counterfactuals. 

For many identifiability problems, complete algorithmic solutions have been developed which, given (partial) causal knowledge, some empirical data and a query, output whether the quantity is identifiable and if so what it is \cite{pearl2009causality, ShpitserEtAl_2008_CompleteIdentificationMethodCausalHierarchy, CorreaEtAl_2020_CalculusForStochasticInterventions}. 
The results establish that there is a strict separation in the identifiability of various quantities, forming what is known as the \emph{causal hierarchy} \cite{pearl2009causality, BareinboimEtAl_2021_PearlsHierarchy, Pearl_2018_TheBookOfWhy}, with the identification of purely statistical quantities, causal effects and counterfactuals forming three levels, each of which requires strictly more detailed causal knowledge than the preceding one. 	
Note that many identifiability problems can be treated string diagrammatically \cite{lorenz2023causal, jacobs2019causal}, in which case the answer to a query is produced through a series of diagrammatic rewrites, licensed by both the status and properties of the compositional structure (e.g.~d-separation conditions for the well-known \emph{do-calculus} rules). 
The causal hierarchy has implications for AI that have been discussed in many works, e.g.~\cite{Pearl_2018_TheoreticalImpedimentsToML, SchoelkopfEtAl_2021_TowardCausalRepresentationLearning}. In our terminology we can say that the causal hierarchy is independent from the presence of an interpretation, but manifests in the difference between the frameworks of statistical models, \rl{CBNs} and FCMs.

\subsection{Frameworks for NLP models} \label{sec:using-NLP-models}

For our last example, we consider how we might define the frameworks for NLP type models, which do not come with a single distinguished input-output process. These include transformers (now with varying input length), and sequence models, as well as CCG, DisCoCat and DisCoCirc models.  What are the meaningful processes we can compute with models of this kind?

Most importantly, for any such model, when given a phrase $\syn{w_1},\dots,\syn{w_n}$ we can compute its representation as a state in the semantics category $\catC$:
\[
\begin{tikzpicture}
	\begin{pgfonlayer}{nodelayer}
		\node [style=map] (0) at (0, 0) {$\syn{w}_1,\dots,\syn{w}_n$};
		\node [style=none] (1) at (0, 1.75) {};
	\end{pgfonlayer}
	\begin{pgfonlayer}{edgelayer}
		\draw (1.center) to (0);
	\end{pgfonlayer}
\end{tikzpicture}
 
\]
Explicitly, for a sequence model such as an RNN this is the state representation \eqref{eq:state-rep}. For a DisCoCat model (or CCG model) the phrase is first sent to a parser which assigns it a grammatical structure equivalent to a diagram, from which we obtain a state such as \eqref{eq:aliceplaysfootball}. For a DisCoCirc model, the parser returns a text circuit for the phrase, to which we apply some fixed initial states $\star$ to each noun wire. A transformer model applies the network \eqref{eq:transformer} for input length $s=n$ to the token labels.

However, computing word \rl{or phrase} representations alone will not be of much use in practice. For actual applications, the framework for an NLP model must also allow us to compute useful outputs from this representation. To do so, each of these models is applied with processes which map this text representation to some output. One way to think of such output processes is as \emph{questions} $(Q_i)^n_{i=1}$ we can ask of the representation, each yielding an answer from some set $A_i$ of answers:
\[
\input{text-to-output2.tikz} 
\]
For example, the question could ask for the likely next word, as in a generative language model. Alternatively each question could return a class such as a sentiment label for the phrase. Or we could have a DisCoCat model followed by a process which assigns a sentiment from $A=\{\text{positive},\text{negative}\}$ to a movie review as below. 
\[
\input{Dcat-review.tikz}
\]
Thus, in general, the framework for any NLP model will include at least the representation states for phrases $(\syn{w}_1,\dots,\syn{w}_n)$ and a set of `question' processes $Q_i$, allowing us to compute outputs as above.

For models where the representation space of a phrase has interesting compositional structure, however, these questions may themselves be structured, leading to a richer framework. For example, a DisCoCirc model is typically trained along with processes for a set of questions, each of which acts on a set of nouns to which it is relevant. A question \natlang{Does Alice know Bob?} would be represented by a process of the following form, with discards on the irrelevant nouns:
\[
\input{dcirc-question-ex.tikz}
\]
Given a text circuit, we can obtain a distribution over possible answers $A$ by applying the text circuit to the initial states, as below.  
\begin{equation} \label{eq:dcirc-fwork-output}
\input{dcirc-text-question-ex.tikz}
\end{equation}

A quantum DisCoCirc model trained with questions of this form is explored in \cite{QDisCoCirc}. Combining \stb{the structure in \eqref{eq:dcirc-fwork-output} with} the fact that the text circuit itself is compositionally structured, it can be seen that DisCoCirc models come with a richer framework that allows for reasoning about the answer to such a query. We will see this explicitly when we consider the explanations offered by diagrams of this kind in Section \ref{sec:expl-from-diags}.

\section{Aspects of conventional XAI}
\label{sec:CI-and-causal-XAI}

In this section we discuss various aspects of conventional approaches to XAI from our compositional perspective. 

\subsection{Influence, causal structure and feature importance} \label{subsec:causalXAI-signalling-etc}

A basic consideration which may help us better understand a model is that of which inputs can affect which outputs. Most models we discussed contain channels (Def.~\ref{def:disc-cat}), where the input and possibly also the output often have a product structure of several variables. For such processes, we can reason about the \emph{relations of influence} between inputs and outputs, also known as \emph{signalling relations}. 

\paragraph{Influence and interventions.}
Given a channel $c \ \colon \ X_1, X_2 \rightarrow Y_1, Y_2$, we say that variable $X_2$ \emph{does not influence (signal to)} $Y_1$ iff the marginal channel into $Y_1$ factorises as follows for some channel $d$: 
\begin{equation} \label{eq:nosignal}
	\input{nosignal_v2.tikz}
\end{equation}
The terminology of `no-signalling' comes from how interventions on an input variable propagate through $c$ -- altering $X_2$ won't affect $Y_1$, and hence no signal can pass from the former to the latter. More precisely, no matter what state $s_2$ is fed into $X_2$, the marginal state at $Y_1$ does not depend on $s_2$:
\[
	\input{nosignal_v2-input-state.tikz}
\]
Note that, seeing as $X_1, X_2$, $Y_1$ and $Y_2$ may have further product sub-structure, we can define (no-)influence from any collection of inputs to a collection of outputs in the same way. 


\paragraph{Causal structure in NNs.}

One way to situate several approaches in conventional XAI is that they are based on the idea that a given neural network-based model implicitly is, or induces, a causal model. This can be understood in several ways, which it will be helpful to clarify.

Firstly, observe that any such model defines a function, i.e.~deterministic morphism $f \colon X_1 \times \cdots  \times X_n \rightarrow Y_1, \times \cdots \times Y_k$, which we can view as an open deterministic causal model (Section \ref{subsubsec:FCMs}) as follows. 
Just like any function, $f$ factorises into $k$ component morphisms $f_{j}$ from its inputs to each output $Y_j$. Moreover, each $f_j$ will in general only depend on a subset $S_j$ of the inputs. (This subset $S_j$ is determined by excluding those inputs for which $f$ satisfies no-influence to $Y_j$.) Hence we can rewrite the function $f$ as below, where each $X_i$ is only copied to $f_{Y_j}$ if $X_i \in S_j$. The resulting diagram is a network diagram, and thus defines an open deterministic causal model whose variables are simply the inputs and outputs, in which $Y_j$ is a child of $X_i$ in the causal structure iff $X_i \in S_j$. 
\[
\input{function-as-det-OCM-3.tikz}
\]

There is also a second, more immediate way in which to view a neural network-based model as an open deterministic causal model. By construction, any such model will be specified as a computational graph, either down to the level of neurons, as in neural network diagrams (Section \ref{subsec:NN-model}), or at a higher level of abstraction, as in the diagram for a transformer (Section \ref{subsec:transformer-models}). As we have seen, any such graph \rlc{can be seen as} a string diagram, and in fact a network diagram. Thus the diagram explicitly represents the function as an open deterministic causal model, now including intermediate variables along with the inputs and outputs.

\paragraph{Post-hoc methods approximating causal properties.}
\rlc{Seeing an NN-based model as inducing an (open deterministic) causal model in this way is increasingly common in the context of XAI methods. 
With the structure and specific properties of this implicit causal model typically unknown, the goal of many of the post-hoc XAI methods can be seen as trying to reveal facts about it.}

\rlc{One causal aspect to explore is the set of influence relations. In practice, however, for a trained deep NN there will typically not be any \emph{no-}influence relations that hold exactly -- that is, no strict equality as in Eq.~\ref{eq:nosignal} -- but at most approximate ones, due to the full connectivity of the string diagram. As a result, interest focuses on the `strength' of influences between variables. The many variations of feature importance (feature attribution) methods -- be they gradient-based \cite{chattopadhyay2019neural, zeiler2014visualizing, shrikumar2017learning, binder2016layer} or perturbation-based \cite{feder2021causalm, pryzant2020causal, vig2020causal, de2021sparse, abraham2022cebab, chan2022causal} -- essentially study the causal effect of features on the output. 
(Also see the discussion in~\cite{GeigerEtAl_2022_InducingCausalStructureForInterpretableNN}.) 
Here features may be inputs or neurons of intermediate layers, and one means of exploring the causal effects is through interventions on these corresponding input or intermediate variables. Note that freely choosing an input state $x$, as is common in `extrapolation' or `data augmentation' techniques, is a special case of this, and in general has to be seen as an act of intervention.\footnote{Given data distribution $D$ over $X$, feeding $x$ into some model with input $X$ can only be seen as conditioning on $x$ (or sampling from $D$), if $x$ is in the support of $D$. 
In contrast, whatever causal model $\modelM_D$ might be behind the generation of $D$, intervening on all variables via $\Do(X_1=x_1,...,X_n=x_n)$ \emph{can} produce any $x=(x_1,...,x_n)$. In this sense, then, although freely choosing the input state $x$ conceptually speaking has to be regarded as of interventional status, it does not rely on any specific causal knowledge and is part of the framework of input-output models in Sec.~\ref{subsec:io-models}.} 
A form of explanation with an explicitly causal flavour, and based on such intervening on inputs, are counterfactual explanations, which we discuss in more detail in Section~\ref{subsec:CFEs-vs-CFs}.}

While one can always study the properties of a deterministic model through interventions on input or intermediate variables, 
arguably the resulting dependence statements only help with the interpretability of the model when the concerned variables have an (at least abstract) interpretation. 
This is usually the case for the input and output variables, 
while intermediate variables such as individual neurons in middle layers tend not to have an interpretation. 
Yet a seemingly common attitude in XAI is that a successfully trained model must have implicitly learned `meaningful' intermediate variables, and by probing or intervening on them, one may discover their interpretation. 
See Sec.~\ref{subsec:Issues_with_XAI} and \cite{FreieslebenEtAl_2023_DearXAICommunity} for a critical assessment of this view. 

Whichever methods are applied, note that, while viewing a model as a causal one in this way has a clarificatory benefit, it does not provide anything formally new 
from the perspective of the causal model framework. 
We are `just' studying a model with all variables `observed', rather than using a causal model to make predictions about hypothetical interventions 
that cannot be performed experimentally.

\paragraph{The explanatory role of causal notions.} 
To the degree that the XAI methods outlined above are explanatory, they explain \emph{the model}. 
The implicit causal structure that is being studied is that of the model, and a priori has nothing to do with the one in the world -- see Sec.~\ref{subsec:model-vs-world} for a more detailed discussion.  
Nonetheless, some argue that the most important way in which humans understand any phenomenon is in causal terms, and that their preferred explanations are causal explanations, and so ideal explanations of an AI model (in particular of input-output pairs) are causal explanations also \cite{FreieslebenEtAl_2023_DearXAICommunity, HarradonEtAl_2018CausalLearning}. 
This attitude naturally leads to seeing a model that 
has fully explicit causal structure and a complete interpretation simply as `its own explanation' \cite{CarloniEtAl_2023_RoleOfCausalityInXAI}. 
Such a model has a kind of `intrinsic interpretability' that, due to the explanatory significance of autonomous causal mechanisms, goes beyond the transparency and traceability aspects of the classic examples of intrinsically interpretable models.   
Such an attitude is well aligned with this work's argument for the benefits of explicit and interpreted compositional structure, and we will return to it in Section~\ref{sec:comp-and-interp}.

\subsection{Counterfactual explanations} \label{subsec:CFEs-vs-CFs}

As mentioned in Section~\ref{sec:context-XAI}, among the most common post-hoc methods in XAI are ones that produce local explanations called \emph{counterfactual explanations} (CFEs) \cite{WachterEtAl_2017_CounterfactualExplanations}. 
Roughly speaking, given a deterministic model with distinguished process $M$, \rlc{which has} input variables $X=X_1 \ldots X_n$ and output variable $Y$,\footnote{$Y$ could also have product structure (but that has no bearing on the discussion of CFEs in this section).} 
a CFE of input-output pair $(x,y)$ is an alternative pair $(x',y')$ with $y \neq y'$ (and $y'$ usually of particular interest to the explainee), and such that $x'$ is as similar to $x$ as possible given that $ M \circ x' = y'$:
\begin{eqnarray}
	\input{basic-CFE_x.tikz} & \hspace{1.5cm} \quad  \hspace{1.5cm} & \input{basic-CFE_x_prime.tikz}  \label{eq:basic-CFE_x} 
\end{eqnarray}
Here similarity is taken with respect to some chosen notion of distance $d(x,x')$. 
For example, $x$ and $x'$ might only differ in as few dimensions as possible, say only in the $i$th variable:
\begin{eqnarray}
	\input{basic-CFE_s_x.tikz} & \hspace{1.5cm} ; \hspace{1.5cm} & \input{basic-CFE_s_x_prime.tikz}   \label{eq:basic-CFE_s_x_prime}
\end{eqnarray}
The non-trivial aspects of CFE methods consist in picking a suitable notion of similarity and in providing a solution to the practical problem of finding CFE pairs. For the sake of concreteness, consider the following much discussed kind of loan example.

\begin{example} \label{ex:CFE-bank-example}
(Adjusted from \cite{Freiesleben_2021_IntriguingRelation}) Suppose a person applies for a loan with a bank that decides deterministically whether to grant it based on a model $M_L$ that takes age, salary, capital, number of open loans, and number of pets as input features. 
Also suppose the application is rejected and moreover that, on the basis of a CFE method, they are told by the bank that ``had they had a £5k higher salary and two pets, they would have got the loan".    
\end{example} 
It is understandable that this kind of `explanation' is popular -- it appeals to the link between counterfactual dependencies and causal relations, and the explanatory significance these have, even 
in everyday life. There are, however, crucial issues with CFEs, such as their lack of uniqueness. See, for example,~\cite{Freiesleben_2021_IntriguingRelation} for insightful discussions. Here we focus on two points: in what sense CFEs are counterfactuals, and then in Section~\ref{subsec:model-vs-world} how the loan example relates to the `model-vs-world' distinction.

\paragraph{CFEs and Pearlian counterfactuals.}
The presentation of CFEs in the literature often claims that a CFE is a counterfactual in the causal model sense (an exception is \cite{CarloniEtAl_2023_RoleOfCausalityInXAI}).  
However, simply varying the inputs of a function does not, in an obvious way, look like a special case of the definition of a counterfactual in Eq.~\eqref{eq:counterfactual-simpler}. 
Of course, the idea of a function, some of whose inputs are changed through intervention, does play an important role in the definition of a counterfactual. 
However, this alone is not enough to justify the use of the term in the sense of the causal model framework. 
For the details see App.~\ref{App:CFE-details}, but the crux of the argument is this: without further causal assumptions beyond just $M$, it does not make sense to regard the alternative state $x'$, which differs from $x$ only in $X_i$ -- or more generally, some subset $S$ of the input variables -- as arising through \emph{only} intervening on $X_i$. And yet this is supposedly the ``had $X_i$ been different" part (and all else the same) of the counterfactual's antecedent. As a result, the pair $(x',y')$ does not constitute a well-defined counterfactual relative to just the given model $M$.



\subsection{An important distinction -- model vs world} \label{subsec:model-vs-world}

Any form of interpretability should be informed by the specific question and purpose that the model is supposed to address \cite{FreieslebenEtAl_2023_DearXAICommunity}. 
A key distinction is whether we need  to reach a degree of transparency and explainability of the \emph{model}, or whether we are concerned with the \emph{world, or phenomenon that the model is about}.
The importance of distinguishing concerns of model vs world in XAI has been stressed in various works 
\cite{FreieslebenEtAl_2023_DearXAICommunity, Beckers_2022_CausalExplanationsAndXAI, FreieslebenEtAl_2022ScientificInferenceAndIML, Sullivan_2022_UnderstandingFromMLModels, Watson_2022_ConceptualChallengesForIML, CarloniEtAl_2023_RoleOfCausalityInXAI, Freiesleben_2021_IntriguingRelation}. 
From a compositional perspective, 
a pertinent question is whether a model's structure is asserted to \emph{correspond directly to structure in the phenomenon the model is about} or not. The answer has a bearing on the types of questions and interpretability concerns that can be addressed with the model. 

While not formal definitions, we introduce the following terminology to aid clarity.
\begin{center}
    \begin{minipage}{0.8\textwidth}
        \begin{itemize}
            \item[\textit{M-type}] Designates questions and purposes, as well as corresponding explanations and notions of \ity, that concern only the model itself.     
            \item[\textit{W-type}] Designates questions and purposes, as well as corresponding explanations and notions of \ity, that (explicitly or implicitly) refer to the world, phenomenon or process that is being modelled.
        \end{itemize}
    \end{minipage}
\end{center}

One might argue that any notion of \ity\ is of M-type -- it can only concern the \ity\ \textit{of a model}. Whereas questions that invoke the world, like asking what intervention will achieve some goal, can only concern what one does \emph{with} a model's \ity. However, it may be useful to extend the distinction to notions of \ity\ and explanations. Explicitly declaring a post-hoc explanation as M-type may help to prevent misuse, such as users confusing M-type statements like `prediction $p$ was caused by feature $x$' for ones that can guide actions.  
Freiesleben writes ``we need to be clear about whether we want to [explain] the model or the modeled process. [...] We can only move from a model [explanation] to a process [explanation] if the model itself, and also the translation of our inputs, preserve the essential structure of the process" \cite{Freiesleben_2021_IntriguingRelation}.  

The distinction is particularly relevant for causal structure. Suppose we have the ideal case of a causal model with a complete interpretation, either given explicitly or somehow induced by a neural model as discussed in Section~\ref{subsec:causalXAI-signalling-etc}. Even in this case, the causal structure is usually simply the structure of the model (M-type), and a priori has nothing to do with the causal structure between the variables in the world (W-type). While this seems clear enough, the causal flavour of the common XAI explanations can easily be misunderstood as causal statements about the world and thus lead to inadvertent misuse. Consider for instance the following situation.

\begin{example}
Suppose a model predicts the probability of suffering a stroke within 5 years based on the input features of age, profession, exercise habits and BMI.
Also suppose that a person is predicted to have a stroke with probability 15\% and that some version of LIME outputs the feature importances for this prediction as $-0.3$ for age, $+0.2$ for profession, $+0.2$ for exercise habits, $-0.1$ for BMI. Assume also that the model in fact perfectly follows the ground-truth distribution, and so 15\% is indeed the true probability of a stroke given these features. 
Even then, the output of LIME may be interesting in telling us (local, approximate) \textit{dependencies in the model}, but it would be foolish to infer actions  \textit{in the world} to reduce the risk of stroke on its basis. 
This is because altering any of these variables in reality may well affect the others, and the influence between some feature and stroke likelihood in the model might have picked up on a spurious correlation, while the influence in the world is mediated via other variables; the model has simply not been trained to address interventional questions on how to reduce stroke likelihood. This has nothing to do with a limitation of LIME, but is the result of a category mistake. 
\end{example}

The care needed in distinguishing between purely associational questions and causal queries is by now an old and well understood point -- see Sec.~\ref{subsec:framework-of-causal-models} for the causal hierarchy.  
Also, the role of \emph{action-guiding explanations} has specifically been emphasised in XAI.  
Beckers argues that ``an important goal of [XAI] is to compensate for [the] mismatch by offering explanations about the predictions of an ML-model which ensure that they are reliably action-guiding" 
and that ``action-guiding explanations are causal explanations, [which require] knowledge of an additional model, namely a causal model of the target system" \cite{Beckers_2022_CausalExplanationsAndXAI}.

A further typical XAI example where the model-vs-world distinction matters, but is rather subtle, is our earlier loan application example.

\begin{example}[CFEs and the world]
Consider again Example \ref{ex:CFE-bank-example} of a bank's loan application system and the CFE for an applicant's refusal to receive a loan. When the model-vs-world distinction is discussed in the literature, this sort of example on loan giving is often presented as an example where the explanations from XAI methods, such as a CFE, \emph{are} about the world and not just the model. 
Indeed the model is \textit{by definition} the mechanism in the world by which the bank does in fact decide on the loan. 
However, while the model's relations between inputs and loan prediction are correct by definition, they don't capture the causal relations that actually obtain in the world, both amongst the input features and also between inputs and whether or not someone actually defaults on a loan (as opposed to getting a loan). 

This structural mismatch can easily lead to wrong conclusions and abuse. First, as is well-known, a customer may exploit the wrong causal relations in order to receive a loan when really nothing has changed that would make them more likely to actually pay it back -- say, by increasing the number of pets \cite{Freiesleben_2021_IntriguingRelation}. 
Second, 
CFEs cannot be taken to guide actions in the world.  
As we have seen above, 
an alternative input $x'$ that appears in the antecedent of a CFE may differ from the original input only in a subset $S$ of features, making it look as if the effect is achievable by suitably intervening on $S$. However, in the world this will generally change other variables, too, and not lead to the assignments as in $x'$.\footnote{Note that this is distinct from the observation that a CFE explanation may not amount to actionable advice due to \emph{operational limitations} because one can't intervene on, for instance, age (see \cite{Freiesleben_2021_IntriguingRelation}).} 
\end{example}

The M-type vs W-type distinction is not only relevant to causal structure. For instance, in models of the DisCoCat and DisCoCirc frameworks (Secs.~\ref{subsec:DisCoCat} and \ref{subsec:DisCoCirc}) the compositional structure is determined by the (grammatical) structure in the phenomenon of natural language. There may well be tasks and concerns that, analogously to the case of causal structure, require getting this structure right. We leave it for future work to identify further examples that demonstrate this form of structure-correspondence aside from the causal one. 

In summary, while there has been some focus on the distinction between associational and causal reasoning (e.g. \cite{Beckers_2022_CausalExplanationsAndXAI}), this doesn't acknowledge the fact that many XAI methods are closely related to causal explanations, and instead the key distinction is the status of the model's structure (causal or otherwise) -- does it correspond to the phenomenon or not? A formalisation of structure such as that offered in this article can help make this distinction sharper.



\section{Observations on compositionality and interpretability} \label{sec:comp-and-interp}

Having seen the basic ingredients of \rl{our compositional approach} (Sections~\ref{sec:compositional-models} and \ref{sec:frameworks}), numerous examples of AI models \rl{in this light} (Section~\ref{sec:examples-compositional-models}), as well as the discussion of \rl{pertinent themes in conventional XAI} (Section~\ref{sec:CI-and-causal-XAI}), let us now extract and summarise the key observations on the relationship between the two central concepts -- compositionality and interpretability. 

\paragraph{Many AI models can be represented in string diagrams.} 

We have seen that a wide range of AI models can be formally described as compositional models, which in practice means representing them as string diagrams. 
This also allows us to distinguish in a formal, yet very intuitive way a model's high-level structure (syntax) and the concrete instantiation (semantics). This is of independent interest aside from the interpretability angle this work focuses on, and helps reason about models and their properties.

\paragraph{Forms of compositionality are familiar in ML.}

While the language of categories and composition may seem novel, we have seen that some forms of composition are in fact ubiquitous in ML and data science. 
At the most basic level, \st{any situation in which we can describe a component of a model as consisting of various aspects or factors is usually an instance of compositionality.} 
In this case, the composition is simply depicted as separate `wires in parallel', with a wire for each aspect. For example, suppose we have a representation space $X$ which we know factorises in terms of $n$ dimensions or features $X_1, \dots, X_n$. In diagrams this appears as below. 
\begin{equation} \label{eq:factor-1}
\input{factor-2.tikz}
\end{equation}
In a neural network model, this means that $X = X_1 \times \dots \times X_n$ is given by a Cartesian product (`direct sum') of the features $X_i$; for example, each $X_i$ could simply be given by $\mathbb{R}$ with $X = \mathbb{R}^n$.\footnote{In a quantum model \rl{a diagram such as the above would mean $X = X_1 \otimes \dots \otimes X_n$ is given by the tensor product of Hilbert spaces $X_i$ and in a statistical model by the tensor product of (classical) probability spaces $X_i$.}} 

As another example, suppose we have a model trained on a data table of staff members, consisting of columns for `age', `gender' and `role', which means the input space factorises in terms of wires for each of these. The data itself is given by a distribution over these factors, and so appears as below.
\begin{equation} \label{eq:factor-2}
\input{data-factor-state.tikz}
\end{equation}

Beyond decomposing a wire (a single space), more general string diagrams, \rl{which allow us} to decompose a process, such as an entire input-output model, also feature in ML. For example, the computational graphs used to describe neural networks \ir{can be seen as}
string diagrams, as we saw in Section~\ref{sec:examples-compositional-models}.

\paragraph{Explicit compositional structure underlies, but doesn't imply, interpretability.}

In general, drawing a string diagram can be seen as a first step towards `making sense of' a model, but \rl{does not necessarily} provide it with an interpretation. Once we have a diagrammatic account of a model, we then can ask whether or not its components do in fact have an interpretation in our sense (Sec.~\ref{subsec:interpretations-of-models}). 
Such an interpretation may be partial (concerning only a subset of the components),  abstract (only assigning meaningful names to the respective components), or  concrete (essentially, also interpreting the variables' state spaces).

For some models the diagram will consist of meaningful interpretable processes, such as causal mechanisms or the conditions in a decision tree. 
However, for most ML models this diagram will consist of components which are not a priori necessarily interpretable. This applies both to models whose diagrams are very low level, such as neurons in a network, or more high level, as in a transformer diagram which contains attention heads whose function is not interpreted. 
For example, the decomposition \eqref{eq:factor-1} need not necessarily make the space $X$ more interpretable; it may or may not be that the features $X_i$ have an interpretation to us. 
In contrast, in \eqref{eq:factor-2} we would expect each variable such as `Age' to come with both an abstract and concrete interpretation, with each specific value interpreted as a specific age. 

So whether or not a model has an interpretation is \emph{not} part of its definition \emph{as a compositional model}. Indeed, one and the same kind of model, be it a linear model or some deep NN architecture, may be interpretable in one situation, but not in another. 
An interpretation is extra data -- wherever it may come from. 
Moreover, the definition of an interpretation (Def.~\ref{def:interpretation}) is broad with the signature $\Human$ of human-friendly terms not prescribed. It is down to the researcher, user or explainee to decide what they regard as giving meaningful interpretations in the given context -- in keeping with the generally acknowledged degree of subjectivity in judging a model's interpretability. 

\paragraph{Standard XAI methods -- interpreting models from the `outside'.}

Even standard \rlc{black-box models} do in fact typically come with some interpretable compositional structure, namely that of their inputs and outputs. Indeed these relate to training data, which itself is often concretely interpretable. 
If so, the overall input to the model thus factorises in terms of individual concretely interpreted input factors, as does the output similarly. Hence, for any such model, the level at which we may draw an interpretable diagram is simply from this `outside' perspective, as below. 
\begin{equation} \label{eq:black-box-in-out}
\input{blackboxinsouts.tikz}
\end{equation}
For example, this might be age, gender and role as in \eqref{eq:factor-2}.  
For a transformer trained on text, from this perspective the inputs would be the individual word labels (rather than their embeddings), while the output could be the predicted next word. 

Since the diagram in \eqref{eq:black-box-in-out} is often  \emph{the only one} of the model we may draw whose variables \st{are immediately interpretable}, and seeing as `explanations' are only explanatory if referring to interpreted terms, it is also the level at which standard XAI methods tend to be applied. 
Indeed, as Section~\ref{sec:CI-and-causal-XAI} argued, most kinds of explanations provided by post-hoc XAI methods can be stated solely in terms of inputs and outputs and usually \rl{convey (or rather, approximate)} causal properties of the model --  essentially, `how does input $j$ affect output $k$?' 
Examples are methods that use Shapley values, in order to answer `which input wire is most important to this output?', or that produce counterfactual explanations. 
However, several criticisms of the `explanations' provided by such methods have been \rl{given; see \cite{Rudin_2019_StopExplainingBlackBoxes, FreieslebenEtAl_2023_DearXAICommunity} and Section~\ref{subsec:Issues_with_XAI} for a summary.} 
\rl{Because of such limitations}, we argue that richer, more principled explanations will require us to look \emph{inside} the model, requiring a model with interpretable \emph{internal} compositional structure. 


%


\paragraph{Intrinsic interpretability is diagrammatic.}

We have seen that the standard examples of intrinsically interpretable models, namely rule-based and linear models, can be naturally depicted in string diagrams (Secs.~\ref{subsec:linear-models} and \ref{subsec:rule-model}). 
The components made visually explicit are the very aspects that are also considered to make the model interpretable. 
For example, the diagram for a decision tree contains boxes for each of its conditions, which are readily interpretable functions (e.g. `Age $<$ 20'), while the diagram for a linear model contains the weights, interpreted as relevance. 

\st{Note that the interpretation is not formally a part of the definition of either linear or rule-based models, but it is}
understood implicitly that the contexts where they are used ensure the existence of a complete and concrete interpretation. For example, a linear regression model is trained relative to a data science problem with all variables having a concrete interpretation as a given, while a rule-based model is understood to have rules that individually do make sense to us. Furthermore, intrinsically interpretable models are usually only considered to be such subject to \st{a sparsity constraint, meaning they are not} too `large' for humans to \emph{easily} trace decisions.
This condition though is not part of an interpretation in our sense. Obviously, it is an important question whether a model is simple enough so that a human gets it `at one glance', but \st{from this work's perspective} this is a question of sparsity and simplicity and not about interpretability as such. 
Even if a rule-based model is `huge', as long every variable and process has a concrete interpretation, in the worst case one may just need \st{an analysis tool 
to help inspect and trace decisions,} but the interpretability is just as manifest and inherent as with a small model.

The literature actually does not give a definition of intrinsic interpretability, but only lists canonical examples. 
\rl{In our terms} they are models with a complete, concrete interpretation subject to the additional sparsity condition.  
Hence, our notion of interpretation is consistent with the one prevalent in the XAI community, for the case of these models,  
but extends it to further kinds of interpreted compositional models.

\paragraph{Compositional frameworks -- interpretability beyond interpreted structure.}

A further aspect of a model's compositional structure that contributes to how (usefully) interpretable it is, but that is not captured by the mere presence of an interpretation \st{in our sense, is the status of the structure. This structure may or may not} license computing quantities of interest by combining its components in novel ways. This aspect is captured by the kind of compositional framework in which a model lives, independently from its interpretation. 

The main examples, discussed in Section~\ref{sec:frameworks}, are: 
input-output models, which despite a possibly explicit and complex compositional structure only allow for the production of outputs given inputs; 
statistical models, which additionally allow for the computation of conditionals; 
and causal models, which allow for interventions, computing counterfactuals and treating causal identifiability problems. 
Indeed, a sparse linear regression model for a loan prediction problem may be a vanilla version of being intrinsically interpretable, but the components themselves have no meaningful relevance to us on their own -- unlike a causal model for the same problem.

\paragraph{Model vs world -- compositional structure of what?}

An increasingly popular perspective sees standard post-hoc methods couched in the \st{language of causal models}. 
Indeed, as discussed in Sec.~\ref{subsec:causalXAI-signalling-etc}, any deterministic model induces a deterministic open causal model and many XAI methods are aimed at revealing causal properties of the given model. 
However, even if the causal structure was explicit and perfectly known, it is causal structure of \emph{the model}, which a priori (and indeed typically) has nothing to do with the causal structure in the world or phenomenon the model is about. 
One must not confuse the two, and only when there \emph{is} a structural correspondence, can XAI explanations, and the model more generally, be action-guiding in the world. 

\st{Related to the model-vs-world distinction are a line of arguments claiming that when a model \emph{has} picked up relevant modular structure in the phenomenon (such as causal structure), then this structure should stay largely unaltered between closely related problem instances, and the model itself should be more robust, train efficiently and exhibit strong generalisation  \cite{SchoelkopfEtAl_2021_TowardCausalRepresentationLearning}.
}

\paragraph{\rl{Compositionally-interpretable models.}}

Bringing together our observations so far, we claim the following. 

\begin{center}{\begin{minipage}{40em}
	Most ways one can `make sense of' an AI model amount to providing it with a \emph{string diagrammatic description}, i.e.~viewing it as a compositional model. 
	Interpreting the AI model then amounts to providing the compositional model with an \emph{interpretation} in our sense. 
	\rl{Models with a complete interpretation, which we refer to as \emph{\CI}\ (CI) models, generalise the notion of \emph{intrinsically interpretable models} and include in particular causal models. 
	Compositional interpretability is especially useful when the interpretation is moreover concrete and the interpreted structure corresponds to \emph{structure in the phenomenon}.} 
\end{minipage}}\end{center}

%

Starting from an AI model we choose a `level of abstraction' at which to understand it as a compositional model (e.g. from neurons at the lowest level, to potentially a single black box at the highest), which amounts to expressing it in terms of string diagrams. 
This does not itself make a model interpretable; for example we can always draw any neural network as a (huge) string diagram. 
However, if the components of the diagram(s) for a model are interpretable, we can say that the model itself is. 

If the interpretation is a concrete one and complete, i.e. covers all generators, one may see the interpreted diagram describing the model as an `explanation' in itself. 
This view on what we call a CI model is consistent with the perspective that an (interpreted) causal model is `its own explanation', as well as our observation that rule-based and linear models are manifestly interpretable from their own diagrams. 

In fact, this view is also consistent with much existing work in the XAI literature, in which a model is specified in terms of a computational graph, and attempts to understand it amount to attempting to interpret the components of this diagram.  
For example, methods exist to attempt to assign an interpretable `concept' to each neuron in a neural network, or attention head in a transformer \cite{hewitt-manning-2019-structural}, both of which can be understood as aiming to assign abstract interpretations to these components of the diagram. 

If it can be made possible to assign rich such interpretations post-hoc to trained networks, then this would indeed yield CI models, and arguably solve the main problems for XAI. 
However, this is unlikely due to various limitations for example pointed out in \cite{FreieslebenEtAl_2023_DearXAICommunity}, including the fact that there is no reason for neural networks to independently use the same concepts as humans, as well as that features such as dropout encourage global, not local, representations of information in neural networks, and so concepts are unlikely to be located in specific parts of a network.  Practically, this process of assigning meanings post-hoc following training is also highly effort intensive. 

While it is arguably uncontroversial that CI models have an ideal form of interpretability, they are not easy to obtain. 
Yet, there are at least two options for employing the concept in practice. 
First, one may weaken the requirement that the model that solves a certain task of interest must itself be a CI model, but rather ensure that it behaves approximately as such, for instance through causal abstraction approaches (Sec.~\ref{subsec:causal-abstraction}).
Second, and maybe most importantly, following the voice of, e.g., Rudin \cite{Rudin_2019_StopExplainingBlackBoxes}, there are situations in practice where the stakes are high enough to justify simply biting the bullet. In these cases, the cost and effort of building a case-specific CI model -- combining expert knowledge, causal discovery algorithms, and whatever else is required -- is worth it and should be investigated more often than currently done so.



\section{Explanations from diagrams} \label{sec:expl-from-diags}

\st{ Our analysis of a range of compositional models and aspects of interpretability led us in the previous section to consider the class of compositionally-interpretable  models. These come with a complete interpretation, and ideally a concrete one, and include the classic intrinsically interpretable models, as well as causal models.\footnote{That is, a causal model in the W-type sense (Sec.~\ref{subsec:model-vs-world}) as originally understood in the causal model literature; to be distinguished from the causal model that any NN model can be seen to induce, which does not a priori yield a CI model.} While such models can be regarded as `their own explanation', it is important to ask: how exactly can the internal diagrammatic structure of a model provide explanations for the specific outputs it produces?}

\st{
In this section we make this concrete by discussing several ways in which interpreted diagrams allow reasoning about the behaviour of a model. These are \emph{structural influence constraints}, which allow reasoning about which inputs affect which outputs; \emph{diagram surgery}, which allows us to act on, or alter the model directly; and finally \emph{rewrite explanations}, which use diagrammatic reasoning to give direct accounts of, and constraints on, particular outputs of a model.}

\subsection{\rl{No-influence arguments}}
\label{subsec:expl-signalling}

As we have seen, perhaps the most prevalent and simple form of composition is the factorisation of a wire as a product of factors. 
Whenever the inputs (and possibly also the outputs) of a process can be decomposed in this manner, we can then reason about which input factors can affect which output factors.
\st{Recall that a channel $c$ from  $A, B$ to $C, D$ exhibits no-influence from $B$ to $C$, so that interventions on $B$ cannot influence $C$ through $c$, when the following holds:} 
\[ 
\input{nosignal.tikz}
\]


The presence of no-influence relations of a (process within a) model provides constraints on which inputs affect which outputs, which can greatly simplify the search for explanations or be explanatory themselves. 
For diagrams consisting of channels, no-influence relations \rl{may in particular} be found by simply examining the \emph{connectivity} structure of the diagram. 
\rl{Indeed, whenever an input variable $X$ has no directed path (reading bottom up) to output variable $Y$ in the diagram, then $X$ cannot influence $Y$ in the overall channel represented by that diagram. 
This fact becomes evident diagrammatically by `letting discards fall through' \cite{kissinger2017equivalence} as in the following example, where there is no directed path from $X_4$ to $Y_1$:} 
\begin{equation} \label{eq:nosignal-in-circuit}
	\input{nosignal-in-circuit.tikz}
\end{equation}
\rl{As a result, models with non-trivial compositional structure come with the added benefit of non-trivial influence relations that can simply be read off the structure of the diagram.} 



\st{A classic example are causal models, where the influence constraints from the absence of paths in the DAG (or equivalently, network diagram) provide great reasoning ability. \rl{Indeed, these constraints are part of the intuitive and explanatory role such causal structure is understood to have.} 
\st{While the concept of no-influence cannot be applied directly to a 
conventional (closed) causal model without inputs,}
 there are closely related no-influence type of statements that can be inferred from the structural properties of such a model, using the same sort of `absence of path' argument as above. For instance, whenever $X$ is not an ancestor of $Y$ in $G$, then in the corresponding network diagram there is no directed path from $X$ to $Y$ and one can easily see, analogously to Eq.~\ref{eq:nosignal-in-circuit}, how an intervention on $X$, say a do-intervention $do(X=x)$, cannot have any effect on the output of $Y$.\footnote{\rl{This fact can also be seen to be a special case of the third rule of the do-calculus rules \cite{pearl2009causality}.}} 
See the following example for an illustration.}

\begin{example} \label{ex:no-influence-in-causal-model}
	\rl{Given a causal model with network diagram as on the left below (also studied in \cite{lorenz2023causal, ShpitserEtAl_2008_CompleteIdentificationMethodCausalHierarchy}) consider a do-intervention $do(X=x)$ and marginalise all variables but $Y$. 
	As is straightforward to see through the below simplifications, the absence of a path from $X$ to $Y$ in the network diagram yields that $\Do(X=x)$ has no effect on $Y$.\footnote{\rl{In the causal model literature this would be stated as the causal effect quantity $P(Y; \Do(X))$ in fact not depending on $X$ and just reducing to the marginal $P(Y)$.}}} 
	\begin{center}
	\input{Sec-9-1-causal-model-example.tikz} 
	\hspace*{0.5cm} $\mapsto$ \hspace*{0.5cm} 
	\input{Sec-9-1-causal-model-example_2.tikz} 
	$=$ \hspace*{0.3cm}
	\input{Sec-9-1-causal-model-example_3.tikz} 
	\hspace*{0.3cm}	$=$ \hspace*{0.3cm}
	\input{Sec-9-1-causal-model-example_4.tikz} 
		\end{center}
\end{example}



\st{
A further example of no-influence argument of a different kind is the following.}

\begin{example} \label{ex:Dcirc-signalling}
Consider a DisCoCirc model, with a verb such as \natlang{likes} in the phrase \natlang{Alice likes Bob}, represented by a channel. The word `likes' will act only on the Alice and Bob wires, and as the identity on any other discourse referents, with the following representation. The resulting channel will only allow signalling between the Alice and Bob wires with all other wires only signalling to themselves. 
\[
\input{likesfactor2.tikz}
\]
For certain forms of text input, this can induce desirable properties of the model. Given a collection of facts relating several agents, and arranged in a chronological order, such as \natlang{Alice hired Bob and then spoke to Claire}, we can consider a DisCoCirc model \rl{of the form below,} which describes the resulting updates of information about each agent. If we wish to understand the final state of \natlang{Bob}, we see that this will not depend on that of \natlang{Claire}, due to the absence of signalling. 
\[
\input{change-dcirc2.tikz}
\]
Intuitively, this is because the interaction between \natlang{Alice} and \natlang{Claire} happens after the final action relevant to \natlang{Bob}, and so should not affect them.\footnote{Note however that this behaviour may not be desirable for arbitrary text input which is not of this `sequence of facts' kind, in which \rl{the sentence or word order in a text may be considered irrelevant}. For such situations, channels may not be appropriate as representations for each word in the model.}
\end{example}

Recall from Section \ref{subsec:causalXAI-signalling-etc} that for conventional architectures such as fully connected neural networks, there typically are \st{\emph{not} any no-influence constraints that are enforced structurally. Hence, the kind of reasoning in terms of strict no-influence arguments is a benefit of models with non-trivial compositional structure, that is, where the connectivity in the diagram is \emph{not} all-to-all. }

We will see more examples of \rl{no-influence arguments} shortly, combined with the forms of explanation we consider next.

\subsection{Diagram surgery} \label{subsec:surgery}

One way to understand the utility of a compositional structure is to see each internal wire in a diagram as \rl{a point} where we may `inspect' or `intervene' on a process, in order to learn more about it. \rl{More generally,} we can learn more about how a diagram leads to a particular output \st{by acting on and altering fragments of the diagram}, either to see how this affects the output \rl{or, broadly speaking, to inspect the process itself.} 

Following the terminology of \cite{jacobs2019causal},  given any diagram $\diagD$ in a category $\catC$, by an act of \emph{diagram surgery} we mean the generation of a new related string diagram $\diagD'$ in $\catC$, \rl{where some fragments of $\diagD$ have changed}.  When $\diagD$ is itself seen as describing a compositional model, then this amounts to an alteration to a new (typically closely related) model. 
A common case of surgery is that which alters a single process (box) in our original diagram, which we may illustrate as follows.
\[ 
\input{diag-surgery-highlight.tikz}
\]
Note that \st{in the above illustration the blank boxes may themselves have complex internal structure, and that the new box $f'$ may have distinct input and output types from $f$, as may the overall diagram $\diagD'$ from $\diagD$.}
Alterations to a diagram can provide various forms of explanations about a model, and the kinds they provide will vary depending on the level of interpretability of the variables on which the surgery acts. Let us now discuss several useful forms of diagram surgery.  


\paragraph{Local surgery.}
Suppose we have some variable $\syn{V}$ occurring within a diagram, which lacks a concrete or even abstract interpretation, but which we would like to learn more about. While we cannot interpret the variable directly, we can instead relate it to further variables $\syn{C_i}$ via some channels of the form $(c_i \colon V \to V \otimes C_i)^n_{i=1}$, where now each variable $\syn{C_i}$ does have a concrete interpretation for each of its states. Applying each such channel amounts to the kind of \emph{local surgery} on our diagram depicted below.
\begin{equation} \label{eq:probe-picture}
\input{probe-picture2a.tikz}
\end{equation}
In general, the channels $c_i$ may alter or `disturb' the state of $V$; for example this will typically be the case for local surgery in quantum models, which we \rl{focus on} in Section \ref{sec:quantum}. However, \st{in typical} classical models, we are free to copy (sharp) states without disturbing them. Hence we may simply \emph{observe} any variable through channels $c_i$ of the following form, which amount to copying the variable and applying some channel $d_i \colon V \to C_i$.  
\begin{equation} \label{eq:copy-probe}
\input{classical-probe.tikz}
\end{equation}

For a representation space $V$ inside a neural network, the channels $d_i$ may be given by a set of classifier networks, where each classifier  $d_i$ maps a state $v$ of $V$ to one \rl{$d_i(v)$} of a finite set of interpreted classes $C_i$, which are taken to describe aspects of $V$. 

\begin{example} 
Consider an RNN, which makes use of an uninterpreted representation space $X$ \rl{and produces an output by applying a classifier from $X$ to a finite set of interpreted classes $C$. More generally, we may train the same representation to come with a family of classifier maps $C_i$ each of which is used in some task. To understand how the state of $X$ evolves as the RNN is applied, we can probe the wire using any of these classifiers $d_i$ as below.}
\[
\input{probe-RNN.tikz}
\]
Now rather than only receiving a class for the final output, we can see how the class is altered as each word of the text is read, to indirectly interpret the model's representation. For example, we may see how the overall sentiment $C_i = \{ \text{positive}, \text{negative} \}$ of the movie review, produced at the final output, is altered by each word such as `not' and `good'. 
\end{example}

\paragraph{Interventions.} 
Another major form of surgery lies in deliberately altering the value of a variable, to inspect the changes this causes to the model. Firstly, given any variable $\syn{V}$, we can \emph{intervene} by applying some fixed channel $c_i$ to $V$, replacing the $V$ wire in our diagram by $c_i$.\footnote{Note this is a special case of local surgery with $C_i = I$.} 
\rl{Provided an interpretation for this intervention $c_i$ itself is given, we can observe the resultant changes to $V$ and to the entire model in order} to attempt to interpret $V$.
\[
\input{localint.tikz}
\]
Indeed we already met interventions in the context of causal models in Sec.~\ref{subsec:causal-models}. The special case of a \emph{do-intervention} $\Do(V=v)$, in which \rl{the variable $V$ is set} to some particular fixed state $v$, can be seen to be equivalent to \rl{the following choice for $c_i$}, in which we simply discard $V$ and replace it with $v$. 
\[
\input{do-int.tikz}
\]
More broadly, we saw how we can go \rl{beyond do-interventions}, defining an intervention on a causal model to be any replacement of the mechanism for a variable (or multiple variables) \rl{so long as the overall model} remains a valid causal model. At the level of the network diagram describing the causal model, this involves a form of diagram surgery \cite{jacobs2019causal} in which we replace the box for mechanism $c_i$ for variable $\syn{X_i}$ with another for a new mechanism $c_i'$:
\begin{equation} \label{eq:cm-ints}
\input{causal-ints.tikz}
\end{equation}
We then connect all inputs of $c_i'$ to the new parents of $X_i$ via copy maps. For a detailed discussion of interventions on causal models in string diagrammatic terms, see \cite{lorenz2023causal}.

 By analogy, we can think of any diagram surgery \rl{that replaces boxes within a diagram as a form of `intervention' on the model. \st{This general notion of intervention is in spirit the same as that of the causal model framework, but now the class of diagrams and (hence models) is generalised, no longer restricted to network diagrams (i.e.~causal models).} }

\begin{example}[Intervening on inputs and CFEs] 
A special case of do-style interventions \rl{is} simply the altering of the inputs to a given process. \rl{This forms the basis of many} standard XAI methods, such as the CFEs discussed in Section \ref{subsec:CFEs-vs-CFs}. Modifying the inputs to a process $f$ \rl{is} the following basic form of surgery: 
\[
\input{change-inputs.tikz}
\]
\rl{In this light we can thus see explanations that involve altering internal components of a model} as generalisations of such explanations. 

\end{example}

\st{Diagram surgery as such can always be applied to a model. What makes it explanatory is when the surgery concerns variables (wires) or larger fragments that have an interpretation and are replaced with new fragments that also have an interpretation.  This distinguishes \CI\ models from black-box models with uninterpreted intermediate components. 
In particular, we may combine diagram surgery with influence arguments to provide explanations for the outputs of CI models, as in the following example.}  

\begin{example}[\rl{Influence constraints} and Surgery]
Let us return to our DisCoCirc model of the text \natlang{Alice hired Bob and then spoke to Claire} from Example \ref{ex:Dcirc-signalling}. Suppose we follow with a question \natlang{Is Bob employed?} and as expected receive the answer \natlang{yes}, and would like to explain how this answer was reached. As before, the structure of the diagram tells us that there will be no-signalling from Claire to Bob and so restrict analysis to the left of the diagram.
\[
\input{change-dcirc.tikz}
\]
To check that the model arrives at the answer in the way we would expect, we can apply diagram surgery by either removing the process for \natlang{hired} or by replacing it with another word such as \natlang{fired}, and verify that these lead to a distinct answer.
\[
\input{change-dcirc2a.tikz}
\]
How does this analysis compare with simply inserting the alternate texts as new inputs to a fixed black box? Firstly, the \rl{influence constraints} have provided guarantees that certain parts of the model won't affect the output (the Claire wire). Secondly, we may as a result restrict to the sub-diagram \rl{above, which provides a localised explanation and may be more efficient to analyse than the entire model}. 


\end{example}

\begin{example}
Consider a model with inputs $X_1, X_2$ and output $O$ which on inputs $(x_1, x_2)$ gives output $o$. The standard form of (counterfactual) explanations \rl{may for instance be of the form} `if we change the value of $x_2$ to $x'_2$ we get the output $o'$ instead'. 

\st{When the model comes with further interpreted compositional structure, we can give a richer form of explanation still.} 
\rl{
Suppose the model factorises as an open causal model via intermediate variables $V_1, V_2$ as below, which have an interpretation. 
Also suppose that inputting $(x_1, x_2')$ yields in particular $V_2=v$ and suppose that through intervening on $V_2$ we discover that regardless of $V_1$ in fact $V_2=v$ necessarily yields $o'$:
\[
\input{causal-model-before-int_v2.tikz}
\]
Thus we now have a more fine-grained explanation, telling us that the change of the output to $o'$ really came down to changing the variable $V_2$, which happens to depend on $X_2$.}\footnote{ \st{Note that the need for care in distinguishing causal structure in the model vs in the `world', especially with a view to action-guidance (as pointed out in Sec.~\ref{subsec:model-vs-world}), applies here just as much as with explanations in terms of inputs and outputs only.
}} 
\end{example}

\subsection{Rewrite explanations} \label{subsec:rewrite-expl}

Our final notion of explanation from diagrams is the most direct, \rl{but} also the most speculative. It is based on diagrammatic `rewriting', a central aspect of the theory of monoidal categories, in which one applies successive equations between pieces of a diagram to prove that it is equal to another. We will aim to demonstrate how rewriting can be applied to models with interpreted compositional structure to provide explanations for their outputs, which we call  \emph{rewrite explanations}. Note however that the ability to give such arguments in practice remains to be demonstrated; we return to this at the end of the section.  

Let us now motivate this approach. Consider a situation where \rl{a compositional model specifies an overall process} $f$ from inputs $X_1,\dots,X_n$ to outputs $Y$. \rl{Suppose that given} input $x_1,\dots,x_n$ it produces \rl{output} $y$, which we would like to explain. 
To do so, we first
\st{consider the `internal structure' of the process $f$ given by the compositional model,} as in the first equality below.
Next, suppose we know of some equations which we may apply to sub-diagrams of this diagram, as in the second equality below. By continually applying such rewriting we find we can arrive at the output $y$. 


\begin{equation} \label{eq:diag-expl-motivation}
\input{rewrite-illustration.tikz}
\end{equation}

Any compositional model could potentially allow for diagrammatic rewriting from the input to the output in this way. However, for this to qualify as an \emph{explanation} for the output $y$, the structure in these diagrams must itself be \rl{\emph{interpreted}}. That is, the inputs $X_j$ and output $Y$ should be (at least abstractly) interpretable, the states  $x_1,\dots,x_n$ concretely interpretable, and all of the components of the diagrams and sub-diagrams used in the above argument should be interpretable also. If that is the case, we consider the argument above as a rewrite explanation. Let us now spell this out formally. 

\begin{definition}[Interpreted Diagram]
Let $\modelM$ be a compositional model with \rl{generators $G$ and} interpretation $\Interp$. An \emph{interpreted diagram} is a string diagram $\diagD$ in \rl{$\catC_\Sig$ (see Def.~\ref{def:interpretation}),} for which $\Interp$ is defined on all variables and boxes in the diagram.
\end{definition}

Explicitly, an interpreted diagram consists of wires labelled by variables $\syn{V}$ of the model for which the abstract interpretation $\Interpabs(\syn{V})$ is defined, and boxes $f \colon \syn{V_1}, \dots, \syn{V_n} \to \syn{W_1}, \dots, \syn{W_m}$ corresponding to morphisms $f \colon V_1 \otimes \dots \otimes V_n \to W_1 \otimes \dots \otimes W_m$ in $\catC$ for which the concrete interpretation $\Interpcon(f)$ is defined.\footnote{More formally, for which $\Interpcon(f)$ is defined on the corresponding morphism $f \colon (\syn{V_i})^n_{i=1} \to (\syn{W_j})^m_{j=1}$ in $\catC_\Sig$. } 


We can now define rewrite explanations themselves. For any diagram $\diagD$ in $\catC$, we denote the induced morphism in $\catC$ from its inputs to outputs by $D=\sem{\diagD}$.

\begin{definition}
[Rewrite Explanation]
Given a compositional model and two diagrams $\diagD, \diagD'$ in $\catC$, a \emph{rewrite explanation} for $(D = D')$ is given by a list of equations between interpreted diagrams
\begin{equation*} \label{eq:diag-equations}
\left(D_j = D'_j \right)^n_{j=1}
\end{equation*}
along with a rewriting proof showing that these imply that $D = D'$. Similarly, a rewrite explanation for an approximate equality $(D \approx D')$ is given by a rewrite proof using a collection of approximate equations between interpreted diagrams $(D_j \approx D'_j)^n_{j=1}$.
 \end{definition}
 
\rl{The basic idea of} a rewrite explanation is as outlined above in \eqref{eq:diag-expl-motivation}, but we will see many more forms of example shortly. Before this, we note two aspects. Firstly, the method involves looking `inside' \rl{a model's overall process} and so requires compositional structure. As such this method is unique to compositional models and cannot be applied to a complete black box \st{in the sense of Sec.~\ref{subsec:XAI_in_nutshell}, that is, when the underlying process (model parameters) is unknown \stb{(e.g.~in a proprietary system)}.}  Secondly, to justify the term `explanation' we have required that the intermediary diagrams $\diagD_j$, $\diagD'_j$ are themselves fully interpretable. Indeed, otherwise one could for example view an entire uninterpreted neural network as a diagram and write (hundreds of) diagrams tracing the input through each neuron in each layer, and this would count as an `explanation' for the eventual output. 
\st{Thirdly, note that at a diagrammatic level rewrites can be seen to involve surgery. However, in Sec.~\ref{subsec:surgery} surgery was used to reason about an \emph{inequivalent} diagram or model, while in rewrite explanations diagrammatic fragments are replaced with ones that are equationally guaranteed to (approximately) represent the same morphism in $\catC$.} 

\begin{remark}[Approximate equalities of diagrams]
As well as strict equations $D = D'$ we will consider approximations of diagrams $D \approx D'$. The latter require the semantics category $\catC$ to come with a notion of approximation $f \approx g$ for morphisms $f, g \colon X \to Y$, which should moreover be quantified, and respected by composition, so that approximate rewrites of sub-diagrams can be extended to an entire diagram. For simplicity we will treat these informally here, but note that these approximations and their properties can be made formal, for example as done for quantum processes in \cite{kissinger2017picture} \footnote{One natural approach, used in \cite{kissinger2017picture}, is to have that each homset $\catC(X,Y)$ embeds into a normed space, with $\| g \circ f \| \leq \| f \| \| g \|$ and $\| f \otimes g \| \leq \|f \| \| g \|$. We then write $f \overset{\epsilon}{\approx} g$ whenever $\| f - g \| \leq \epsilon$.  }. 
\end{remark}

Let us now meet \rl{a suite of toy examples to illustrate the idea} of rewrite explanations.

\begin{example} \label{ex:rule-based-comp-ex}
Consider the decision list model \eqref{eq:decision-list} from Section \ref{subsec:rule-model}. Suppose that on a given input $(s,a,p)$, consisting of sex $s$, age $a$, and number of priors $p$, the model predicts arrest within 2 years, i.e.~outputs `yes'. An explanation for this output could take as its sub-diagrams the following:
\[
\input{rule-explanations.tikz}
\]
These diagrams have an interpretation as the statements `age is between 21-23' and `has 2-3 prior arrests'. An explanation is then given by the following rewrite proof (using the standard properties of the $\star$ and `first' boxes) that they lead to the output `yes':
\[
\input{rule-explanation-arg.tikz}
\]
\[
\input{rule-explanation-arg-2.tikz}
\]
 This argument simply follows the logical structure of the model to reach the output. Hence in this case the rewrite explanation corresponds to the standard sense of `intrinsic interpretability' for rule-based models, capturing how they intuitively give explanations by simply inspecting how they arrive at their conclusion.  
\end{example}


\omitfornow{
\begin{example} \label{ex:decision-tree-explanations}
Consider a compositional model for a decision tree, similar to that from Section \ref{subsec:rule-model}. The model decides whether to grant an applicant a loan. The input data on each applicant $X$ has information for their home ownership, savings and employment status. Suppose that for an applicant $x$ in this case a loan is not granted. An explanation could be given by the following sub-diagrams:
\begin{equation} \label{eq:tree-example-eqns}
\input{tree-xai-explanation.tikz}
\end{equation}
where we have written $\star = (x,`no')$ and $x = (x,`yes')$, and the following rewrite argument.
\[
\input{tree-xai-arg.tikz}
\]
\[
\input{tree-xai-arg2.tikz}
\]
Observe that the equations \eqref{eq:tree-example-eqns}  used have a direct interpretation as the statements `applicant is a homeower' and `applicant is not employed'. The rewrite argument simply corresponds to the usual tracing through of question answers through the decision tree. Indeed, tracing through the tree in this manner is how we might expect to find the equations \eqref{eq:tree-example-eqns} themseles. 

Hence a rewrite explanation such as this corresponds to the standard notion of `explanation' for an output provided by a decision tree. Other rule-based models provide a simple case of rewrite explanations by `tracing through' their inputs in just the same way. 
\end{example}
}

Tracing through the inputs in other rule-based models such as decision trees provides simple examples of rewrite explanations in just the same way. Like rule-based models, interpreted causal models also allow for simple rewrite explanations by tracing inputs through the diagram, as in the following example. 

\begin{example}
Consider the causal model for slipperiness of the floor outlined in Example \ref{ex:sprinkler-interpretation}. Here for simplicity we label the variables and boxes with their interpretations \rl{(see Rem.~\ref{rem:specifying-interrpetations})}.  Suppose that for the season `autumn' as an input, the model predicts that the floor will be slippery. An example rewrite explanation could be given as follows. Rather than listing the sub-diagrams used we will merely give the rewriting argument, from which they may be read off.
\[
\rlb{\input{slippery-xai-in-one_v2.tikz}}
\]
\end{example}

The following non-example shows how rewrite explanations are typically \emph{not} available for black-box models whose internal components are \rl{not interpreted}, such as typical neural networks or transformers.

\begin{example}[Non-Example]  \label{ex:comp-expl-nonexample}
Suppose we have a language model for deciding whether to grant a loan, which is only interpretable at the level of overall inputs and outputs. We observe that the model seems to consider employed homeowners to be `reliable' and that `reliable' applications should be given loans, in the sense that the following approximate equalities hold. 
\[ 
\input{blackbox-xai.tikz} 
\]
Despite this knowledge, we cannot derive a proof that an employed homeowner should be given a loan, 
\rl{because the corresponding approximate equality relating all components may or may not hold in general for this sort of model}:\footnote{Structurally, this stems from a lack of relation between the 2-input and 3-input instances of the model.}
\[
\input{blackbox-xai-2.tikz}
\]
Alternatively, one might instead try to give a rewrite explanation for a model such as a transformer, by starting from the diagram \eqref{eq:transformer} showing the attention heads, as in Section \ref{subsec:transformer-models}, and tracing the inputs through to the output. However, without interpretations for the components of this diagram this could not yield a rewrite explanation in our sense. 
\end{example}

In contrast, more richly compositionally structured NLP models do allow for such reasoning. The following example shows that even sequential composition, as featured in an RNN, can produce rewrite explanations. 

\begin{example} \label{ex:comp-expl-sequence-model}
Consider a text sequence model as in Section \ref{subsec:NLP-sequence-model} whose representation is used to determine whether an applicant is successful in applying for a loan. Suppose the input text states that the applicant is an employed homeowner, and suppose that their application is successful. A rewrite explanation for this output could include the following, where we find that following the information that the applicant is an employed homeowner, the model's representation is approximately that for a `reliable' person. Secondly, we find that a `reliable' person is typically provided with a loan. 
\[ 
\input{sequence-model-explanation-1.tikz} 
\qquad \qquad \qquad \qquad 
\input{sequence-model-explanation.tikz} 
\]
Combining these two diagrams together, we can apply rewriting to explain the outcome above, as follows. 
\[ 
\input{sequence-model-xai-arg.tikz} 
\]

\end{example}

In the next example, we make use of additional factorisation structure on the representation space of a sequence model.

\begin{example}
Consider a text sequence model whose wires are internally structured in the manner of \rl{conceptual space models} from Section \ref{subsec:conceptual-space-models}, containing representations of foods in terms of the domains of colour $C$ and taste $T$. Suppose the model answers that a yellow banana is not bitter. We might provide an explanation for this output by finding that the following equations hold, interpreted roughly as `yellow is a colour', `a yellow banana is sweet' and `sweet is not bitter'. 
\[
\input{consp-xai-1.tikz}
\qquad \ \ 
\input{consp-xai-eqns.tikz}
\qquad \ \ 
\input{consp-xai-3.tikz}
\]
Together these imply that asking whether a yellow banana is bitter yields the output `no', as follows.
\[
\input{consp-xai-argument.tikz}
\]
\end{example}

Text circuit models, such as DisCoCirc models, provide further examples of models with factorisation structure, where now the factorisation is over discourse referents rather than domains. These allow for rewrite explanations which relate agents in a text, as in the following examples.



\begin{example} \label{ex:dcirc-follows}
Consider a DisCoCirc model of the text \natlang{Bob is in the kitchen. Claire is in the Garden. Alice follows Bob. Where is Alice?'}. Suppose that the words in our model satisfy the following rules.
\begin{equation} \label{eq:Dcirc-comp-expl}
\input{text-circuit-xai-rules-2c.tikz}
\qquad  \qquad \quad 
\input{text-circuit-xai-rules-2b.tikz}
\end{equation}
The first states that when an agent follows another they are in the same location. The second states that if an entity is \natlang{in} another entity, and we ask where the first one is, the answer is given by the location of the second one. 
  These equations can give a rewrite explanation for why the model will always correctly answer that \dc{Alice} is in the kitchen, as follows.  
\[
\input{text-circuit-xai-2a.tikz}
\]
\[
\input{text-circuit-xai-2b.tikz}
\]
It is natural to ask: where would the rules in \eqref{eq:Dcirc-comp-expl} come from? One answer is that \rl{the} model can be trained in a manner to encourage them to hold, \rl{that is `enforce' them approximately by construction.} \st{Another is that} they may simply be found to hold experimentally following training. Alternatively, they may be \rl{`hard-coded' in an exact sense} into the structure of the model, for example as follows. Suppose that each wire factorises in terms of a `representation' wire, depicted in bold, and  a `location' wire depicted as a dashed line. Then the rules \eqref{eq:Dcirc-comp-expl} follow automatically from the following structure for our keywords. 
\[ 
\input{text-circuits-xai-semantics.tikz}
\]
\end{example}

\begin{example}
Consider a DisCoCirc model \rl{of the} text describing a western film plot from \cite{coecke2021mathematics}, shown below, involving the characters \dc{Harmonica}, \dc{Claudio}, \dc{Frank} and \dc{Snaky}. Suppose that we ask the question \natlang{Is Claudio alive?}, and receive the answer $n$ for `no'. An explanation can be given as follows, where for brevity we only depict the rewrite argument itself, and for this example depict the initial states $\star$ supplied to each wire. In the first and third step we use the fact that the boxes are channels. In the second step we apply a sub-diagram stating that \dc{Alive?} always returns `no' when following \dc{hanged}. 
\[
\input{western-states.tikz}
\]
\end{example}

\omitfornow{
\begin{example}
Consider a DisCoCat model upon whose representations a classifier is trained to determine whether applicants are eligible for a loan. Given the text `Alice is employed and a homeowner' the model grants a loan. An explanation could be given by observing that `employed and a homeowner' is approximately equal to an internal state which we found to be approximately equal to `reliable' (or an internal state interpreted as relating to relability), along with equations stating that `is reliable' always returns a successful application. The rewrite proof is then as follows. 
\[
\input{dcat-xai-basic.tikz}
\]

For example, a specific model may represent `is' and `and' via the Frobenius structure $\tinycopy$ similarly to \cite{Frobanatomy}, so that the rewrite proof looks as follows.  
\[
\input{dcat-style-xai.tikz}
\]
\end{example}
}

\paragraph{Rewrite explanations in practice.}
For intrinsically interpretable models, rewrite explanations include the familiar notion of passing an input through the model, with every step being interpretable and thus yielding an overall explanation. For further kinds of \rl{\CI\ models}, however, they are more novel but typically also require additional equations to hold between processes within the model. It remains to be shown in practice that one may train sufficiently rich \rl{CI models} coming with such equations. 
As discussed in Example \ref{ex:dcirc-follows}, this could be demonstrated either by training useful \rl{CI models}, which have equations as built-in structure or explicitly encouraged in the training function. Otherwise it would require showing that one can find such equations experimentally from a given compositional model which has not been trained to make the equations hold. Each of these approaches suggest directions for further experimental work in future, which would demonstrate that rewrite explanations can be useful in practice outside of the case of intrinsically interpretable models.

\section{Quantum models} \label{sec:quantum}

The categorical approach can capture models using a wide range of semantics categories, including that of \st{quantum processes}. 
In this section we focus our attention on quantum models and their relation to interpretability. We have seen several examples of quantum models, including quantum conceptual space models (Section \ref{subsec:conceptual-space-models}), quantum variants of RNNs including unitary RNNs (Section \ref{subsec:NLP-sequence-model}), and DisCoCat and DisCoCirc models in QNLP (Sections \ref{subsec:DisCoCat} and \ref{subsec:DisCoCirc}).

For concreteness, recall that a quantum compositional model will take its semantics in a category such as $\Quant{}$, the category of finite-dimensional Hilbert spaces and completely positive maps (see Section \ref{subsec:quantum-processes}).\footnote{Alternatively, a broader setting such as the category $\FCStar$ of finite-dimensional C*-algebras allows one to also include finite classical systems, which are often used to record measurement outcomes.} We may implement such a model using a quantum computer, where each variable is represented by some number of qubits, and each generator by a channel decomposed in terms of states, unitary gates, and discarding.\footnote{\rl{For certain models like DisCoCat (see Sec.~\ref{subsec:DisCoCat}) the generators include maps like the `cap', which are not channels, but instead non-trace-preserving completely positive maps leading to a quantum implementation, which additionally requires post-selection.}} For example, the diagram for a text circuit similar to Example \ref{ex:dcirc-follows} can be given a quantum implementation by representing each generator (word) by a \emph{parameterised quantum circuit}, where a separate set of parameters is stored for each generator. 
\[
\input{quantum-example-3.tikz}
\]

The utility of such quantum AI models, practically trained in quantum machine learning (QML) setups, remains a subject of ongoing research. 
\rlb{Suppose though there is a range of tasks in AI for which quantum models do prove valuable, then their interpretability may become an increasing concern just as for classical models.}

\rl{More specifically, we} claim that a compositional framework for interpretability, such as that offered here, will be helpful and in fact necessary to study interpretability of quantum models. Indeed, comparison against classical AI will be necessary and so a \rl{\emph{model-agnostic approach is required}}. Moreover, quantum models are naturally \st{defined} compositionally, being typically specified in terms of (quantum circuit) diagrams, and in fact formal diagrammatics has roots in quantum information and foundations \cite{abramsky2004categorical,coecke2018picturing}. 
Finally, it has been argued that \st{\emph{training compositionally} offers a way to circumvent serious obstacles to training quantum AI models arising from the Barren plateaus in typical parameter landscapes. To train compositionally here means to train local components in simulation, and combine these into larger circuits at test time.} 
An example of just such a compositional training setup is the recent quantum implementations of the DisCoCirc framework for QNLP \cite{QDisCoCirc}.

\paragraph{Interpreting quantum models.} 

What, then, can our setup tell us about interpreting quantum models? 
\st{Firstly, while one may attempt to argue that for specific domains, such as cognition or NLP, quantum semantics is especially natural to use and thus in a sense may be regarded as more interpretable, we will not do so here (for arguments for quantum models of cognition see \cite{busemeyer2012quantum}). 
 \st{Instead we will simply claim that} while a quantum implementation might provide computational efficiency, it is a model's compositional structure that underlies its interpretability in our sense.}

Indeed, many of our arguments for the interpretability benefits of rich compositional structure apply equally to quantum models as to classical \rl{ones}, as can any arguments which only assume semantics based on symmetric monoidal categories.\footnote{As opposed to cd-categories, which assume classical `copying' structure.} For example, the notion of abstract interpretation, and explanation methods based on \rl{influence constraints}, diagram surgery, and rewrite explanations may all be carried over to quantum models, so that many examples from Section \ref{sec:expl-from-diags} could be taken with quantum semantics. 
Examples \rl{in the literature} of quantum models with abstract interpretations \rl{include the} QNLP models \rl{in \cite{meichanetzidis2023grammar, lorenz2023qnlp},} as well as the quantum conceptual space models in \cite{QonceptsFull2024} (see Example~\ref{ex:consp-conrel}). 

Despite this, one aspect of interpretability which may pose more challenges for quantum models is the notion of a \rl{\emph{concrete}} interpretation. Recall that a variable has such an interpretation when we can assign specific meanings to each of its states. 
\rl{For a variable $V$ with \emph{classical} semantics in $\mathbb{R}$, abstractly interpreted as some feature or concept, 
if $V$ is said to also have a concrete interpretation then it is usually obvious and uncontroversial what it means to interpret each state $r \in \mathbb{R}$ -- this could be the `degree' to which this feature is present, the spatial position of an object, a time stamp or a person's age in years etc.}  
\rl{In contrast, for a variable $V$ with \emph{quantum} semantics given by a $d$-dimensional complex Hilbert space $\hilbH$, its set of states are all corresponding density operators (i.e. trace-1 positive operators on $\hilbH$, see Sec.~\ref{subsec:quantum-processes}). 
Hence, even if $V$ is abstractly interpreted as some feature or concept, it is not clear what a corresponding concrete interpretation of $V$'s states is in general.\footnote{\st{Note that, at least in the first instance, this has nothing to do with the more than a century old debate in physics about the appropriate `interpretation of quantum theory'. In quantum theory each density operator has a completely uncontroversial role within the formalism for computing probabilities of physical events, but the issue here is that not even this kind of immediate reading of the mathematical state is generally available when using quantum semantics for classical variables in AI.}}
For the most elementary quantum system given by a single qubit, i.e. where $d=2$, each state can be parametrised by three real numbers, which can then be visualised as the two angles and the radius of a point in the so called `Bloch ball'. 
Now,} while some variables may have a natural Bloch ball interpretation, such as the example of an RGB \emph{colour} sphere as in \cite{QonceptsFull2024,yan2021qhsl}, in general providing an interpretation for each point of this \rl{ball} is not obvious. 
Note, however, that such interpretations are provided for a number of states of a qubit in the quantum DisCoCirc model in \cite{QDisCoCirc}. 


\rl{The challenge of concrete interpretations in the quantum case has a further aspect, namely} when considering composites of variables. 
For classical deterministic models, we can interpret a state $(v,w)$ of a composite $V \times W$ as simply a pair of interpreted aspects corresponding to $v$ and $w$.\footnote{Probabilistic classical models may in fact possess correlated states (distributions), which cannot be written as product states. However any such state can be written as a probability distribution over product states, and interpreted as an `uncertain' product state. } However, the combination of quantum systems $\hilbH, \hilbK$ is described by their tensor product $\hilbH \otimes \hilbK$. A typical state of this composite is \emph{entangled}, meaning it cannot be written as a product of states of each factor independently.\footnote{\rl{Importantly, it also cannot be written as a probability distribution over product states -- unlike in the case of classical correlation.}} Even when provided with concrete interpretations for (the individual states of) all variables of a model, how are we to interpret such entangled states of their composite?

Despite this problem with concrete interpretations, it may simply be that the other aspects of interpretability listed above, nonetheless suffice to deem a quantum model interpretable. 
In particular the special case of diagram surgery given by the application of `local surgery' from Section \ref{subsec:surgery}, may be especially helpful for quantum models. Given a quantum variable $\sem{\syn{V}} = \hilbH$ we can imagine training a model to come with a family of channels $m_i \colon \hilbH \to \hilbH \otimes C_i$, where $C_i$ is a finite set of concretely interpreted classical outcomes. A channel of this form is called a \emph{quantum instrument}. These would allow us to `probe' the quantum system, whose states may not be concretely interpretable, by applying such an instrument and observing the classical outcomes $C_i$. For example, we may have a variable representing images and probes relating this system to interpretable aspects of the image such as \domain{colour}, \domain{brightness}, or \domain{shape}, encoded as classical variables. 
This notion of `probing' a wire in a compositional model as a form of mediated interpretation may be useful in endowing quantum models with interpretations, by relating them to classical attributes, and would be interesting to explore in future work.

\section{\rl{Conclusions and future directions}} \label{sec:future}

In this article we have presented a compositional viewpoint on AI models and their interpretability, \rl{employing the formal notion of a compositional model. Essentially any AI model, including deterministic, probabilistic or quantum models, can be seen as a compositional model, providing both the grounds for meaningful comparison of different models, and the basis for our notion of an interpretation -- namely, the components of its compositional structure.} 

\rl{While any model can be analysed from this viewpoint, it also naturally leads to the stronger notion of \emph{\CI}\ models -- ones with a complete, abstract interpretation in our sense. Standard neural networks, transformers and other black-box models do not (as such) yield CI models, but classic instances of intrinsically interpretable models, as well as causal models, amongst other kinds, do. The interpretability of CI models may be further distinguished and made stronger in a number of ways, all of which are rooted in the compositional view: a model's interpretation may be concrete; the status of its compositional structure may license certain rules for computing with the components (the \rlb{compositional framework}); the structure may or may not directly correspond to structure in the phenomenon; it may allow for explanations in terms of structural influence constraints; or may facilitate rewrite explanations in terms of diagrammatic equations.} 

Overall, we suggest that taking a compositional view may help to broaden both the search for interpretable AI models, and the kinds of explanations they can provide. There are many directions in which one may continue this research in the pursuit of interpretable compositional models.

\paragraph{\st{Finding more CI models.}}
Broadly, we can say that the space of \rl{CI models} remains to be explored in full detail. Our paradigm examples 
\rl{include intrinsically interpretable models (see Secs.~\ref{subsec:linear-models}, \ref{subsec:rule-model} and \ref{sec:comp-and-interp}), but also} causal models (Sec.~\ref{subsec:causal-models}) and DisCo models in NLP (Sec.~\ref{subsec:DisCoCirc}), where compositional structure comes respectively from causal and grammatical structure. 
However, further kinds of compositional models should be developed in the future. 
\rl{Potential examples may include}
cognitively-inspired models such as conceptual spaces (Section \ref{subsec:conceptual-space-models}), which in particular should be enriched beyond possessing merely states (instances) and effects (concepts) but also `conceptual processes' between domains. 

In Section \ref{sec:quantum} we discussed aspects of interpretability related to quantum models. \st{For now, perhaps the most pressing area in quantum AI is simply the development of effective models, which may become more widespread as quantum hardware improves.} 
\st{Ideally, from our perspective, one would develop quantum models whose compositional structure is directly related to human intelligible terms so as to yield a CI model. Research towards such models within QNLP includes for example \cite{QDisCoCirc}.}

\paragraph{\rl{Learning and relating compositional structures.}}


A major question in the pursuit of \CI\ models is the following: where does the  compositional structure come from? There are cases in which one may indeed specify the signature and hence structure category $\catS$ of a model `by hand'; for example in a data science scenario with \rl{given interpreted variables, in which one explores either intrinsically interpretable models for a particular task or specific causal relations between them.}
In \rl{some} other cases it is known how to apply structure to the data, such as parsers which overlay raw text with grammatical structure, which may then be converted to string diagrams.

However, for \rl{many} applications in AI one hopes to instead train a model from unstructured data. If the resulting model is to possess interpretable variables and processes on them, how should these arise? In answering this question there may be a spectrum of approaches in how much supervision we give the model. It is perhaps naive to expect a model to learn human interpretable \rl{variables with no input of human concepts or knowledge of some form or other in the training,} but one aims to reduce this as much as possible. 
We note that a special case of this problem is in the learning of \rl{causal variables and their causal structure} from (to some extent) unstructured data, which in causal ML is known as \emph{causal representation learning} \cite{SchoelkopfEtAl_2021_TowardCausalRepresentationLearning}. Solving this, and now the broader problem of \emph{compositional representation learning}, remains a holy grail for structured AI, and a major research goal for the future. 

While we have largely focused on defining and giving examples of compositional models \rl{and their interpretations}, there is much more mathematical machinery that could be developed to allow us to \emph{relate compositional models to one another}. This would be interesting to explore in future work, building on standard notions of model relations from categorical logic. 
\rl{An example of the kind of interesting relation one may study is one that relates models at different levels of abstraction or fine-grainedness. 
App.~\ref{subsec:refining-models} indicates} some first steps in this direction, by introducing the notions of morphisms of compositional models and the related \emph{refinement} of diagrams. The latter formalises the intuitive \rl{idea} of refining a model or \rl{process}, viewed as a diagram, to another typically more detailed one with further variables and processes. Special cases include specifying a (causal) model of some distribution or process, refining a causal model to an FCM, or specifying a precise computational implementation of a given model.  


\st{Related to both the refinement of models and the learning of interpretable structure is the concept of \emph{causal abstraction}, discussed in App.~\ref{subsec:causal-abstraction}. This concerns an abstraction relation between causal models, which does not just require equality of diagrams and models, but also constrains how interventions at the different levels relate to each other.}  
In future work it would be interesting to further develop the categorical theory behind causal abstraction, and relate it to our theory of refinements of models. More broadly, is there an interesting notion of `abstraction' for general compositional models, based on some form of diagram surgery?

\paragraph{\rl{The potential of diagrammatic equations.}}

In Section \ref{subsec:rewrite-expl} we introduced the new notion of \emph{rewrite explanations} as an explainability technique applicable to \rl{CI} models. More work is needed to demonstrate that such explanations can be used in practice. In particular, how can we train models coming with the kinds of equations used in rewrite explanations? Should these equations be imposed to hold strictly in the setup of the model, encouraged to hold approximately via minimising a loss function, or simply found experimentally after training? 
As well as this, more work is needed to demonstrate the use of rewrite explanations on trained models, and develop the technique to a fully-fledged explainability tool if this is viable, complete with software for generating and examining explanations.

\rl{A related idea in this spirit of \emph{transparency from equations} is to develop transparent models by specifying their behaviours through equations that their components are trained to (approximately) satisfy.} 
\st{Rather than considering equations that merely feature in the explanation of particular outcomes, here one considers them as making the entire compositional model more functionally transparent.} 
An example we have already met are the equations \eqref{eq:strict-encoder-binv} and \eqref{eq:strong-encoder} for encoder-decoder models, which the loss functions of VAEs enforce approximately in training. Another example are the \emph{get-put} rules developed for \emph{lenses} in database theory, which are now widely studied categorically \cite{bohannon2006relational,foster2007combinators,johnson2010algebras}. These consist of a `get' map, which given a variable $\syn{X}$ returns values of some type $\syn{C}$, and a `put' map which allows to `insert' a new state for this value. These satisfy various equations which define their mutual behaviour and interactions, including the following: 
\[
\input{getput1.tikz} \qquad \qquad \qquad \input{getput2.tikz}
\]
The article \cite{PatternLanguage} aims to give a compositional approach to specifying ML models using equations which constrain loss functions, just as in these examples.  Equational constraints such as these could help to align a model to have desirable properties, for example by removing harmful biases \rl{and allowing one to reason about the way the model will behave, as well as possibly facilitating rewrite explanations of specific outcomes in our sense.} 




\paragraph{\rl{Promising links to related fields.}}

\rlc{As discussed in Section \ref{subsec:compositionality-context}, perhaps the currently most prominent area of AI research \rl{that is} centered on `compositionality' is in the exploration of \emph{compositional behaviour} of ML systems.} 
These include the ability of a trained model to compositionally generalise; for example to generalise from few training examples to many new unseen ones, by combining or composing known examples. It is natural to expect compositionally (i.e. categorically) structured models to come with built-in compositional behavioural abilities, at least in ways which relate to their explicit structure. For example, DisCoCirc models may be able to apply strict logical reasoning to large texts even when trained only on small fragments, as demonstrated in the quantum model of \cite{QDisCoCirc}. However, spelling out precisely in what ways explicit compositional structure leads to compositional behaviour remains a major direction for future work. 

Our perspective should also be tied in future to a number of other structural perspectives on AI, including the area of \emph{geometric deep learning} \cite{bronstein2021geometric}, which could provide further examples of structured compositional models. In this approach the notion of `structure' is largely present in symmetry groups on the training data, which are respected by the model (through `equivariance'); as in the classic example of translational equivariance of a Convolutional Neural Network (CNN). In fact, recently a categorical perspective and generalisation of geometric deep learning has been offered \cite{gavranovic2024categorical} which would be interesting to compare with our work.

A complementary area of applied category theory may be called \st{\emph{categorical learning}}, consisting of categorical descriptions of ML systems and their training setup, including the learning process, parameter updating and optimization. A number of recent works have provided categorical accounts of machine learning in terms of bidirectional transformations \cite{fong2019backprop,shiebler2021category,cruttwell2022categorical}. In future work, it would be interesting to explicitly connect our viewpoint with these approaches which take the training process into account. 

\section{Acknowledgements}
We would like to thank \rlc{Giovanni De Felice, Gabriel Matos, Vincent Wang-Maścianica, Caterina Puca, Konstantinos Meichanetzidis and the rest of the Oxford team of Quantinuum, as well as Lachlan McPheat, for feedback and many helpful and in-depth discussions.} 

\addcontentsline{toc}{section}{References}
\bibliographystyle{alpha}
\bibliography{CI}

\appendix

\section{Further compositional model examples }

\subsection{Linear models} \label{appendix:linear-models}

\irs{Here we show to how to describe linear structure in categories such as $\NN$ diagrammatically, in a manner closely related to the approach of `graphical linear algebra' \cite{bonchi2017interacting}. }

Let $\catC$ be a cd-category with a distinguished object $R$. Let us call an object $V$ \emph{$R$-linear} when it comes with distinguished morphisms as in \eqref{eq:linear-generators}, but now with $\mathbb{R}$ replaced by $R$, called \emph{addition} and \emph{scalar multiplication}, as well as a distinguished state $0$ of $V$, which together satisfy the following conditions. 

Firstly, $R$ is itself $R$-linear, coming with its own addition and scalar multiplication maps, as well as distinguished state denoted by $1$. Secondly, each of the morphisms \eqref{eq:linear-generators} is a deterministic channel. Next, addition is associative and commutative, with unit zero:
\[
\input{plus-monoid.tikz}
\qquad \qquad 
\input{plus-commutative.tikz}
\qquad \qquad 
\input{plus-unit.tikz}
\]
Multiplication on $R$ itself is commutative:
\[
\input{mult-comm.tikz}
\]
For any state $r$ of $R$ we define a \emph{$r$-multiplication} morphism on $V$ as in \eqref{eq:rmult}. We require that scalar multiplication satisfies the following. 
\[
\input{mult2.tikz}
\qquad 
\qquad 
\input{multzero2.tikz}
\qquad 
\qquad 
\input{mult12.tikz}
\]
Finally, we require that addition and multiplication are related by the following, which intuitively encode the relations  that $rv + sv = (r+s)v$ and $r(v + w) = rv + rw$ respectively, for for all $v,w \in V$ and $r, s \in R$.
\[
\input{mult3.tikz}
\qquad \qquad \qquad 
\input{mult4.tikz}
\]
The key example is the category $\NN$ where $R=\mathbb{R}$ and on each object $V=\mathbb{R}^n$ we have that addition is given by $(v,w) \mapsto v + w$ and scalar multiplication by $(r, v) \mapsto r \cdot v$, as expected. Then for each $r \in \mathbb{R}$, \eqref{eq:rmult} is given by $v \mapsto r \cdot v$. 

From these maps one may go on to define further linear operations in string diagrammatic terms. For example, consider the inner product and matrix multiplication morphisms that feature in the transformer architecture.
\[
\input{Matmul.tikz} \ \ :: \ \ (A,v) \mapsto (\sum_j A_{i,j}v_j)^t_{i=1}
\qquad \qquad \qquad 
\input{inner-prod-general.tikz} \ \ :: \ \ (q,j) \mapsto (\langle q_i, k_j \rangle)^{t,s}_{i,j=1}
\]
These decompose as elementary string diagrams as follows. 
\[
\input{Matmul-general2.tikz} 
\]
\[ 
\input{inner-product-general.tikz} 
\] 
where each $\cdot$ denotes the map $(v,w) \mapsto \langle v, w \rangle$.\footnote{One may consider adding the inner product as another elementary generator on each linear object in the language of computational models. Note, however, that on any fixed choice of $Q=K=\mathbb{R}^d$, it can be written as a string diagram in terms of only $+$ and multiply.}.

\subsection{Decision trees} \label{appendix:dec-tree}

An alternative string diagram describing the tree \eqref{eq:tree-as-sd-1}, which shows explicitly which variable each question depends on, is the following. For any question represented by $Q \colon V \to \mathbb{B}$ let us define an alternative form of `controlled $Q$' box: 
\[
\input{C-box-yes-no-4.tikz}
\]
Now the $Y, N$ wires are simply the Booleans $\mathbb{B}$. Again the $Y$ (`yes') wire carries true iff the `control' $\mathbb{B}$ is true and $Q(v)$ is true, and $N$ carries true iff the control is true and $Q(v)$ is false. We may then represent the same decision tree as a function of $T$ (temperature) and $D$ (days since 2011) by the following computational model, which explicitly shows the variables used in each question:
\[
\input{tree-converted-2.tikz}
\]
Here the first question is passed `true' to its control wire, omitted from the diagram.

\subsection{Stairs readers} \label{appendix:stairs-reader}

A stairs reader may be equivalently described as an RNN where the transition function takes the form: 
\[
\input{stairs-def.tikz} \qquad \qquad \text{ where } \qquad \qquad \input{stairs-embed.tikz}
\]
for the, usually linear, stairs function $g$. For any RNN, the function $e$ above sends each word $\syn{w}$ to its state representation and may be called the \emph{embedding} map. From this the state representation of a sequence $(w_1,\dots,w_n)$ shown in section \ref{subsec:NLP-sequence-model} follows. 
Hence a representation of a phrase is given by repeatedly applying the stairs map to the vector embeddings of each word.

\subsection{CCG models} \label{app:CCG-model}

In this section we consider compositional models based on the linguistic formalism \emph{Combinatory Categorial Grammar} (CCG). This approach is due to \cite{yeung2021ccg}, to which we refer readers for an introduction to CCG and more details.

Let $\lexicon$ be a lexicon consisting of words $\syn{w} : \syn{t}$ where each is assigned a type $\syn{t}$ within a CCG built from a set of basic types such as $\n$ (noun), $\s$ (sentence), etc. For example we may have \natlang{Alice}$ : \n$, \natlang{sleeps} $: \n \ccgto \s$. Given a sequence of typed words, one may apply the \emph{rules} of CCG, such as \emph{forward application} (FA) or \emph{backward application} (BA), to specify its grammatical structure and derive whether it forms a valid sentence. 

For example, we can depict a derivation of how `Alice likes Bob' forms a sentence as below. 
\begin{equation} \label{eq:CCG-ex}
\input{CCGdiag2states.tikz}
\end{equation}

Now let us see how to define a suitable structure category to give formal meaning to such a diagram. This is based on the observation that the rules of CCG are closely related to the structure of the following form of category \cite{yeung2021ccg}. A monoidal category $\catB$ is said to be \emph{biclosed} when it is both left and right-closed.\footnote{Formally, there are natural isomorphisms $\catB(A, C \closedfrom B) \simeq \catB(A \otimes B, C) \simeq \catB(B, A \closedto C)$.} This means for each pair of objects $A, B$ there are objects $A \closedfrom B$ and $A \closedto B$, such that morphisms with inputs $A, B$ and output $C$ are in bijection with those of either of the left or right-hand types below. 
\[ 
\input{biclosed.tikz} 
\]
As shown in \cite{yeung2021ccg}, all the basic rules of CCG exist in any biclosed category where, given types corresponding to objects $t_1, t_2$, the CCG type $t_1 \ccgto t_2$ corresponds to the object suggested by the notation, and similarly for $\ccgfrom$. 
In particular, any CCG lexicon $\lexicon$ generates a free bi-closed category $\FreeBi(\lexicon)$. The variables are the basic types $\syn{n},\syn{s}$, from which all further types are constructed using $\closedfrom, \closedto$, and there is a generator state $\syn{w}$ of $t$ for each word $\syn{w}$ of type $t$ in $\lexicon$. For example, there are generators given by the state $\natlang{Alice}$ of $n$ and $\natlang{likes}$ of $(n \closedto s) \closedfrom n$ as in \eqref{eq:CCG-ex}. Then using the bi-closed structure, any diagram for a CCG derivation, such as \eqref{eq:CCG-ex}, becomes a morphism in this category. 

To now give semantics to CCG derived sentences, we must give a model of this structure, as follows. 

\begin{definition}
Let $\lexicon$ be a CCG lexicon. By a \emph{CCG model} of $\lexicon$ we mean a compositional model where $\catS = \FreeBi(\lexicon)$, $\catC$ is biclosed, and $\sem{-}$ is a biclosed monoidal functor.  
\end{definition}

The fact that $\sem{-}$ is bi-closed means $\sem{t_1 \closedto t_2} = \sem{t_1} \to \sem{t_2}$ and $\sem{t_1 \closedfrom t_2} = \sem{t_1} \closedfrom \sem{t_2}$ for all types. 

Explicitly then, to give a CCG model amounts to specifying a biclosed category $\catC$, an object for each basic type, e.g. $n = \sem{\syn{n}}$, and a state $\sem{\syn{w}}$ of the appropriate object in $\catC$ for each $\syn{w} : t$ in the lexicon, e.g. a state $\natlang{Alice}$ of $n$ and a state $\natlang{likes}$ of $(n \closedto s) \closedfrom n$, in $\catC$. Then all other CCG types and rules come with an automatic meaning in $\catC$ from the bi-closed structure, and so a diagram such as \eqref{eq:CCG-ex} in $\catS$ is mapped to the `same' form of diagram in $\catC$. 

\begin{example}
One commonly used semantics is the category $\catC=\FVecR$ of finite-dimensional real vector spaces and linear maps, which we saw in Example \ref{ex:dcat-vec} was compact. In fact, any compact category is bi-closed with $A \closedto B = A^* \otimes B$ and $A \closedfrom B = A \otimes B^*$. Alternatively, the category $\Setcat$ of sets and functions is an example of a bi-closed category that provides a `functional' and `truth-theoretic' form of semantics for CCG models.
\end{example}

DisCoCat and CCG models are closely related, since rigid monoidal categories are a special case of biclosed ones, which we outline in Appendix \ref{appendix:disco-to-ccg}. 


\paragraph{CCG phrase structure models}
We can weaken the requirement that our categories are bi-closed to more readily obtain models from a CCG lexicon $\lexicon$, at the expense of having models which no longer obey the combinatorial rules of CCG. Models of the following kind have been studied in the CCG case by Hermann et al. \cite{hermann-blunsom-2013-role} and in the more general phrase-structure case by Socher et al.~\cite{socher}.

By a \emph{CCG phrase structure model} of $\Sigma$ we mean a compositional model with a variable $\syn{t}$ for each CCG type, a generator state $\syn{w}$ of $\syn{t}$ for each $\syn{w} \colon \syn{t}$ in $\lexicon$, and a generator for each possible CCG rule and set of types to which it may be applied. Thus, by construction, we have all the generators needed to give meaning to a diagram such as \eqref{eq:CCG-ex}, without requiring the category to be bi-closed. 

Any CCG model induces a phrase structure model via its bi-closed structure, where now $\sem{t_1 \ccgto t_2} = \sem{t_1} \closedto \sem{t_2}$ etc. For a general CCG phrase structure model, however, there is no longer any enforced relation between a type such as $\n \ccgto \s$ and $\n, \s$, with the former being `just a name'. 

\begin{example}
A CCG phrase structure model in $\NN$ amounts to specifying a vector space $\mathbb{R}^{d[t]}$ (i.e. a dimension) for each CCG type $t$, a vector $w \in \mathbb{R}^{d[t]}$ for each $\syn{w} : t$ in $\lexicon$, and a neural network for each typed CCG rule. A sentence such as `Alice likes Bob' is first mapped to a CCG derivation by a parser, and then to the resulting state given by the composite \eqref{eq:CCG-ex}. 
\end{example}

Since specifying a morphism (e.g. neural network) for every typed CCG rule may be cumbersome, it is natural to weaken this structure further. Firstly, one can ignore the types on words, so that there is a single variable and every input or output to a generator is of this type, with a single generator for each rule type (`FA', `BA' etc.). Then diagrams appear as in the left-hand below. Secondly, one may replace all rule generators with a single generator denoted `UNIBOX', so that diagrams appear as in the right-hand below.\footnote{By definition, each of these structures is weaker so that a model of an earlier structure specialises to a later one. Formally these specify structure categories $\catS_{\mathrm{Notypes}}$ and $\catS_{\mathrm{UNI}}$ related to the generic CCG structure category $\catS_{\mathrm{CCG}}$ by quotient functors $\catS_\mathrm{CCG} \to \catS_{\mathrm{Notypes}} \to \catS_{\mathrm{UNI}}$ so that any model of $\catS_{\mathrm{UNI}}$ forms a model of $\catS_{\mathrm{Notypes}}$ and any such model forms one of $\catS$.}

\[
\input{CCG-typesgone.tikz}
\qquad \qquad  \qquad \qquad \qquad 
\input{CCGunibox.tikz}
\]
Thus a compositional model of the left-hand kind in $\NN$ consists of a single vector space with vectors for each word and neural networks for each rule type, while the latter kind have the single network `UNIBOX'. Models of each kind are implemented in the \texttt{lambeq} software \cite{kartsaklis2021lambeq}.

\paragraph{Interpretation}
Similarly to DisCoCat models, in a CCG model the variable $\syn{t}$ for each CCG type has an abstract interpretation as that type, and the generator $\syn{w}$ for each word has an abstract interpretation as (the meaning of) that word. Each generator $\syn{FA}, \syn{BA}, \dots$ for a CCG rule has an abstract interpretation as encoding this form of grammatical composition. Combining these, we can informally say that we interpret a sentence diagram such as \eqref{eq:CCG-ex} as encoding the meaning of the sentence, complete with grammatical structure. 

A CCG phrase structure model similarly has an abstract interpretation for each word and composition rule, though now all variables $\syn{t}$ receive the same interpretation (as simply `a word'), since all word types coincide. Since multiple CCG rules are mapped to the same semantics, their interpretations also coincide, in the case of a `UNIBOX' model reducing to a single interpretation (as `word composition').


\subsection{Relation between DisCoCat and CCG models} \label{appendix:disco-to-ccg}

Any rigid monoidal category is a special case of a biclosed category, allowing  DisCoCat models to also be used as CCG models. Here we sketch this briefly; for more details see \cite{yeung2021ccg,de2022categorical}. Given any CCG lexicon $\lexicon$ we can specify a corresponding pre-group grammar $\lexicon'$ by mapping each type $t_1 \ccgto t_2$ to $t_1^l \cdot t_2$ and $t_1 \ccgfrom t_2$ to $t_1 \cdot t_2^r$. Similarly, any rigid category forms a biclosed category by setting $A \closedto B = A^l \otimes B$ and $A \closedfrom B = A \otimes B^r$. 
For example, the morphism in a biclosed category corresponding in CCG to the forward application rule in CCG is equal to the following. 
\[ 
\input{FAmap.tikz}
\]
Formally, this specifies a biclosed monoidal functor $\FreeBi(\lexicon) \to \FreeRigid(\lexicon')$, so that any DisCoCat model of $\lexicon'$ lifts to a CCG model of $\lexicon$. 
As noted in \cite{yeung2021ccg}, this mapping can be useful in allowing us to readily inspect the equality of apparently distinct CCG parsings of the same sentence, such as the following (where FC denotes the function composition rule): 
\begin{equation} \label{eq:bigbadwolf}
\input{CCG-to-wolf.tikz}
\end{equation}
Relations such as the above can be seen as a major advantage of both CCG models and DisCoCat models, which make use of categorical composition. In contrast, generic CCG phrase structure models will typically not be faithful to such structural features of grammar; for example, the two representations in \eqref{eq:bigbadwolf} will not coincide.

\subsection{Higher order morphisms in DisCoCirc} \label{app:dcirc-hocat}

Here we make the notion of higher order morphisms used in DisCoCirc models more precise. Higher order maps have been studied in various forms \cite{wilson2022mathematical,roman2020open}, and in particular they are often allowed to be applied not only to morphisms of some fixed type $f \colon A \to B$ but also to morphisms with any extra wires such as $g \colon A \otimes C \to B \otimes D$, leading to further naturality conditions that they should satisfy.  However, this ability is not required for DisCoCirc, and so we use only the following simpler notion. 

In any category $\catC$, by a \emph{higher order morphism} we mean a function
\begin{equation} \label{eq:h.o.-type}
h \colon \prod^n_{i=1} \catC(A_i,B_i)  \to \catC(C,D)
\end{equation}
for some objects $A_i, B_i$, for $i=1,\dots,n$, and $C, D$. Thus, given morphisms $f_i \colon A_i \to B_i$ in $\catC$, for $i=1\dots,n$, it returns a morphism $h(f_1,\dots,f_n) \colon C \to D$. As defined earlier, a h.o.-category is a monoidal category $\catC$ with a distinguished set $H_\catC$ of higher order morphisms. 

By a \emph{h.o.-functor} we mean a monoidal functor $F \colon \catC \to \catD$ between h.o.-categories along with a function $H_F \colon H_\catC \to H_\catD$, such that each higher order morphism $h$ of type \eqref{eq:h.o.-type} in $H_\catC$, is mapped to one of type
\[
H_F(h) \colon \prod^n_{i=1} \catD(F(A_i),F(B_i)) \to \catD(F(C),F(D))
\]
satisfying $H_F(h)(F(f_1),\dots,F(f_n)) = F(h(f_1,\dots,f_n))$ for all inputs $f_1,\dots, f_n$. Then a DisCoCirc model may be defined as a h.o.-functor as in Definition \ref{Def:dcirc-model}.

\begin{example}
$\TextCircuits_\lexicon$ forms a h.o.-category as expected, with the generators of shape \eqref{eq:dcirc-all} as its higher order morphisms. Here each generator takes the appropriate shapes of text circuits as inputs and returns as output the text circuit in which they have been inserted into the box.
\end{example}

\begin{example}
Any closed SMC forms a h.o.-category where higher order morphisms \eqref{eq:h.o.-type} correspond to morphisms $h' \colon (A_1 \closedto B_1) \otimes \dots \otimes (A_n \closedto B_n) \to (C \closedto D)$, by setting $h(f_1,\dots,f_n)$ to be the morphism induced by the state $h' \circ (f'_1 \otimes \dots \otimes f'_n)$ of $C \closedto D$, where $f'_j$ is the state of $A_j \closedto B_j$ induced by $f_j \colon A_j \to B_j$. In this way any Cartesian closed category (such as $\Setcat$) or any compact category is a h.o.-category. 
\end{example}

\subsection{Conceptual encoders and decoders} \label{subsec:concept-encoders} 

In this section we will explore how concepts, similar to those of the conceptual space models of Section \ref{subsec:conceptual-space-models}, can be implemented practically in encoder-decoder type setups as in Section \ref{subsec:encoder-decoder-model}. This is achieved by equipping a latent space $Z$ with a collection of concepts, represented on $Z$ in one of two ways: either as states, or as effects. Practical implementations are given by the Conceptual VAE and quantum concepts models introduced in \cite{QonceptsFull2024} (and in each of these models $Z$ additionally takes on  the structure of a conceptual space model, factorising in terms of domains). 

\paragraph{Concepts as generative.}

Let us introduce the first such approach.  We can enrich an encoder-decoder model by equipping our latent space $Z$ with a collection of concepts $(C_i)^n_{i=1}$, each given by a state of $Z$ and with a name from some set of labels $L=\{1,\dots,n\}$. By a \emph{conceptual encoder-decoder model} we mean a compositional model in a cd-category with variables $\syn{X}, \syn{Z}$ as well as an additional  \emph{concept labels} variable $\syn{\SpaceofCon}$, and generators shown as below, such that all are represented as channels, and that is trained to approximately satisfying the following equation, similarly to an encoder-decoder model. 
\begin{equation} \label{eq:con-vae-main}
\input{conc-vae-master-2.tikz}
\end{equation}
Thus we have generators $d, e$ for the decoder and encoder as before. The data is now labelled, given as a distribution over both inputs $X$ and concept labels $\SpaceofCon$, which typically is represented by a finite set. Rather than a single prior over $Z$, the channel $\conrep$ maps each concept label $i \in L$ to a normalised state (i.e. distribution, for a model in $\FStoch$) of $Z$, which we call the \emph{generative concept} with label $i$:
\[
\input{concept-i.tikz}
\]
Passing this distribution to the decoder $d \colon Z \to X$ then yields a distribution over inputs $X$, from which we may sample to generate new instances of the concept. 
Note that marginalizing $Z$ in \eqref{eq:con-vae-main} tells us that we can recover the original data by choosing a concept label $i$, according to their distribution in the data, and then generating instances using $d \circ C_i$, in this way. 

In the case where $\SpaceofCon$ is discrete and each concept does appear in the data, such a model is equivalent to providing an encoder-decoder model `for each concept'. That is, for each label $i$, 
$(X, Z, d, e, C_i, \mathsf{Data}|_i)$ forms an encoder-decoder model as in \eqref{eq:strict-encoder-binv}. Here $\mathsf{Data}|_i$ is given by conditioning the data to those with concept label $i$, and the prior over $Z$ is now given by $C_i$.

\paragraph{Concepts as classifiers.}
Alternatively, as in a conceptual space model (Section \ref{subsec:conceptual-space-models}) we can treat concepts as effects, and so use them as \emph{classifiers}, i.e.~maps which send each latent point $z$ to a value $C(z)$, indicating how well it fits the concept. From such values one may determine which concept best fits an instance. From this perspective we only require an encoder from some input space $X$ into a latent space $Z$.




By a \emph{classification concept model} we mean a model with variables $\syn{X}, \syn{Z}$, $\syn{L}$ and generators $\syn{Data}$ and $\syn{e}$ of the same form as a conceptual encoder-decoder model (but not necessarily with a decoder), and where the representation map is now given by an effect on $\syn{Z}, \syn{\SpaceofCon}$  as below, so that plugging in a concept label $i \in \SpaceofCon$ yields a concept $C_i$ as an effect on $Z$: 
\[ 
\input{Ci-from-valid.tikz} 
\]

We can pass any data point $x \in X$ to the encoder to produce a representation $e \circ x$ over $Z$, and then apply each concept to obtain a scalar $C_i \circ e \circ x$, representing how well each concept fits, similarly to \eqref{eq:instance}. We typically assume such a model comes with a generator giving us a prior distribution over concept labels, intuitively telling us how likely each concept is. When $X$ and $L$ are classical spaces, we can then carry out conditioning to produce a \emph{classification} map $X \to \SpaceofCon$ which sends each input $x$ in $X$ to a distribution over possible concept (labels):
\begin{equation} \label{eq:classifier-map-concepts}
\input{classif-inputs.tikz}
\end{equation}
For an input $x \in X$, this outputs the probability distribution $P(C_i | x)$ over concepts $C_i$. Here the dashed box denotes probabilistic conditioning, treated in more detail in Section \ref{subsec:stats-models}.

\paragraph{Connecting both views of concepts.}
In a classical classification concept model, the classificatory and generative views of concepts can be related by the following, where $\mu$ is a given measure over $Z$ (e.g. the Lebesgue measure, when using a suitable category of `continuous' probability channels). 
\begin{equation} \label{eq:concept-effect-state-map}
\input{concept-effect-state-map.tikz} 
\end{equation}
This states that each effect $C_i$ forms a \emph{density} for the state $C_i$ with respect to the measure $\mu$. In a quantum model, one instead associates a concept effect $C$ (i.e. positive operator) with a normalised state (density matrix) $\rho_C = \normop(C) = \frac{1}{\Tr(C)}{C}$.

Let us now meet two examples of conceptual encoder-decoder setups, both from \cite{QonceptsFull2024}.  

\begin{example} [Conceptual VAE] \label{Ex:ConVAE}
In \cite{QonceptsFull2024} a model is presented which is both a conceptual encoder-decoder model and concept classifier model, called the \emph{Conceptual VAE}. The model makes use of both an encoder and decoder, relating a latent space $Z$ to an input space of images $X$. The model is trained on data given by simple 2d shapes generated from the `Spriteworld' software, such as the following:
\begin{figure}[H]
\centering
\begin{subfigure}{3cm}
		\centering
\includegraphics[scale=0.6]{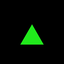}
	\end{subfigure}
\begin{subfigure}{3cm}
		\centering
\includegraphics[scale=0.6]{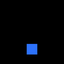}
	\end{subfigure}
\begin{subfigure}{3cm}
		\centering
\includegraphics[scale=0.6]{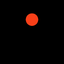}
	\end{subfigure}
\begin{subfigure}{3cm}
		\centering
\includegraphics[scale=0.6]{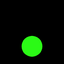}
	\end{subfigure}
\end{figure}
The conceptual VAE uses a classical probabilistic semantics where each channel takes a Gaussian form.\footnote{In particular it takes semantics in the continuous setting of the category $\ConSp$ introduced in Example \ref{ex:consp-conrel}, rather than merely $\FStoch$.} Each concept $C_i$ may be seen either as a state of $Z$, given by a Gaussian distribution, or as an an effect on $Z$, via the corresponding density function, as in \eqref{eq:concept-effect-state-map}. 


The latent space $Z$ factorises as a conceptual space, with domains $Z_1 = \dc{shape}$, $Z_2 = \dc{colour}$, $Z_3 = \dc{position}$ and $Z_4 = \dc{size}$, where an instance corresponds to a single domain. Each domain is taken to be one-dimensional, with $Z_i = \mathbb{R}$. Examples of concepts include single domain concepts such as \concept{red}, \concept{square}, \concept{large}, as well as as `entangled' concepts which depend on multiple domains.

The training is achieved by a variation of the standard ELBO loss for VAEs \eqref{eq:ELBO-loss} to take into account the concept $C_i$ for each training instance as a prior over $Z$. This relates to minimising a KL-divergence for each concept in just the same way that the standard ELBO loss \eqref{eq:ELBO-loss} relates to a KL-divergence between the two sides of \eqref{eq:strict-encoder-binv}.

\end{example}

\begin{example}[Quantum concepts model] 
In \cite{QonceptsFull2024} it is shown how one may use the same dataset as the Conceptual VAE, consisting of 2d images from the Spriteworld software, to train a concept classifier model with a quantum semantics. Again $Z$ forms a conceptual space with domains $Z_1 = \dc{shape}$, $Z_2 = \dc{colour}$, $Z_3 = \dc{position}$ and $Z_4 = \dc{size}$, and the same concept labels. Semantics is now taken in $\catC=\Quant{}$, so that each domain is associated with a Hilbert space $Z_i = \hilbH_i$, and the overall space $Z = \hilbH_1 \otimes \dots \otimes \hilbH_4$ with their tensor product. The encoder sends each input image in $X$ to a pure quantum state (instance) over the Hilbert space $Z$. Each concept is represented as a positive operator, and is tested against the instance by applying an appropriate quantum measurement as in \eqref{eq:instance}. The model is implemented using qubits via (simulated) quantum circuits. 
\end{example}

\paragraph{Interpretation}
In a conceptual encoder-decoder model, or classification concept model, the input space $\syn{X}$ typically has a concrete interpretation, as does the set of concept labels $\syn{L}$, both being applicable to the training data. In general $\syn{Z}$ has only a (rather broad) abstract interpretation as a representation space, $\syn{e}, \syn{d}$ as an encoder and decoder, and $C$ as sending each label in $L$ to its concept. Each specific state or effect $C_i$ has an interpretation as that specific concept.  The classification map \eqref{eq:classifier-map-concepts} has a concrete interpretation as assigning each input to its likelihoods of being interpreted as each concept (label). In the examples above, the space $\syn{Z}$ has the further structure of a conceptual space, factorising in terms of domains $\syn{Z_1},\dots,\syn{Z_n}$ which each come with a specific abstract interpretation. 

As remarked for conceptual space models, endowing an encoder-decoder with a set of concepts $C_i$ (either states or effects) can help us interpret the latent space $\syn{Z}$ further (while not necessarily providing a full concrete interpretation), by allowing us to understand a point $z$ by seeing how similar it is to each concept.

\subsection{Disentangled representations} \label{app:disent-reps}

A common approach to attempting to provide a latent space $Z$ with further structure is to ask that it factorises as a product $Z = Z_1 \otimes \dots \otimes Z_n$ of suitably `independent' and `meaningful' factors $Z_i$, similarly to the domains in a conceptual space. A literature in ML exists exploring the requirements and learnability of such `disentangled representations'. In \cite{higgins2018towards} it is suggested that a disentangled representation can be identified by endowing each factor with a group of symmetries, with the product of these groups acting on $Z$. 

To treat this in our setup, first observe that any group $G$ may be viewed as a category with a single object $\star$ and a morphism $\syn{g} \colon \star \to \star$ for each $\syn{g} \in G$. Composition is group multiplication $g \cdot h$ , and the identity is the unit $1 \in G$. Here every morphism is an isomorphism. 

Now a \emph{disentangled representation} may be defined as a compositional model with variables $\syn{Z_1},\dots,\syn{Z_n},\syn{Z}$ and a generator $\syn{g_i} \colon \syn{Z_i} \to \syn{Z_i}$ for each group element $g_i \in G_i$, subject to all equations which hold in the group\footnote{More efficiently, one may merely specify a sufficient set of generators and relations for each group $G_i$ in the usual group-theoretic sense, and these will generate the same structure category $\catS$.}, and the equation stating that $\syn{Z} = \syn{Z_1} \otimes \dots \otimes \syn{Z_n}$. The model therefore specifies objects $Z_i$ in the semantics category $\catC$, whose composite $Z$ comes with a \emph{$G$-action} for the product group $G=G_1 \times \dots \times G_n$. That is, for each $g=(g_1,\dots,g_n)$ we have an isomorphism: 
\begin{equation} \label{eq:group-element}
\input{group-product.tikz}
\end{equation}
such that the following hold:
\[
\input{group-rep.tikz}
\]
Through the choice of semantics category one can impose further structure; for example, the category $\catC=\cat{Vec}$ of vector spaces and linear maps imposes that each group action is linear, forming a \emph{group representation} \cite{higgins2018towards}.

\paragraph{Interpretation}
Typically a disentangled representation comes with an abstract interpretation for each factor and symmetry group, for example that $Z_1$ is an image space with $G_1$ as translation, while $Z_2$ is a colour space with $G_2$ as rotations on the color wheel. In \cite{higgins2018towards} these symmetries are understood as `in the world', known to us, and the representation manifests them in the model. This interpretation may or may not be concrete; for example we may know that $\syn{Z_2}$ represents colour but not precisely the colour corresponding to each element $z \in Z_2$.

\section{CFEs vs Pearlian counterfactuals}
\label{App:CFE-details}

As mentioned in Sec.~\ref{subsec:CFEs-vs-CFs}, although in the literature a CFE is often said to be a counterfactual in the causal model sense, just varying over the possible inputs of a function, on the face of it, does not look like a special case of the definition of a counterfactual in Eq.~\eqref{eq:counterfactual-simpler} -- a definition relative to a \emph{closed} SCM. 
In addition to this immediate observation, it is worth having a closer look to see whether it perhaps holds true in a less obvious way. 

In the definition of a counterfacutal, the conditioning on the observed facts that obtained in the actual world, effectively updates the probability distribution over the exogenous `noise' variables $U_i$, which in the context of data science problems typically are \emph{unobserved}. 
However, if one knew the exact values of the $U_i$ that led to the observed facts, then one can of course equivalently just consider the hypothetical world `on its own' with these corresponding values for the $U_i$'s fed in. 
For instance, in Example~\ref{ex:aspirin-counterfactual}, if one knew that it was precisely $U_A=u_A$ and $U_H=u_H$ that led to $A=n$, $H=y$ then the counterfactual becomes a `one-world' computation with the intervention in the world where $U_A$ and $U_H$ have been fixed correspondingly. 

One might then wonder whether this is how a CFE is to be seen as a counterfactual.  
The model $\modelM$ in Eq.~\eqref{eq:basic-CFE_x} induces an open deterministic causal model -- a very simple one with one single mechanism $M$ for $Y$ with parents  $X_1,...,X_n$. Crucially, the input variables $X_i$ are generally not statistically independent from one another -- in contrast to the exogenous variables $U_i$ in an SCM -- and there typically is a data distribution $D$ over $X_1,...,X_n$ which has $x$ in its support and which does not factorise. 
\[
	\input{CFE-vs-CF_1.tikz}
\]


To try and make sense of the claim, one might suppose there is also an unknown SCM $\modelM_X$ which generates $D = \sem{\modelM_X}$, and such that if one made it explicit, then the alternative pair $(x',y')$ would always become a counterfactual in the strict sense relative to the combination of $\modelM_X$ and $\modelM_L$. 
That is, one may wonder whether $y'$ would (assuming $\modelM_X$ and $\modelM_L$) necessarily be the answer to the counterfactual question: ``given that $X=x$ and $Y=y$ in the actual world, what would $Y$ have been had $X=x'$?". 
However, this is not the case as shown in Eq.~\eqref{eq:CFE-vs-CF_2_v2}, where we factorised the SCM $\modelM_X$ into the deterministic part $\modelF_X$ and the state $\model{L}_X$ on its exogenous variables (see Sec.~\ref{subsec:framework-of-causal-models}).  
The inequality is essentially due to the following. The question is whether the alternative input $x'$, differing from $x$ only in component $x'_i$, can arise through intervening only on $X_i$ while allowing all other input variables $X_j$ to be untouched and evolve according to their usual deterministic evolution. The answer is that this is not always possible, depending on $\modelM_X$, and in particular on whether $X_i$ has descendants in the causal structure of $\modelM_X$, i.e.~on causal assumptions not part of the model $\modelM_L$. 

\begin{equation}
	\input{CFE-vs-CF_2_v2.tikz} 
	\label{eq:CFE-vs-CF_2_v2}
\end{equation}

As a result, the pair $(x',y')$ does not constitute a well-defined counterfactual relative to just $\modelM_L$ -- neither on the face of it, nor can one fill in the details by envoking an implicit model.

\section{Relating models}  \label{sec:further-theory}

\subsection{Refinements of models}
 \label{subsec:refining-models}

\rl{A first kind of relation between compositional models, which is natural to consider, is that of refining, that is the refining of variables and/or processes between the variables. 
Refinements include, for example, the notions of more fine-grained (causal) explanations and concrete (neural) implementations.} 

\begin{definition}
\irs{
Let $\modelM=(\catS,\sem{-},\catC)$ and $\modelM'=(\catS',\sem{-}',\catC)$ be compositional models in $\catC$. 
A \emph{morphism} of models $R \colon \modelM \to \modelM'$ is a functor\footnote{Here we assume that $R$ is the same kind of functor as the models (e.g. monoidal, cd-functor).} as below such that the following commutes. 
\[
\input{morphism-models-2.tikz}
\]}
We say $R$ is \emph{variable-preserving} if $R(\syn{V})$ is a variable of ${\modelM'}$, for every variable $\syn{V}$ of $\modelM$, and \emph{faithful} if $\syn{V} \mapsto R(\syn{V})$ is injective. By a \emph{refinement} of a diagram $\diagD$ in $\catC$ we mean a diagram $\diagD'$ in $\catC$ along with a morphism $R \colon \modelM_\diagD \to \modelM_{\diagD'}$ of their induced compositional models in $\catC$, such that $R(\diagD) = R(\diagD')$. 
\end{definition}

The notion of refinement in particular formalises the following intuitive notion. 
Consider a morphism $f$ in $\catC$ with specified inputs and outputs, viewed as a diagram: 
\[
\input{fbox.tikz}
\]
By a \emph{compositional model of $f$} we mean a diagram $\diagD$ in $\catC$ which forms a faithful refinement of this diagram. 

Explicitly, this means that $\diagD$ is again a diagram in $\catC$ with $\sem{\diagD} = f$. When variable-preserving it futher means that its inputs $(\syn{X'_i})^n_{i=1}$ have $\sem{\syn{X'_i}} = \sem{\syn{X_i}} = X_i$, and similarly for its ouputs. Hence the diagram $\diagD$ has `the same' inputs and outputs as $f$ but has potentially further structure `internally'.

\begin{example}
Suppose we have an input-output model given by a morphism $f$ from inputs $X$ to outputs $Y$, and then specify that these factorise in terms of variables $X_1,\dots,X_n$ to $Y_1,\dots,Y_m$. Then we obtain a (non variable-preserving) refinement of diagrams expressed by the first equality below. 
\[
\input{refinement1.tikz}
\]
Suppose we outline the further factorisation in terms of an `encoder' and `decoder' over $Z$ as in the second model. This amounts to a refinement of diagrams corresponding to the second equation. Here the refinement is faithful and variable-preserving and has also introduced a new variable $\syn{Z}$. 


\end{example}

\begin{example}
An \emph{open causal model} $\modelM$ of a channel $c$ is \rl{precisely one} whose diagram $\diagD_\modelM$ in $\catC$ forms a faithful variable-preserving model of $c$. 
For example, consider a distribution $\omega$ over variables $S, L, A$ corresponding to the left-hand diagram below in $\Stoch$, \rl{where, following our earlier Ex.~\ref{Ex_CBN}, the variables $S, L$ and $A$ may for instance stand for a person's choice to smoke, whether or not they develop lung cancer, and their age, respectively. 	
A causal model of $\omega$ in the above sense then amounts to a diagram in $\catC$ with the same output variables, such that the following equality holds, that is, it is precisely the usual notion of a `causal model for a distribution', i.e. a model that is a candidate causal explanation of that distribution.} 
\[
\input{causal-model-of.tikz}
\]
\end{example}

\begin{example}
Consider a causal model $\modelM$ in $\catC$ with variables $V = \{X_i\}^n_{i=1}$, and suppose we refine this to an FCM $\modelF$ with $V$ as its endogenous variables in the usual sense. This corresponds to a morphism $\modelM \to \modelF$ where we identify the $\syn{X_i}$ and their semantics $X_i$ in both models and map each mechanism $\syn{c_i}$ to the composite of $\syn{f_i}$ and $\syn{\lambda_i}$ as syntax, such that the semantics also matches as expected:
\[
\input{de-SCM-refine.tikz}
\]
In other words, the left-hand diagram in $\catC$ is refined to that on the right, for each mechanism in $\modelM$.

We can depict the whole refinement through an equality of diagrams where we understand this to mean that variables labelled the same on both sides are identified and \rl{each dashed box on the left corresponds to the respective} dashed box on the right. 
\rl{Again returning to our earlier smoking example from Ex.~\ref{Ex_CBN} such a refinement} may be depicted as: 
\begin{equation}
	\input{FCMrefinement.tikz} 
\end{equation}
\end{example}

\begin{example}
A computational implementation of a diagram $\diagD$ in $\catC$ can often be described as a (typically non-variable preserving) refinement to a diagram $\diagD'$ consisting of elementary computational units (e.g. wires of the form $\mathbb{R}$ and neurons, or qubits and ZX-diagrams).
\[
\input{dcirclevel3.tikz}
\qquad 
=
\qquad
\scalebox{0.9}{\input{dcircneural3_v2.tikz}}
\]
\end{example}




\subsection{Causal abstraction} \label{subsec:causal-abstraction}

\rl{Rather than, as with refinements, going down in the level of abstraction or fine-grainedness, let us now consider a prominent example of a relation that concerns the other direction. 
As we saw in Sec.~\ref{sec:CI-and-causal-XAI} and also Sec.~\ref{sec:comp-and-interp},} an ideal kind of model is one with explicit causal structure between interpreted variables which reflects that of the phenomenon in the world. Yet this poses a challenge: such models are notoriously hard to come by. An interesting approach to arriving at such models is via \emph{causal abstraction}. 
\rl{Importantly, and as we will see shortly, this is not just `looking at refinement from the other direction', but a much more constrained and hence stronger notion.}

Causal abstraction is a relation between two causal models, originally defined independently from ML in the philosophy of causation and causal modeling \cite{ChalupkaEtAl_2016_MultiLevelCauseEffectSystems, ChalupkaEtAl_2017_CausalFeatureLearning, BeckersEtAl_2019_AbstractingCausalModles, beckers2020approximate}. While there are various versions of the concept, roughly a `high-level' causal model $\highlevel$ is a causal abstraction of a `low-level' causal model $\lowlevel$ when there is an (`abstraction') map from the low to the high-level that plays well with the causal structure, so that interventions on $\highlevel$ have the same effect as on $\lowlevel$ once suitably translated across levels. The intuition is that if $\lowlevel$ is an NN model trained on some task, seen as a deterministic open causal model, and $\highlevel$ captures our causal knowledge of the phenomenon that underlies the task, then the causal abstraction relation gives a strong kind of interpretability for $\lowlevel$.\footnote{An obvious question may be: why this is interesting if we are already \emph{given} a high-level causal model -- why not just work with that in the first place? 
The point is that $\lowlevel$ takes as input data not directly related to the variables of $\highlevel$ -- the nature of the task means that `our' causal model $\highlevel$ cannot be directly applied to it, while one would also like to leverage the power of NN-based ML.} 
 
There is a series of works exploring this idea. 
In \cite{GeigerEtAl_2021_CausalAbstracttoinOfNN, geiger2024finding} Geiger \etal\ study, given models $\lowlevel$ and $\highlevel$, how to find and check \emph{alignment maps} with respect to which the given models may stand in a causal abstraction relation, while in  \cite{GeigerEtAl_2022_InducingCausalStructureForInterpretableNN} it is studied how to use a given model $\highlevel$ during training to \emph{induce} a model $\lowlevel$ such that the former is a causal abstraction of the latter. 

\paragraph{Abstraction in diagrams.} Causal abstraction is well aligned with a compositional perspective, and naturally discussed in diagrammatic terms. 
Let $\lowlevel$ and $\highlevel$ be two deterministic open causal models; the `low-level' model $\lowlevel$ with input (exogenous) variables $R$ and non-input (endogenous) variables $V_L$; 
and `high level' model $\highlevel$ with inputs $U$ and non-inputs $V_H = \{X_1,\dots,X_n\}$.\footnote{Here $V_L$ and $V_H$ are all considered outputs of the respective models, but this assumption is just for a simpler presentation and could easily be dropped.} 
The particular semantics category is irrelevant, but for concreteness let's assume $\NN$.

For each high level variable $X \in V_H$, suppose we are given a subset $\pi(X) \subseteq V_L$, such that $\pi(X) \cap \pi(Y) = \emptyset$ for $X \neq Y$ (the `localist' assumption), and a surjective map 
\[
 	\tau_X	\ : \ \pi(X) \ \rightarrow X
\]
Suppose we also have a surjective map:\footnote{Here $\pi(X)$, $R$ and $U$ in the respective (co-)domains each refer to the corresponding product variables (rather than the sets of variables).} 
\[
	\tau \ : \ R \ \rightarrow U
\]

In slight abuse of notation, for any $S \subseteq V_H$, denote the corresponding product variable $\pi(S) = \prod_{X\in S} \pi(X)$, and $\tau_S$ the corresponding product function. In the strongest sense, $\highlevel$ is a causal abstraction of $\lowlevel$ with respect to these alignment maps iff  
$\forall S \subseteq V_H$:
\begin{equation}
	\input{def-causal-abstraction_black_open_noE_setS.tikz}
	\label{eq:def-causal-abstraction}
\end{equation}
\rl{Here we used the notion of \emph{opening} a causal model from \cite{lorenz2023causal}: $\text{open}_X$ essentially `drops' the causal mechanism of $X$ so that a do-intervention on $X$ can be seen as the composition of first opening at $X$ and then feeding in a state of choice; i.e. for any state $x$ the corresponding do-intervention is given by $\Do(X=x)(\modelM) = \text{open}_X(\modelM) \circ x$.} 

In practice, though, with $\lowlevel$ a deep NN, it is unrealistic to even check every such equality of processes, which would require equality for all input states on $R$ and all possible weights of neurons in $\pi(X)$ for all $X$. 
Instead, Geiger \etal\ propose a weaker abstraction condition that quantifies over only the set of input states that can arise through $\lowlevel$ from the actual data distribution on $R$. Let $D \subseteq \text{St}(R)$ denote the set of input states that come with the problem.  In the weaker sense, $\highlevel$ is a causal abstraction of $\lowlevel$ iff the LHS and RHS of~\eqref{eq:def-causal-abstraction} are equal whenever fed in any state of the following form, for $S \subseteq V_H$, $r,r' \in D$.   
\begin{equation}
	\input{def-causal-abstraction-int-int_just_states_black.tikz}	
	\label{eq:def-causal-abstraction-int-int-states}
\end{equation}

\paragraph{Interpretability from abstraction. }
What are the interpretability benefits of such an abstraction relation? Suppose that the high-level model $\highlevel$ does come with a complete interpretation $\Interp_H$ (as is the point here). 
If a causal abstraction relation obtains as above, then morally this induces a (partial) abstract interpretation for  $\lowlevel$ by `factoring through' $\highlevel$, i.e.~by assigning the interpretation $\Interp_H(X)$ to the set of variables (neurons) $\pi(X)$ for each $X \in V_H$. 
Note, though, that this is not formally an interpretation of $\lowlevel$ itself, which would have to concern individual generators of $\lowlevel$.\footnote{
Nor does the partitioning of $V_L$ induced by $\pi$ generally induce a correspondingly coarsened causal model $\widetilde{\lowlevel}$ with variables $(\pi(X))_{X\in V_H}$ in a straightforward sense, which then could have a partial abstract interpretation induced by $\Interp_H$ in the formal sense. Given causal model $\modelM$ with variables $V$ and causal structure $G_{\modelM}$, a partition $(Y_1,...,Y_k)$ of $V$, i.e. $\cup_{j=1}^k Y_j = V$, does not in general induce a causal model with coarsened variables $Y_1,...,Y_k$, by just lumping together mechanisms to form new ones. This only works if the partitioning plays well with the causal structure -- roughly speaking, if it corresponds to a way of `partitioning' the network diagram $D_{\modelM}$ into fragments. 
Moreover, even if a partitioning does, formally speaking, define a coarsened causal model $\widetilde{\modelM}$, the latter generally becomes detached from the phenomenon the original $\modelM$ is describing, for in reality one cannot intervene on the new coarsened variables independently -- generically, coarsening destroys the autonomy of mechanisms.} 

While causal abstraction thus won't generally yield an explicit interpreted causal structure for the model $\lowlevel$ itself, it provides a yet richer form of `induced' causal model than discussed in Sec.~\ref{sec:CI-and-causal-XAI}, and we agree in spirit with the judgement in \cite{GeigerEtAl_2022_InducingCausalStructureForInterpretableNN} that it 
``is not a story about the reasoning a neural network might use to achieve its behaviour, but instead is an intervention-based method that determines how it does, in fact, achieve its behaviour. We can interpret the semantic content of neural representations using the high-level variables they are aligned with, and understand how those neural representations are composed using the high-level parenthood relation." \cite{GeigerEtAl_2022_InducingCausalStructureForInterpretableNN} 
Moreover, Geiger \etal\ argue that many standard techniques in XAI 
can be seen as causal abstraction analysis, turning the concept into a general approach to ``providing faithful and interpretable explanations of AI models" \cite{GeigerEtAl_2021_CausalAbstracttoinOfNN, GeigerEtAl_2022_InducingCausalStructureForInterpretableNN}. 
One might also argue that causal abstraction answers a key question that comes with causal representation learning \cite{SchoelkopfEtAl_2021_TowardCausalRepresentationLearning}, namely `how much' causal knowledge and in `what form' does one have to provide to the model design and training so as to guarantee the learning of meaningful causal relata. 

There are notable generalisations of causal abstraction, including approximate ones and those where the localist assumption is dropped.
We leave a full categorical presentation, including possible extensions from causal to more general compositional models, for future work; \rl{also see Sec.~\ref{sec:future}.}

\end{document}